\def\year{2021}\relax
\documentclass[letterpaper]{article} 
\usepackage{aaai21} 
\usepackage{times}  
\usepackage{helvet} 
\usepackage{courier}  
\usepackage[hyphens]{url}  
\usepackage{graphicx}

\usepackage{natbib}  
\usepackage{caption} 
\usepackage[utf8]{inputenc} 
\usepackage[T1]{fontenc}    
\usepackage{url}            
\usepackage{booktabs}       
\usepackage{amsfonts}       
\usepackage{nicefrac}       
\usepackage{microtype}      
\usepackage{amsmath}
\usepackage{booktabs}  
\usepackage{caption}   
\usepackage{dsfont}    
\usepackage{multirow}  
\usepackage{mathtools} 
\usepackage{xcolor}    
\usepackage{siunitx}   
\usepackage{subfig}    

\usepackage[hidelinks]{hyperref} 

\usepackage[titletoc]{appendix}

\newcommand{\bfth}{\boldsymbol{\theta}}
\newcommand{\miniboone}{\textsc{Miniboone}}
\newcommand{\power}{\textsc{Power}}
\newcommand{\gas}{\textsc{Gas}}
\newcommand{\hepmass}{\textsc{Hepmass}}
\newcommand{\bsds}{\textsc{Bsds300}}

\newcommand{\bfb}{\boldsymbol{b}}

\newcommand{\bfq}{\boldsymbol{q}}
\newcommand{\bfs}{\boldsymbol{s}}
\newcommand{\bfu}{\boldsymbol{u}}
\newcommand{\bfv}{\mathbf{v}}
\newcommand{\bfw}{\boldsymbol{w}}
\newcommand{\bfx}{\boldsymbol{x}}
\newcommand{\bfy}{\boldsymbol{y}}
\newcommand{\bfz}{\boldsymbol{z}}

\newcommand{\bfA}{\boldsymbol{A}}

\newcommand{\bfE}{\boldsymbol{E}}

\newcommand{\bfJ}{\boldsymbol{J}}
\newcommand{\bfK}{\boldsymbol{K}}

\newcommand{\bfQ}{\boldsymbol{Q}}

\newcommand{\bfX}{\boldsymbol{X}}

\newcommand{\hf}{\frac{1}{2}}
\newcommand{\model}{OT-Flow} 
\def\eu{\ensuremath{\mathrm{e}}}
\def\du{\ensuremath{\mathrm{d}}}
\def\tr{\operatorname{tr}}
\newcommand{\R}{\ensuremath{\mathds{R}}}
\newcommand{\E}{\ensuremath{\mathds{E}}}
\newcommand{\D}{\ensuremath{\mathds{D}}}

\newcommand{\bigO}{\ensuremath{\mathcal{O}}}
\newcommand{\bbR}{\R}

\usepackage{pifont}
\definecolor{myred}{RGB}{215,48,39}
\definecolor{mygreen}{RGB}{26,152,80}
\newcommand{\cmark}{\textcolor{mygreen}{\ding{51}}}
\newcommand{\xmark}{\textcolor{myred}{\ding{55}}}

\title{OT-Flow: Fast and Accurate Continuous Normalizing Flows \\ via Optimal Transport}
\author{
	Derek Onken,\textsuperscript{\rm 1}
	Samy Wu Fung,\textsuperscript{\rm 2}
	Xingjian Li,\textsuperscript{\rm 3}
	Lars Ruthotto,\textsuperscript{\rm 3,1}
    \\
}
\affiliations {
    \textsuperscript{\rm 1} Department of Computer Science, Emory University \\
    \textsuperscript{\rm 2} Department of Mathematics, University of California, Los Angeles \\
    \textsuperscript{\rm 3} Department of Mathematics, Emory University \\
    donken@emory.edu, swufung@math.ucla.edu, xingjian.li@emory.edu, lruthotto@emory.edu
}

\begin{document}

\maketitle

\begin{abstract}
	A normalizing flow is an invertible mapping between an arbitrary probability distribution and a standard normal distribution; it can be used for density estimation and statistical inference. Computing the flow follows the change of variables formula and thus requires invertibility of the mapping and an efficient way to compute the determinant of its Jacobian. To satisfy these requirements, normalizing flows typically consist of carefully chosen components. Continuous normalizing flows (CNFs) are mappings obtained by solving a neural ordinary differential equation (ODE). The neural ODE's dynamics can be chosen almost arbitrarily while ensuring invertibility. Moreover, the log-determinant of the flow's Jacobian can be obtained by integrating the trace of the dynamics' Jacobian along the flow. Our proposed OT-Flow approach tackles two critical computational challenges that limit a more widespread use of CNFs. First, OT-Flow leverages optimal transport (OT) theory to regularize the CNF and enforce straight trajectories that are easier to integrate.	Second, OT-Flow features exact trace computation with time complexity equal to trace estimators used in existing CNFs. On five high-dimensional density estimation and generative modeling tasks, OT-Flow performs competitively to state-of-the-art CNFs while on average requiring one-fourth of the number of weights with an 8x speedup in training time and 24x speedup in inference.
\end{abstract}

\section{Introduction} 
\label{sec:intro}

	A normalizing flow~\cite{rezende2015} is an invertible mapping $f \colon \bbR^d \to \bbR^d$ between an arbitrary probability distribution and a standard normal distribution whose densities we denote by $\rho_0$ and $\rho_1$, respectively.
	By the change of variables formula, for all $\bfx \in \R^d$, the flow must satisfy~\cite{rezende2015,papamakarios2019normalizing}%
	\begin{equation} \label{eq:finite_flow}%
		\log \rho_0(\bfx) = \log \rho_1(f(\bfx)) + \log \left| \, \det \nabla f(\bfx) \,  \right| .
	\end{equation}%
	Given $\rho_0$, a normalizing flow is constructed by composing invertible layers to form a neural network and training their weights.
	Since computing the log-determinant in general requires $\bigO(d^3)$ floating point operations (FLOPS), effective normalizing flows consist of layers whose Jacobians have exploitable structure (e.g., diagonal, triangular, low-rank). 
	
	\begin{figure}[t]
		\centering
		\includegraphics[width = 0.83\columnwidth]{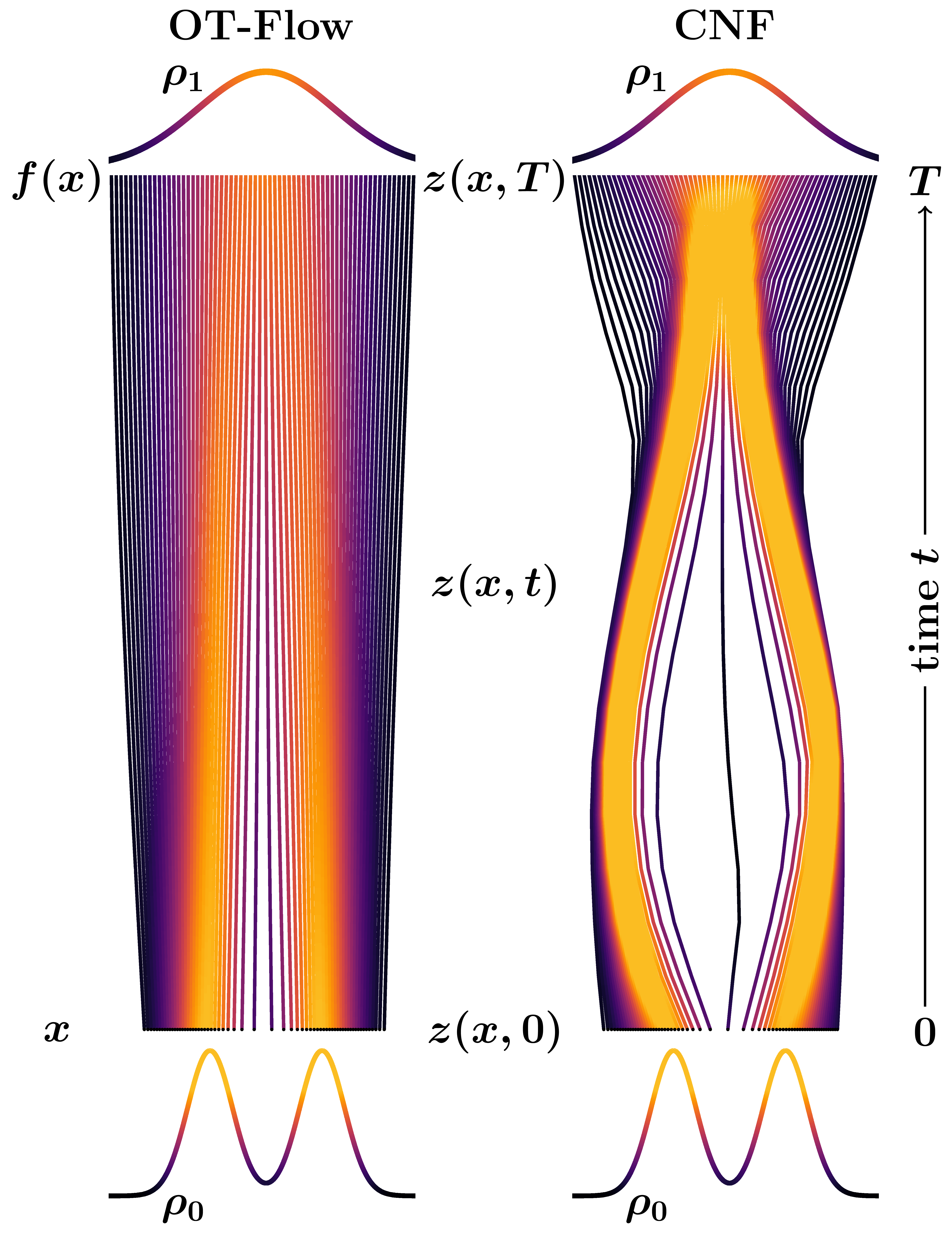}
		\caption{Two flows with approximately equal loss (modification of Fig.~1 in~\citealp{grathwohl2019ffjord,finlay2020train}). While \model{} enforces straight trajectories, a generic CNF can have curved trajectories.}
		\label{fig:pretty_cnf}    
	\end{figure}

	\begin{figure*}[t]
	  \centering
		\includegraphics[width=0.99\textwidth]{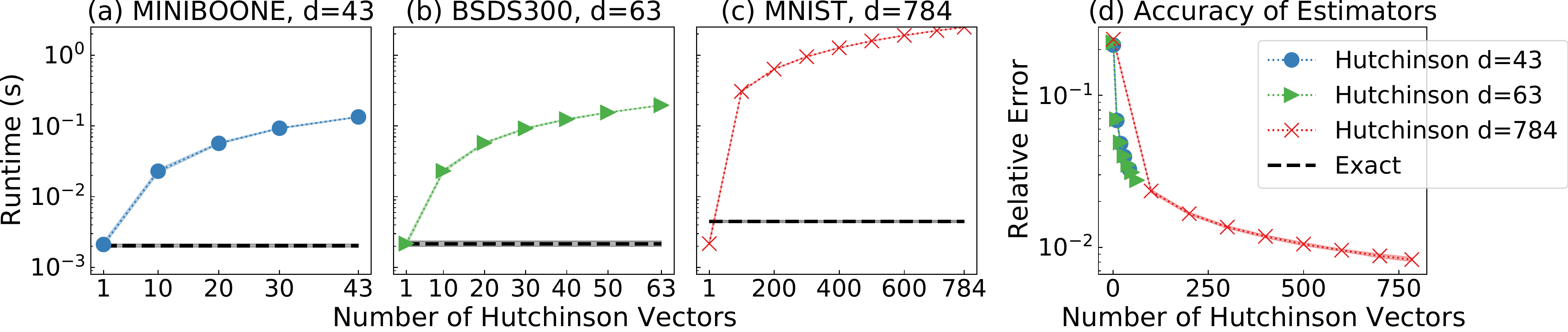}
	  \caption{Performance comparison of trace computation using exact approach presented in Sec.~\ref{sec:trace} and Hutchinson's trace estimator using automatic differentiation. (a-c): runtimes (in seconds) over dimensions 43, 63, and 784, corresponding to the \miniboone{}, \bsds{}, and MNIST data sets, respectively. (d): relative errors vs. number of Hutchinson vectors for different dimensions. We present means with shaded 99\% error bounds computed from twenty runs via bootstrapping (App.~\ref{app:errors}). }
	  \label{fig:trace_compare}
	\end{figure*}

	Alternatively, in continuous normalizing flows (CNFs), $f$ is obtained by solving the neural ordinary differential equation (ODE)~\cite{chen2018neural,grathwohl2019ffjord} 
	\begin{equation} 
		\begin{split}
		  \label{eq:neural_odes}
		  \partial_t \left[ \begin{array}{c}
		      \bfz(\bfx,t)\\
		      \ell(\bfx,t)\\
		    \end{array} \right]
		    =
		    \left[  \begin{array}{c}
		      \bfv \big(\bfz(\bfx,t), t; \bfth \big) \\
		      \tr \big( \nabla \bfv(\bfz(\bfx,t),t; \bfth) \big) \\
		    \end{array}  \right], \\
		    \left[  \begin{array}{c}
		      \bfz(\bfx,0) \\
		      \ell(\bfx,0) \\
		    \end{array}  \right] = 
		    \left[  \begin{array}{c}
		      \bfx \\
		      0    \\
		    \end{array}  \right],
		\end{split}
	\end{equation} 
	for artificial time $t \in [0,T]$ and $\bfx\in\R^d$.
	The first component maps a point $\bfx$ to $f(\bfx)=\bfz(\bfx,T)$ by following the trajectory $\bfz \colon \R^d \times [0,T] \to \R^d$ (Fig.~\ref{fig:pretty_cnf}).
	This mapping is invertible and orientation-preserving under mild assumptions on the dynamics $\bfv \colon \R^d \times [0,T] \to \R^d$.
	The final state of the second component satisfies $\ell(\bfx,T)=\log\det\nabla f(\bfx)$, which can be derived from the instantaneous change of variables formula as in~\citet{chen2018neural}. 
	Replacing the log determinant with a trace reduces the FLOPS to $\bigO(d^2)$ for exact computation or $\bigO(d)$ for an unbiased but noisy estimate~\cite{zhang2018monge,grathwohl2019ffjord,finlay2020train}.

	To train the dynamics, CNFs minimize the expected negative log-likelihood given by the right-hand-side in~\eqref{eq:finite_flow}~\cite{rezende2015,papamakarios2017masked,papamakarios2019normalizing,grathwohl2019ffjord} via
	\begin{equation}
		\label{eq:KLbasedLoss}
		\begin{split}
		\min_{\bfth} \;\; \E_{\rho_0(\bfx)} \;  \left\{ C(\bfx,T) \right\} , \quad \text{for} \\
		C(\bfx,T) 
		\coloneqq
		\hf \| \bfz(\bfx,T) \|^2 
		- \ell(\bfx, T) + \frac{d}{2}\log(2 \pi),
		\end{split}
	\end{equation}
	where for a given $\bfth$, the trajectory $\bfz$ satisfies the neural ODE~\eqref{eq:neural_odes}. 
	We note that the optimization problem~\eqref{eq:KLbasedLoss} is equivalent to minimizing the Kullback-Leibler (KL) divergence between $\rho_1$ and the transformation of $\rho_0$ given by $f$ (derivation in App.~\ref{app:G} or~\citealp{papamakarios2019normalizing}).

	CNFs are promising but come at considerably high costs.
	They perform well in density estimation~\cite{chen2017continuous,grathwohl2019ffjord,papamakarios2019normalizing} and inference~\cite{ingraham2018learning,papamakarios2019normalizing}, especially in physics and computational chemistry~\cite{noe2019boltzmann,brehmer2020madminer}.
	CNFs are computationally expensive for two predominant reasons.
	First, even using state-of-the-art ODE solvers, the computation of~\eqref{eq:neural_odes} can require a substantial number of evaluations of $\bfv$; this occurs, e.g., when the neural network parameters lead to a stiff ODE or dynamics that change quickly in time~\cite{ascher2008numerical}.
    Second, computing the trace term in~\eqref{eq:neural_odes} without building the Jacobian matrix is challenging.
    Using automatic differentiation (AD) to build the Jacobian requires separate vector-Jacobian products for all $d$ standard basis vectors, which amounts to $\bigO(d^2)$ FLOPS. Trace estimates, used in many CNFs~\cite{zhang2018monge,grathwohl2019ffjord,finlay2020train}, reduce these costs but introduce additional noise (Fig.~\ref{fig:trace_compare}). Our approach, \model{}, addresses these two challenges.

	\paragraph{Modeling Contribution}
	Since many flows exactly match two densities while achieving equal loss $C$ (Fig.~\ref{fig:pretty_cnf}), we can choose a flow that reduces the number of time steps required to solve~\eqref{eq:neural_odes}. 
	To this end, we phrase the CNF as an optimal transport (OT) problem by adding a transport cost to~\eqref{eq:KLbasedLoss}. 
    From this reformulation, we exploit the existence of a potential function whose derivative defines the dynamics $\bfv$.
    This potential satisfies the Hamilton-Jacobi-Bellman (HJB) equation, which arises from the optimality conditions of the OT problem. 
    By including an additional cost, which penalizes deviations from the HJB equations, we further reduce the number of necessary time steps to solve~\eqref{eq:neural_odes} (Sec.~\ref{sec:OMT-Flow}).
	Ultimately, encoding the underlying regularity of OT into the network absolves it from learning unwanted dynamics, substantially reducing the number of parameters required to train the CNF.

	\paragraph{Numerical Contribution}
	To train the flow with reduced time steps, we opt for the discretize-then-optimize approach and use AD for the backpropagation (Sec.~\ref{sec:trace}).
	Moreover, we analytically derive formulas to efficiently compute the exact trace of the Jacobian in~\eqref{eq:neural_odes}.
	We compute the \emph{exact} Jacobian trace with $O(d)$ FLOPS, matching the time complexity of \emph{estimating} the trace with one Hutchinson vector as used in state-of-the-art CNFs~\cite{grathwohl2019ffjord,finlay2020train}.
	We demonstrate the competitive runtimes of the trace computation on several high-dimensional examples (Fig.~\ref{fig:trace_compare}).
	Ultimately, our PyTorch implementation\footnote{Code is available at https://github.com/EmoryMLIP/OT-Flow .}
	of \model{} produces results of similar quality to state-of-the-art CNFs at 8x training and 24x inference speedups on average (Sec.~\ref{sec:experiments}).

	\begin{figure*}[t]
	  \centering
		\includegraphics[width=0.47\linewidth]{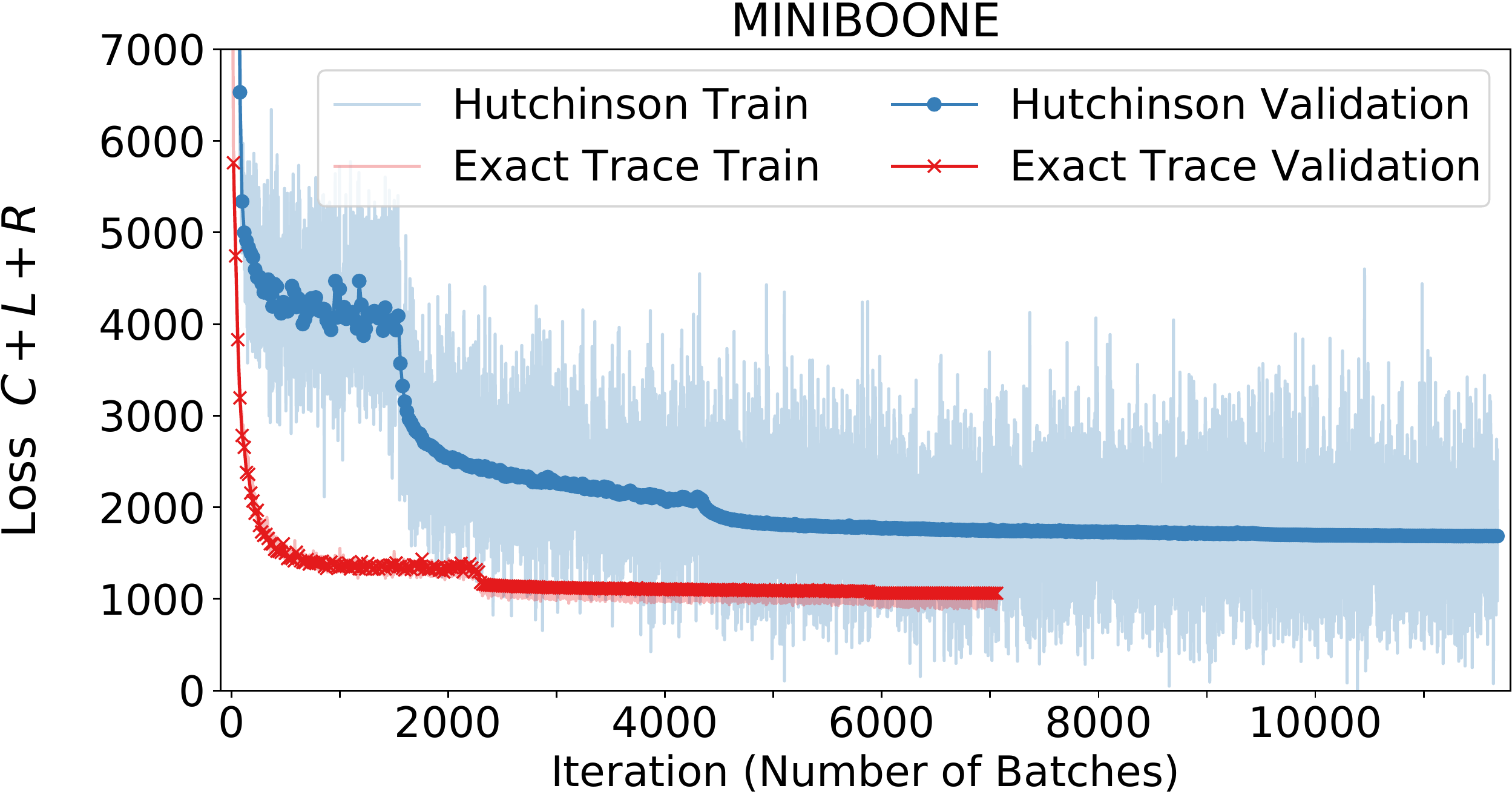} \quad
		\includegraphics[width=0.47\linewidth]{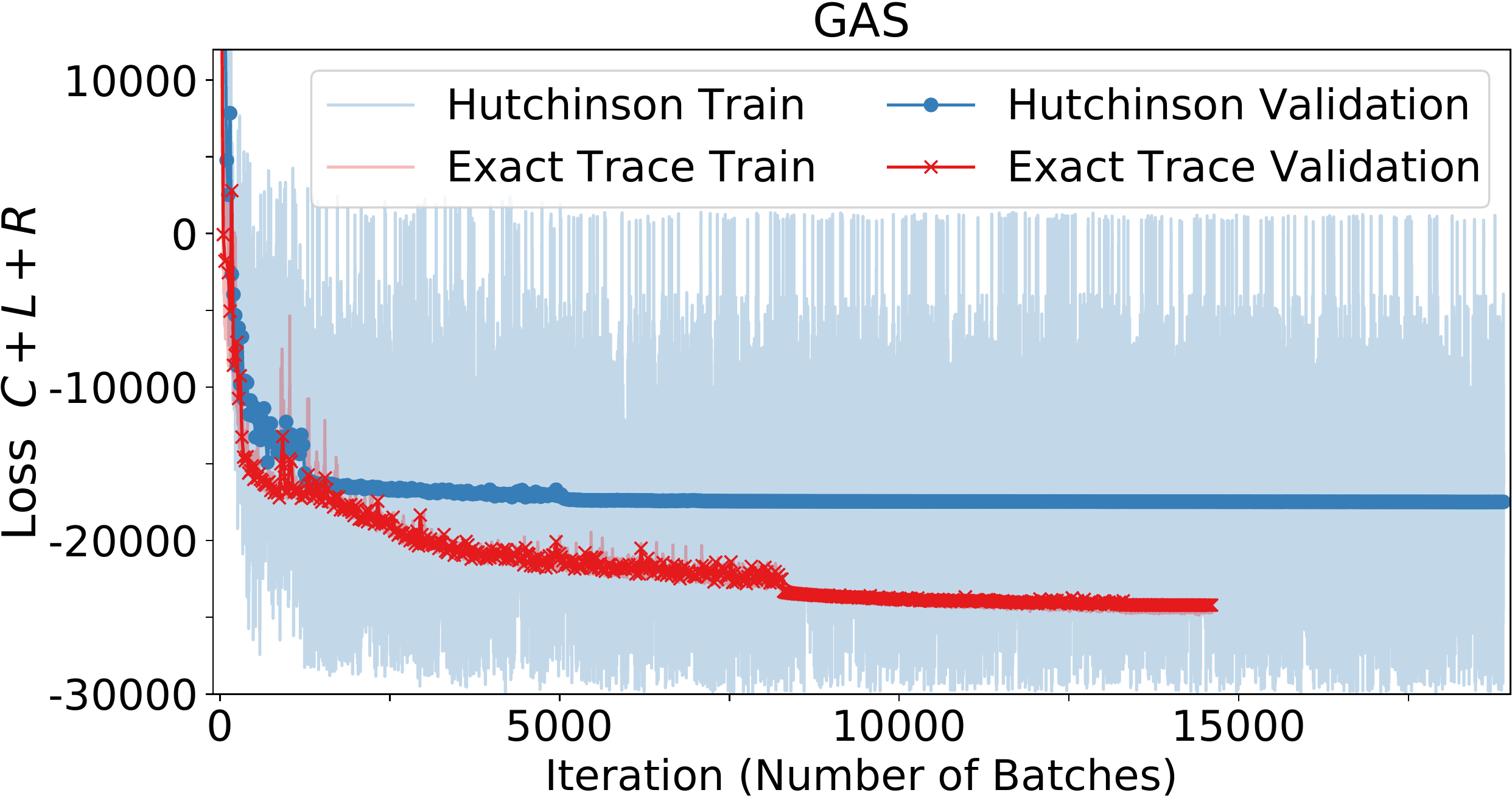}
	  \caption{The exact trace computation in \model{} leads to faster decay of validation loss and lower training loss variance compared to an identical model using a randomized trace estimator (also used in FFJORD and RNODE) for two data sets.}
	  \label{fig:hutch}
	\end{figure*}

\section{Mathematical Formulation of \model{}}
\label{sec:OMT-Flow}

    Motivated by the similarities between training CNFs and solving OT problems~\cite{benamou2000computational,peyre2018computational}, we regularize the minimization problem~\eqref{eq:KLbasedLoss} as follows. 
    First, we formulate the CNF problem as an OT problem by adding a transport cost.
    Second, from OT theory, we leverage the fact that the optimal dynamics $\bfv$ are the negative gradient of a potential function $\Phi$, which satisfies the HJB equations. Finally, we add an extra term to the learning problem that penalizes violations of the HJB equations. This reformulation encourages straight trajectories (Fig.~\ref{fig:pretty_cnf}).

    \paragraph{Transport Cost} We add the $L_2$ transport cost
    \begin{equation}
		\label{eq:L2}
		L(\bfx,T) = \int_0^T \frac{1}{2}\| \bfv \big( \bfz(\bfx,t),t \big) \|^2 \, \du t,
	\end{equation}
    to the objective in~\eqref{eq:KLbasedLoss}, which results in the regularized problem
    \begin{equation}
		\label{eq:LregODE}
		\min_{\bfth} \;\; \E_{\rho_0(\bfx)} \; \Big\{C(\bfx, T) + L(\bfx,T) \Big\} \quad \text{s.t. } \; \eqref{eq:neural_odes}.
	\end{equation}
    This transport cost penalizes the squared arc-length of the trajectories. 
    In practice, this integral can be computed in the ODE solver, similar to the trace accumulation in~\eqref{eq:neural_odes}.
    The OT problem~\eqref{eq:LregODE} is the relaxed Benamou-Brenier formulation, i.e., the final time constraint is given here as the soft constraint $C(\bfx, T)$. This formulation has mathematical properties that we exploit to reduce computational costs~\cite{evans1997partial,villani2008optimal,lin2019fluid,finlay2020train}.
    In particular, \eqref{eq:LregODE} is now equivalent to a \emph{convex} optimization problem (prior to the neural network parameterization), and the trajectories matching the two densities $\rho_0$ and $\rho_1$ are straight and non-intersecting~\cite{gangbo1996geometry}. This reduces the number of time steps required to solve~\eqref{eq:neural_odes}.
    The OT formulation also guarantees a solution flow that is smooth, invertible, and orientation-preserving~\cite{ambrosio2008gradient}.

    \paragraph{Potential Model} 
    We further capitalize on OT theory by incorporating additional structure to guide our modeling. 
    In particular, from the Pontryagin Maximum Principle~\cite{evans1983introduction,evans2010partial}, there exists a potential function $\Phi \colon \R^d \times[0,T] \to\R$ such that
	\begin{equation}
		 \label{eq:pontryagin}
		 \bfv(\bfx,t; \bfth) = - \nabla \Phi(\bfx,t;\bfth).
	\end{equation}
	Analogous to classical physics, samples move in a manner to minimize their potential.
	In practice, we parameterize $\Phi$ with a neural network instead of $\bfv$.
	Moreover, optimal control theory states that $\Phi$ satisfies the HJB equations~\cite{evans1983introduction}
	\begin{equation}
		\label{eq:HJB_main}
		\begin{split}
		 -\partial_t &\Phi(\bfx,t) + \frac{1}{2}\|\nabla \Phi(\bfz(\bfx,t),t)\|^2 = 0
		 \\ 
		\Phi(\bfx,T) &= 1 + \log\big(\rho_0(\bfx)\big)  - \log\big(\rho_1(\bfz(\bfx,T))\big) 
		\\
        &- \ell(\bfz(\bfx,T),T).
		\end{split}
	\end{equation}
	We derive the terminal condition in App.~\ref{app:HJBReg}.
	The existence of this potential allows us to reformulate the CNF in terms of $\Phi$ instead of $\bfv$ and add an additional regularization term which penalizes the violations of~\eqref{eq:HJB_main} along the trajectories by 
	\begin{equation}
		\label{eq:HJB_ode}
		 R(\bfx, T) = \int_0^T \left| \partial_t \Phi \big(\bfz(\bfx,t),t \big) - \frac{1}{2}\|\nabla \Phi \big( \bfz(\bfx,t),t \big)\|^2 \right| \, \du t.
	\end{equation}
    This HJB regularizer $R(\bfx, T)$ favors plausible $\Phi$ without affecting the solution of the optimization problem~\eqref{eq:LregODE}.

    With implementation similar to $L(\bfx,T)$, the HJB regularizer $R$ requires little computation, but drastically simplifies the 
    cost of solving~\eqref{eq:neural_odes} in practice.
    We assess the effect of training a toy Gaussian mixture problem with and without the HJB regularizer (Fig.~\ref{fig:effectOfHJBReg} in Appendix). For this demonstration, we train a few models using varied number of time steps and regularizations. For unregularized models with few time steps, we find that the $L_2$ cost is not penalized at enough points. Therefore, without an HJB regularizer, the model achieves poor performance and unstraight characteristics (Fig.~\ref{fig:effectOfHJBReg}). This issue can be remedied by adding more time steps or the HJB regularizer (see examples in~\citealp{yang2019,ruthotto2020machine,lin2020apac}). Whereas additional time steps add significant computational cost and memory, the HJB regularizer is inexpensive as we already compute $\nabla \Phi$ for the flow.

    \paragraph{\model{} Problem} In summary, the regularized problem solved in \model{} is 
	\begin{equation}
		\label{eq:MFGFlowObj}
		\begin{split}
		\min_{\bfth} \;\; \E_{\rho_0(\bfx)} \; \Big\{C(\bfx, T) + L(\bfx,T) + R(\bfx,T) \Big\}, \\ \text{subject to} \quad \eqref{eq:neural_odes}, 
		\end{split}
	\end{equation}
	combining aspects from~\citet{zhang2018monge},~\citet{grathwohl2019ffjord},~\citet{yang2019}, and~\citet{finlay2020train} (Tab.~\ref{tab:comparison}).
	The $L_2$ and HJB terms add regularity and are accumulated along the trajectories. As such, they make use of the ODE solver and computed $\nabla \Phi$ (App.~\ref{app:implementation}).

\begin{table*}[t]
\centering
\renewcommand{\arraystretch}{1.1}
\addtolength{\tabcolsep}{-1pt} 
\begin{tabular}{lccccccccc}
	\toprule
	\multirow{2}{*}{\vspace{-5pt}Model} &  \multicolumn{5}{c}{Formulation} & \multicolumn{3}{c}{Training Implementation } & Inference\\
	\cmidrule(lr){2-6} \cmidrule(lr){7-9} \cmidrule(lr){10-10}
 & ODEs~\eqref{eq:neural_odes} & $\Phi$ & $L$ & $R$ & $\| \nabla \bfv \|_F^2$ & ODE Solver & Approach & Trace & Trace\\
	\midrule
	FFJORD 				& \cmark & \xmark & \xmark & \xmark & \xmark & Runge-Kutta (4)5  & OTD & Hutch w/ Rad & exact w/ AD loop\\
	RNODE 				&\cmark & \xmark & \cmark & \xmark & \cmark & Runge-Kutta 4 & OTD & Hutch w/ Rad & exact w/ AD loop\\
	Monge-Ampère	&\cmark & \cmark & \xmark & \xmark & \xmark & Runge-Kutta 4 & DTO & \multicolumn{2}{c}{Hutch w/ Gauss}\\
	Potential Flow &\cmark & \cmark & \xmark & \cmark & \xmark & Runge-Kutta 1 & DTO & \multicolumn{2}{c}{exact w/ AD loop}\\
	\textbf{\model{}} 	& \cmark & \cmark & \cmark & \cmark & \xmark & Runge-Kutta 4 & DTO & \multicolumn{2}{c}{efficient exact (Sec.~\ref{sec:trace})}\\
	\bottomrule
\end{tabular}
\addtolength{\tabcolsep}{1pt} 
\caption{All methods share the underlying neural ODEs but differ in use of a potential $\Phi$, regularizers ($L$, $R$, $\| \nabla \bfv \|_F^2$), ODE solver, approach (discretize-then-optimize DTO or optimize-then-discretize OTD), and trace computation (exact using automatic differentiation AD, Hutchinson's estimator with a single vector sampled from a Rademacher or Gaussian distribution).}
\label{tab:comparison}
\end{table*}

\section{Implementation of \model{}} 
\label{sec:trace}

We define our model, derive analytic formulas for fast and exact trace computation, and describe our efficient ODE solver.

\paragraph{Network}
	We parameterize the potential as%
	\begin{equation}%
	\label{eq:NNArchitecture}
	\begin{split}
	  \Phi(\bfs ; \bfth) = \, &\bfw^\top N(\bfs;\bfth_N) + \frac{1}{2} \bfs^\top (\bfA^\top 
	  \bfA)\bfs + \bfb^\top \bfs + c,\\ &\text{where} \quad\bfth = (\bfw, \bfth_N, \bfA, \bfb, c).
	\end{split}
	\end{equation}%
	Here, $\bfs = (\bfx,t) \in \R^{d+1}$ are the input features corresponding to space-time, $N(\bfs;\bfth_N) \colon \R^{d+1} \to \R^m$ is a neural network chosen to be a residual neural network (ResNet)~\cite{he2016deep} in our experiments, and $\bfth$ consists of all the trainable weights: $\bfw \in \R^m$, $\bfth_N \in \R^p$, $\bfA \in \R^{r \times (d+1)}$, $\bfb \in \R^{d+1}$, $c \in \R$.
	We set a rank $r=\min (10,d)$ to limit the number of parameters of the symmetric matrix $\bfA^\top \bfA$. Here, $\bfA$, $\bfb$, and $c$ model quadratic potentials, i.e., linear dynamics; $N$ models the nonlinear dynamics. This formulation was found to be effective in~\citet{ruthotto2020machine}.

\paragraph{ResNet}
	Our experiments use a simple two-layer ResNet. When tuning the number of layers as a hyperparameter, we found that wide networks promoted expressibility but deep networks offered no noticeable improvement. 
	For simplicity, we present the two-layer derivation (for the derivation of a ResNet of any depth, see App.~\ref{app:trace} or~\citealt{ruthotto2020machine}). 	
	The two-layer ResNet uses an opening layer to convert the $\R^{d+1}$ inputs to the $\R^{m}$ space, then one layer operating on the features in hidden space $\R^{m}$
	\begin{equation}
	\begin{split}
	    \bfu_0 & = \sigma(\bfK_0 \bfs + \bfb_0) \\ 
	   N(\bfs;\bfth_N) =  \bfu_1 & = \bfu_0 + h \, \sigma(\bfK_1 \bfu_0 + \bfb_1).
\bf	\end{split}
	\end{equation}
	We use step-size $h{=}1$, dense matrices $\bfK_0 \in \R^{m \times (d+1)}$ and $\bfK_1 \in \R^{m \times m}$, and biases $\bfb_0, \bfb_1 \in \R^{m}$. We select the element-wise activation function 
	$\sigma(\bfx) =\log(\exp(\bfx) + \exp(-\bfx))$, which is the antiderivative of the hyperbolic tangent, i.e., $\sigma'(\bfx) = \tanh(\bfx)$. Therefore, hyperbolic tangent is the activation function of the flow $\nabla \Phi$.

\newcommand{\rottextTwo}[1]{\rotatebox{90}{\parbox{22mm}{\small \centering#1}}}
\begin{figure*}[t]
  \centering
  \setlength{\tabcolsep}{1.2pt}
  \renewcommand{\arraystretch}{0.5}
  \begin{tabular}{cccccccc}
  	\rottextTwo{Data\\$\bfx$}
  	&
  	\includegraphics[width=0.12\textwidth]{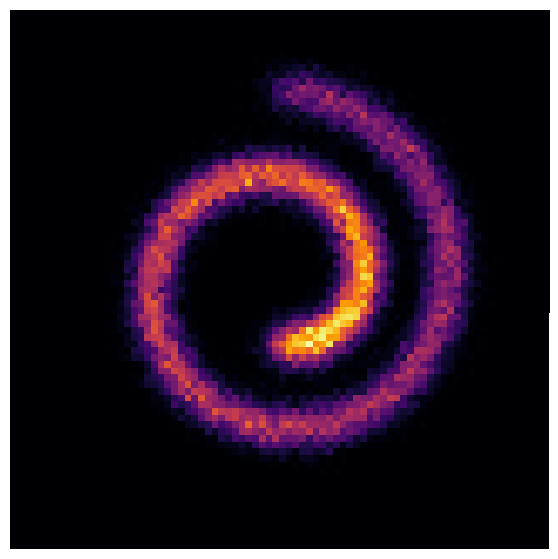}
  	&
  	\includegraphics[width=0.12\textwidth]{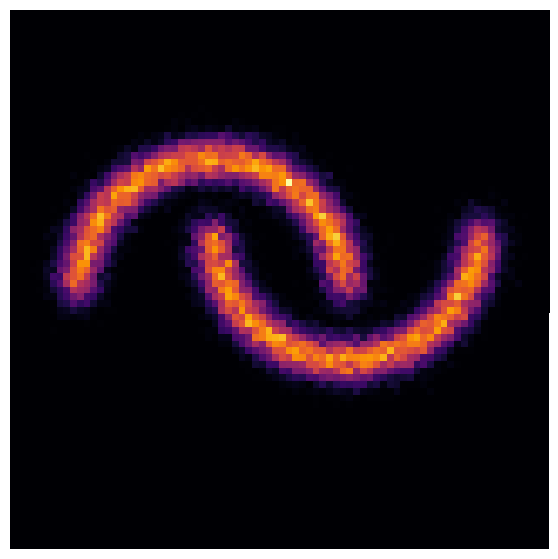}
  	&
  	\includegraphics[width=0.12\textwidth]{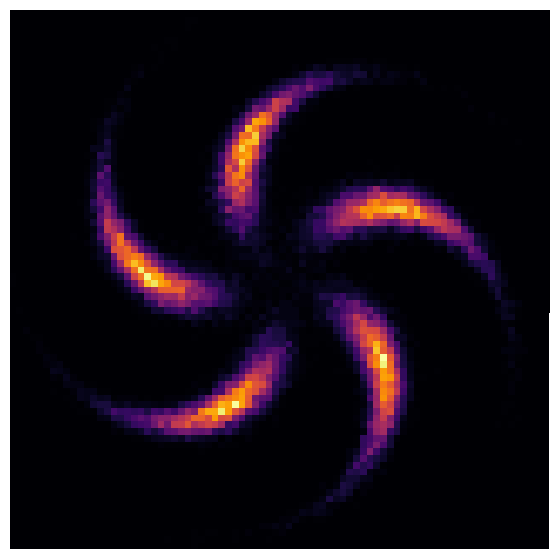}
  	&
  	\includegraphics[width=0.12\textwidth]{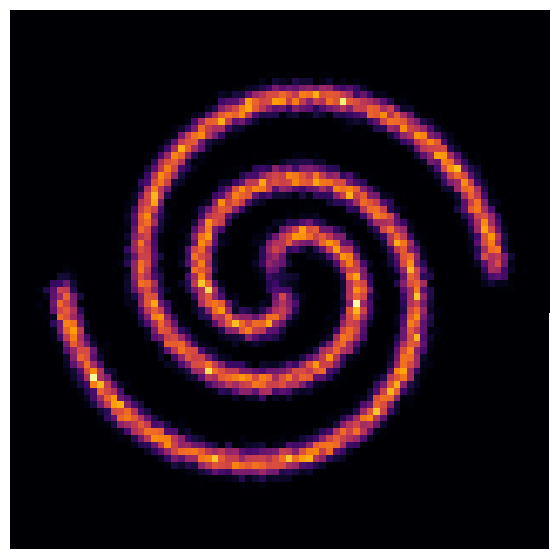}
  	&
  	\includegraphics[width=0.12\textwidth]{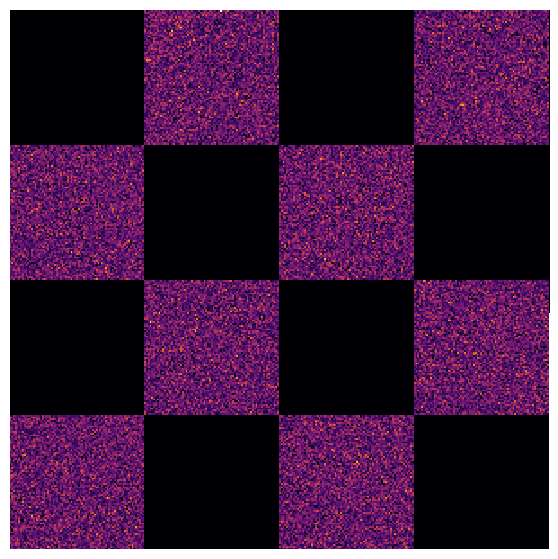}
  	&
  	\includegraphics[width=0.12\textwidth]{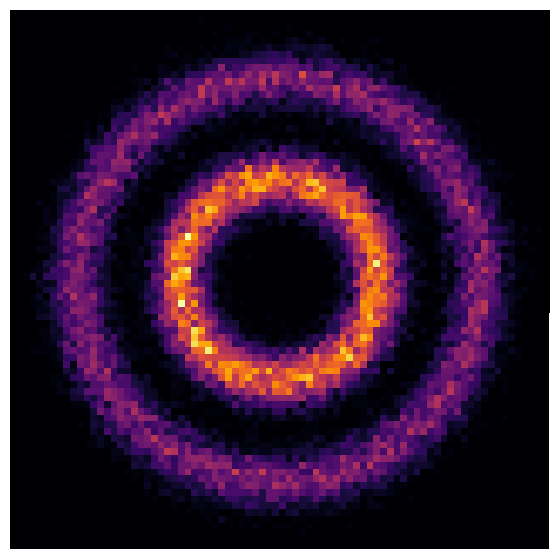}
  	&
  	\includegraphics[width=0.12\textwidth]{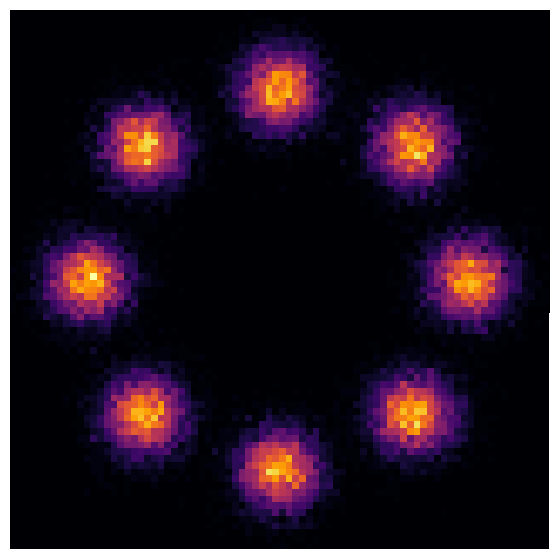}
  	\\
    \rottextTwo{Estimate\\$\rho_0$}
  	&
  	\includegraphics[width=0.12\textwidth]{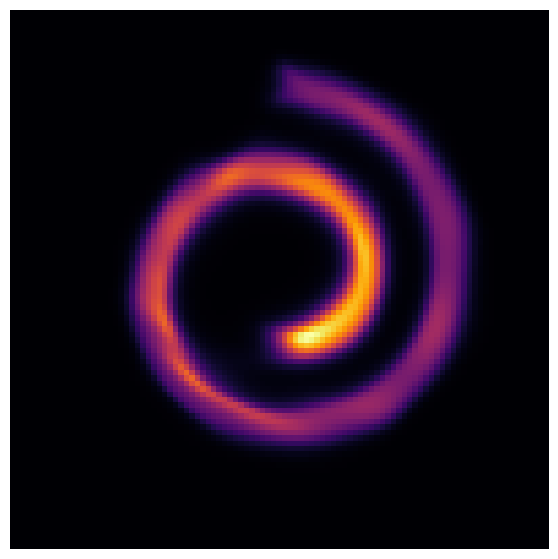}
  	&
  	\includegraphics[width=0.12\textwidth]{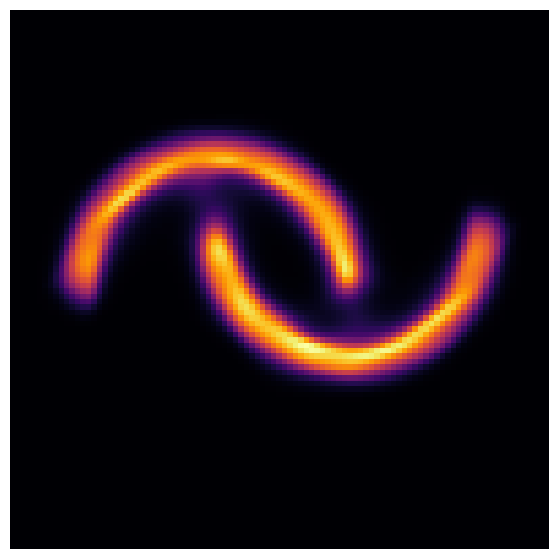}
  	&
  	\includegraphics[width=0.12\textwidth]{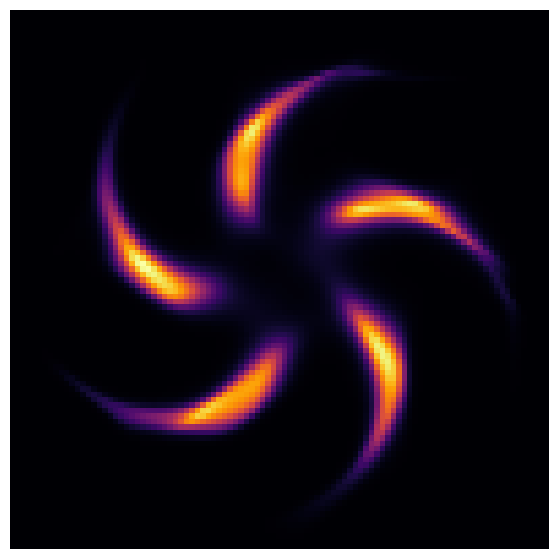}
  	&
  	\includegraphics[width=0.12\textwidth]{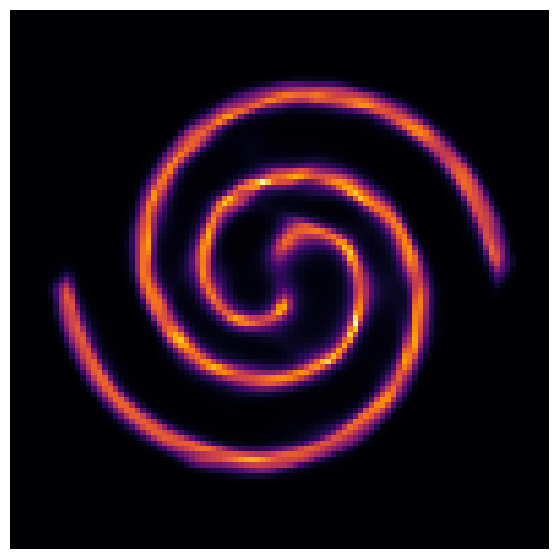}
  	&
  	\includegraphics[width=0.12\textwidth]{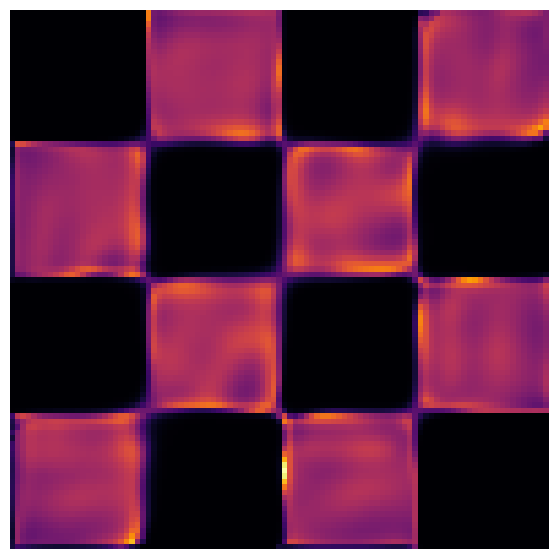}
  	&
  	\includegraphics[width=0.12\textwidth]{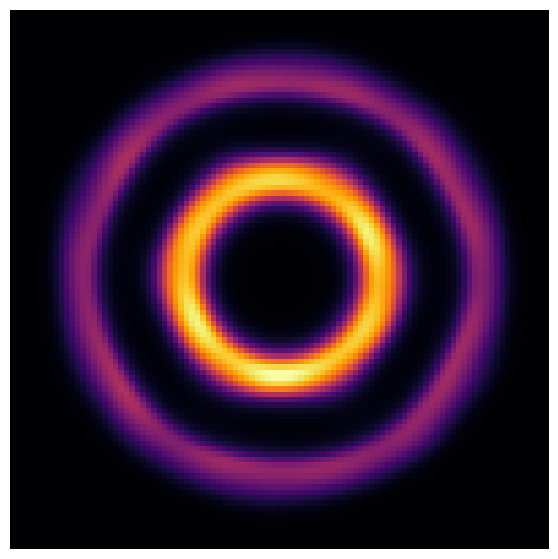}
  	&
  	\includegraphics[width=0.12\textwidth]{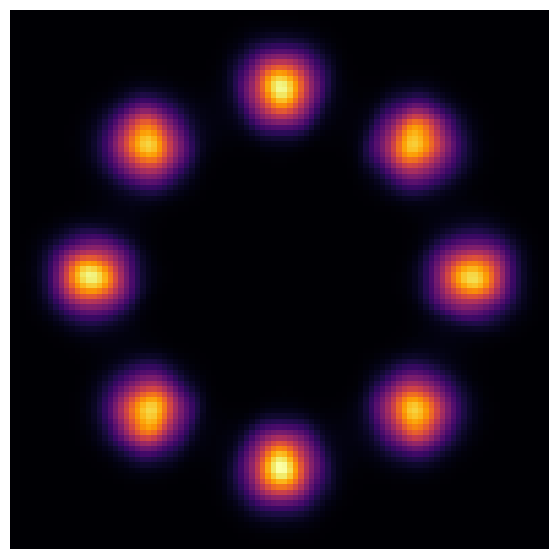}
  	\\
  	\rottextTwo{Generation\\$f^{-1}(\bfy)$}
  	&
  	\includegraphics[width=0.12\textwidth]{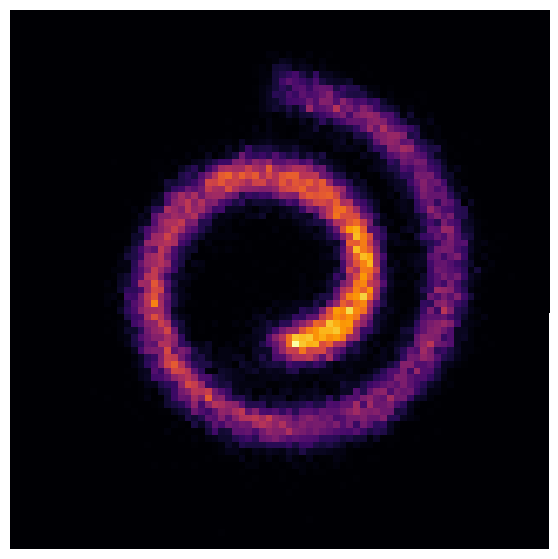}
  	&
  	\includegraphics[width=0.12\textwidth]{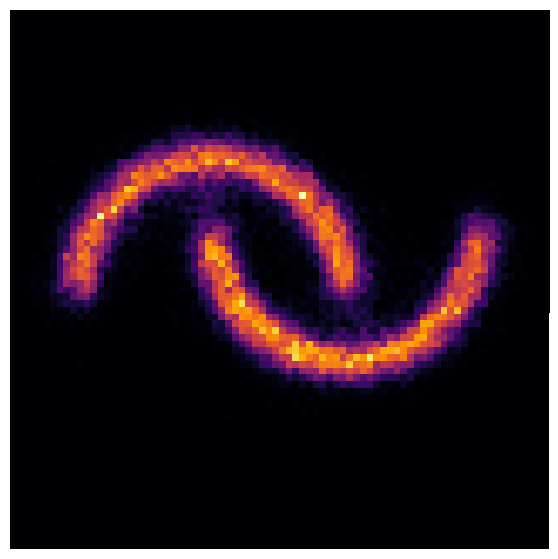}
  	&
  	\includegraphics[width=0.12\textwidth]{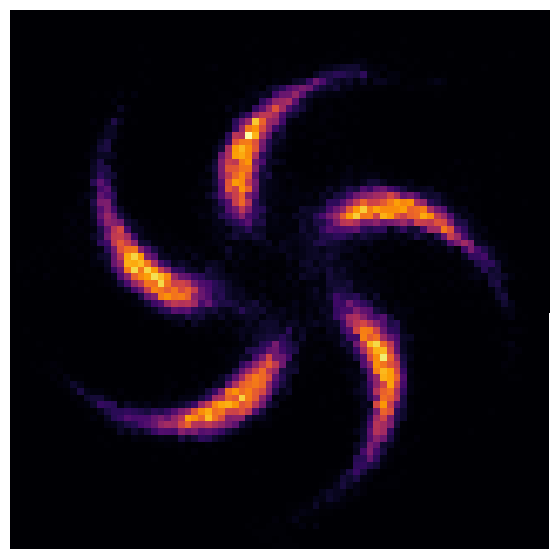}
  	&
  	\includegraphics[width=0.12\textwidth]{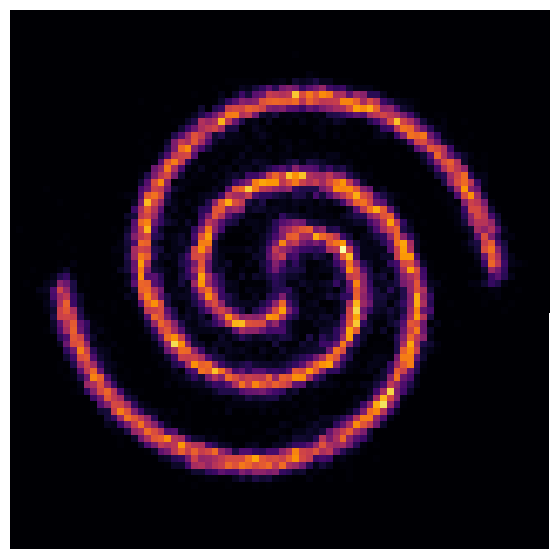}
  	&
  	\includegraphics[width=0.12\textwidth]{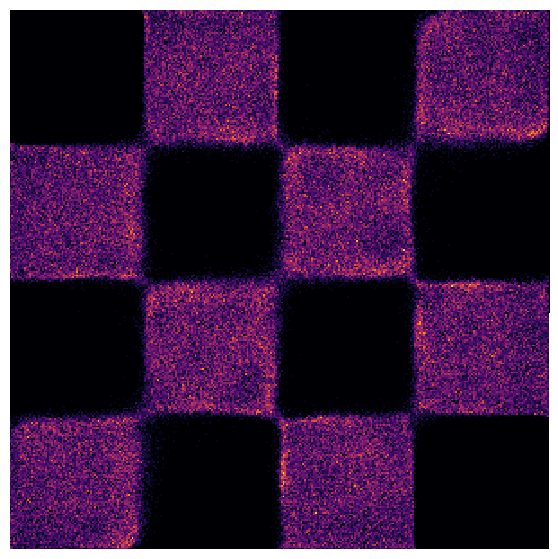}
  	&
  	\includegraphics[width=0.12\textwidth]{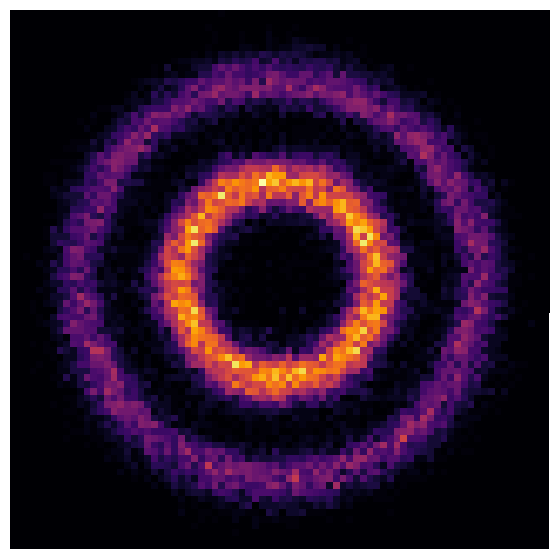}
  	&
  	\includegraphics[width=0.12\textwidth]{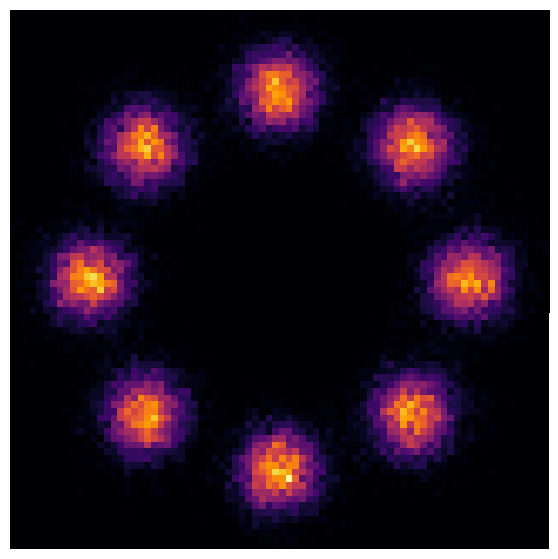}
  \end{tabular}
  \caption{Density estimation on 2-D toy problems. \textbf{Top:} samples from the unknown distribution. \textbf{Middle:} density estimate for unknown $\rho_0$ computed by inverse flowing from $\rho_1$ via~\eqref{eq:neural_odes}.
  \textbf{Bottom:} samples generated by inverse flow where $\bfy \sim \rho_1(\bfy)$.}
  \label{fig:toyModels}
\end{figure*}

\paragraph{Gradient Computation}
	The gradient of the potential is 
	\begin{equation}\label{eq:DPhi}
	    \nabla_{\bfs} \Phi(\bfs;\bfth)  = \nabla_{\bfs} N(\bfs;\bfth_N) \bfw + (\bfA^\top \bfA) \bfs + \bfb,
	\end{equation}
	where we simply take the first $d$ components of $\nabla_{\bfs} \Phi$ to obtain the space derivative $\nabla \Phi$.
	The first term is computed using chain rule (backpropagation)
	\begin{equation}
	  \begin{split}
	      \bfz_{1} &= \bfw + h  \bfK_{1}^\top {\rm diag}(\sigma'(\bfK_{1} \bfu_{0} + \bfb_{1})) \bfw,
	      \\  
	      \bfz_0 &= \bfK_{0}^\top {\rm diag}(\sigma'(\bfK_{0} \bfs + \bfb_{0})) \bfz_1, \quad \text{where}\\
	      \nabla_{\bfs} &N(\bfs;\bfth_N) \bfw = \bfz_{0}.
	  \end{split}
	\end{equation}
	Here, ${\rm diag}(\bfq) \in\bbR^{m\times m}$ denotes a diagonal matrix with diagonal elements given by $\bfq \in \R^m$. Multiplication by diagonal matrix is implemented as an element-wise product.

\paragraph{Trace Computation} 

	We compute the trace of the Hessian of the potential model.
	We first note that%
	\begin{equation} 
	    \label{eq:laplace}
	    \begin{split}
	     \tr \big(\nabla^2 \Phi(\bfs;\bfth)\big) =  \tr \Big(\bfE^\top \, \nabla_{\bfs}^2 \big( N(\bfs;\bfth_N ) \bfw \big) \, \bfE \Big)\\
	      + \tr \Big(\bfE^\top (\bfA^\top \bfA) \, \bfE \Big),
		\end{split}
	\end{equation}
	where the columns of $\bfE \in \R^{(d+1)\times d}$ are the first $d$ standard basis vectors in $\R^{d+1}$. All matrix multiplications with $\bfE$ can be implemented as constant-time indexing operations. 
	The trace of the $\bfA^\top \bfA$ term is trivial. We compute the ResNet term via
	\begin{equation} \label{eq:trace}
	\begin{split}
	     &\tr \big(\bfE^\top \, \nabla_{\bfs}^2 ( N(\bfs;\bfth_N) \bfw) \,\bfE \big)  = t_0 + h \, t_1,
	     \;\; \text{where}\\ 
	     &t_0 = \big(\sigma''(\bfK_0 \bfs + \bfb_0) \odot \bfz_1 \big)^\top \big((\bfK_0 \bfE)\odot(\bfK_0 \bfE)\big) \mathbf{1}, \\
	     &t_1 = \big(\sigma''(\bfK_1 \bfu_{0} + \bfb_1) \odot \bfw\big)^\top \big((\bfK_1 \bfJ)\odot(\bfK_1 \bfJ)\big) \mathbf{1},
	\end{split}
	\end{equation}
	where $\odot$ is the element-wise product of equally sized vectors or matrices, $\mathbf{1} \in \R^{d}$ is a vector of all ones, and $\bfJ = \nabla \bfu_{0}^\top=$ $ {\rm diag} \big( \sigma'(\bfK_0 \bfs + \bfb_0) \big) ( \bfK_0 \bfE)$. For deeper ResNets, the Jacobian term $\bfJ = \nabla \bfu_{i-1}^\top \in \R^{m\times d}$ can be updated and over-written at a computational cost of $\bigO(m^2 \cdot d)$ FLOPS (App.~\ref{app:trace}).
	
	The trace computation of the first layer uses $\bigO(m\cdot d)$ FLOPS, and each additional layer uses $\bigO(m^2 \cdot d)$ FLOPS (App.~\ref{app:trace}).
	Thus, our exact trace computation has similar computational complexity as FFJORD's and RNODE's trace estimation.
	In clocktime, the analytic exact trace computation is competitive with the Hutchinson's estimator using AD, while introducing no estimation error (Fig.~\ref{fig:trace_compare}).
	Our efficiency in trace computation~\eqref{eq:trace} stems from exploiting the identity structure of matrix $\bfE$ and not building the full Hessian.
	
	We find that using the exact trace instead of a trace estimator improves convergence (Fig.~\ref{fig:hutch}). Specifically, we train an \model{} model and a replicate model in which we only change the trace computation, i.e., we replace the exact trace computation with Hutchinson's estimator using a single random vector. The model using the exact trace (\model{}) converges more quickly and to a lower validation loss, while its training loss has less variance (Fig.~\ref{fig:hutch}). 
	
	Using Hutchinson's estimator without sufficiently many time steps fails to converge~\cite{onken2020do} because such an approach poorly approximates the time integration \emph{and} the trace in the second component of~\eqref{eq:neural_odes}. Whereas FFJORD and RNODE estimate the trace but solve the time integral well, \model{} trains with the exact trace and notably fewer time steps (Tab.~\ref{tab:large}). At inference, all three solve the trace and integration well.

    \paragraph{ODE Solver} For the forward propagation, we use Runge-Kutta 4 with equidistant time steps to solve~\eqref{eq:neural_odes} as well as the time integrals~\eqref{eq:L2} and~\eqref{eq:HJB_ode}. The number of time steps is a hyperparameter.
    For validation and testing, we use more time steps than for training, which allows for higher precision and a check that our discrete \model{} still approximates the continuous object.
    A large number of training time steps prevents overfitting to a particular discretization of the continuous solution and lowers inverse error; too few time steps results in high inverse error but low computational cost. 
	We tune the number of training time steps so that validation and training loss are similar with low computational cost.
    
    For the backpropagation, we use AD.    
    This technique corresponds to the discretize-then-optimize (DTO) approach, an effective method for ODE-constrained optimization problems~\cite{collis2002analysis,abraham2004effect,becker2007optimal,leugering2014trends}.
    In particular, DTO is efficient for solving neural ODEs~\cite{li2017maximum,gholami2019anode,onken2020do}. 
    Our implementation exploits the benefits of our proposed exact trace computation combined with the efficiency of DTO.

\begin{table*}[t]
\centering
    \begin{tabular}{clrrrrrrrrr}
    \toprule
     \multirow{2}{*}{\vspace{-5pt}Data Set} & \multirow{2}{*}{\vspace{-5pt}Model} & \multirow{2}{*}{\vspace{-5pt}\# Param} &  \multicolumn{4}{c}{Training} & \multicolumn{4}{c}{Testing} \\
    \cmidrule(lr){4-7} \cmidrule(lr){8-11} 
     &  &  & Time (h) & \# Iter & $\frac{\text{Time}}{\text{Iter}}$(s) & NFE & Time (s) & Inv Err & MMD & \multicolumn{1}{c}{$C$}\\
    \midrule
    \multirow{3}{*}{\shortstack[*]{\textbf{\power{}}\\$d=$ 6}} &
    	\model{}  & 18K & 3.1  & 22K & 0.56 & 40   & 10.6  & 4.10e-6  & 4.68e-5 & -0.30\\
        & RNODE   & 43K & 25.0 & 32K & 2.78 & 200  & 88.2  & 5.95e-6  & 5.64e-5 & -0.39\\
        & FFJORD  & 43K & 68.9 & 29K & 8.63 & 583  & 72.4  & 7.60e-6  & 4.34e-5 & -0.37\\
    \midrule
    \multirow{3}{*}{\shortstack[*]{\textbf{\gas{}}\\$d=$ 8}} &
    	\model{}  & 127K & 6.1  & 52K & 0.42 & 40  & 30.9  & 1.79e-4  & 2.47e-4 & -9.20\\
        & RNODE   & 279K & 36.3 & 59K & 2.23 & 200 & 763.7 & 2.53e-5  & 8.03e-5 & -11.10\\
        & FFJORD  & 279K & 75.4 & 49K & 5.54 & 475 & 892.4 & 1.78e-5  & 1.02e-4 & -10.69\\
    \midrule
    \multirow{3}{*}{\shortstack[*]{\textbf{\hepmass{}}\\$d=$ 21}}  &
    	\model{} &  72K & 5.2  & 35K & 0.53 & 48  & 47.9   & 2.98e-6 & 1.58e-5 & 17.32\\
        & RNODE  & 547K & 46.5 & 40K & 4.16 & 400 & 446.7  & 1.91e-5 & 1.58e-5 & 16.37\\ 
        & FFJORD & 547K & 99.4 & 47K & 7.56 & 770 & 450.4  & 2.98e-5 & 1.58e-5 & 16.13\\ 
    \midrule                    
    \multirow{3}{*}{\shortstack[*]{\textbf{\miniboone{}}\\$d=$ 43}}  & 
       	\model{} & 78K  & 0.8 &  7K & 0.44 & 24  & 0.8  & 5.65e-6  & 2.84e-4 & 10.55\\
       	& RNODE  & 821K & 1.4 & 15K & 0.33 & 16  & 33.0 & 4.42e-6  & 2.84e-4 & 10.65\\
       	& FFJORD & 821K & 9.0 & 16K & 2.01 & 115 & 31.5 & 4.80e-6  & 2.84e-4 & 10.57\\
    \midrule
    \multirow{3}{*}{\shortstack[*]{\textbf{\bsds{}}\\$d=$ 63}} & 
    	\model{} &  297K & 7.1   &  37K & 0.70 & 56   & 432.7   & 5.54e-5 & 4.24e-4 & -154.20\\
        & RNODE  & 6.7M  & 106.6 &  16K & 23.4 & 200  & 15253.3 & 2.66e-6 & 1.64e-2 & -129.75\\ 
        & FFJORD & 6.7M  & 166.1 &  18K & 33.6 & 345  & 20061.2 & 3.41e-6 & 6.52e-3 & -133.94\\      
    \bottomrule
    \end{tabular}
	\caption{Density estimation on real data sets.
	We present the number of training iterations, the number of function evaluations for the forward ODE solve (NFE), and the time per iteration. For \bsds{} training, FFJORD and RNODE were terminated when validation loss $C$ hit -140. All values are the average across three runs on a single NVIDIA TITAN X GPU with 12GB RAM. We present the standard deviations computed from the three runs in Tab.~\ref{tab:error_bounds} located in the Appendix.} 
	\label{tab:large}
\end{table*}%

\section{Related Works}
\label{sec:relatedWorks}
	
	\paragraph{Finite Flows} 
	Normalizing flows~\cite{tabak2013family,rezende2015,papamakarios2019normalizing,kobyzev2019normalizing} use a composition of discrete transformations, where specific architectures are chosen to allow for efficient inverse and Jacobian determinant computations. 
	NICE~\cite{dinh2014nice}, RealNVP~\cite{dinh2016density}, IAF~\cite{kingma2016improved}, and MAF~\cite{papamakarios2017masked} use either autoregressive or coupling flows where the Jacobian is triangular, so the Jacobian determinant can be tractably computed. GLOW~\cite{kingma2018glow} expands upon RealNVP by introducing an additional invertible convolution step. 
	These flows are based on either coupling layers or autoregressive transformations, whose tractable invertibility allows for density evaluation and generative sampling.
	Neural Spline Flows~\cite{durkan2019neural} use splines instead of the coupling layers used in GLOW and RealNVP. Using monotonic neural networks, NAF~\cite{huang2018neural} require positivity of the weights. UMNN~\cite{wehenkel2019unconstrained} circumvent this requirement by parameterizing the Jacobian and then integrating numerically.

	\paragraph{Infinitesimal Flows} 
	Modeling flows with differential equations is a natural and common concept~\cite{suykens1998,welling2011bayesian,neal2011mcmc,salimans2015markov}.
	In particular, CNFs~\cite{chen2017continuous,chen2018neural,grathwohl2019ffjord} model their flow via~\eqref{eq:neural_odes}.

	To alleviate the expensive training costs of CNFs, FFJORD~\cite{grathwohl2019ffjord} sacrifices the exact but slow trace computation in~\eqref{eq:neural_odes} for a Hutchinson's trace estimator with complexity $\bigO(d)$~\cite{hutchinson1990stochastic}. 
	This estimator helps FFJORD achieve training tractability by reducing the trace cost from $\bigO(d^2)$ to $\bigO(d)$ per time step.
	However, during inference, FFJORD has $\bigO(d^2)$ trace computation cost since accurate CNF inference requires the exact trace (Sec.~\ref{sec:intro},Tab.~\ref{tab:comparison}).
	FFJORD also uses the optimize-then-discretize (OTD) approach and an adjoint-based backpropagation where the intermediate gradients are recomputed.
	In contrast, our exact trace computation is competitive with FFJORD's trace approach during training and faster during inference (Fig.~\ref{fig:trace_compare}). 
	\model{}'s use of DTO has been shown to converge quicker when training neural ODEs due to accurate gradient computation, storing intermediate gradients, and fewer time steps~\cite{li2017maximum,gholami2019anode,onken2020do} (Sec~\ref{sec:trace}).

	\paragraph{Flows Influenced by Optimal Transport}
 	To encourage straight trajectories, RNODE~\cite{finlay2020train} regularizes FFJORD with a transport cost $L(\bfx,T)$. 
 	RNODE also includes the Frobenius norm of the Jacobian $\| \nabla \bfv \|_F^2$ to stabilize training. 
	They estimate the trace and the Frobenius norm using a stochastic estimator and report 2.8x speedup. 
	Numerically, RNODE, FFJORD, and \model{} differ. Specifically, \model{}'s exact trace allows for stable training without $\| \nabla \bfv \|_F^2$ (Fig.~\ref{fig:hutch}).
	In formulation, \model{} shares the $L_2$ cost with RNODE but follows a potential flow approach (Tab.~\ref{tab:comparison}).

	Monge-Ampère Flows~\cite{zhang2018monge} and Potential Flow Generators~\cite{yang2019} similarly draw from OT theory but parameterize a potential function (Tab.~\ref{tab:comparison}).
	However, \model{}'s numerics differ substantially due to our scalable exact trace computation (Tab.~\ref{tab:comparison}).
	OT is also used in other generative models~\cite{sanjabi2018,salimans2018improving,lei2018geometric,lin2019fluid,avraham2019parallel,tanaka2019discriminator}.

\section{Numerical Experiments} \label{sec:experiments}

	We perform density estimation on seven two-dimensional toy problems and five high-dimensional problems from real data sets. We also show \model{}'s generative abilities on MNIST.

\begin{figure*}
    \centering
    \addtolength{\tabcolsep}{-5pt} 
    \subfloat[\miniboone{} dimension 16 vs 17\label{fig:miniboone1}]{
	   	\begin{tabular}{ccc}
	   	\small
	   	Samples & \model{} & FFJORD\hspace{16pt} \\
	   	$\bfx \sim \rho_0(\bfx)$ & $f(\bfx)$ & $f(\bfx)$\hspace{16pt} \\
	   	\includegraphics[height=65pt]{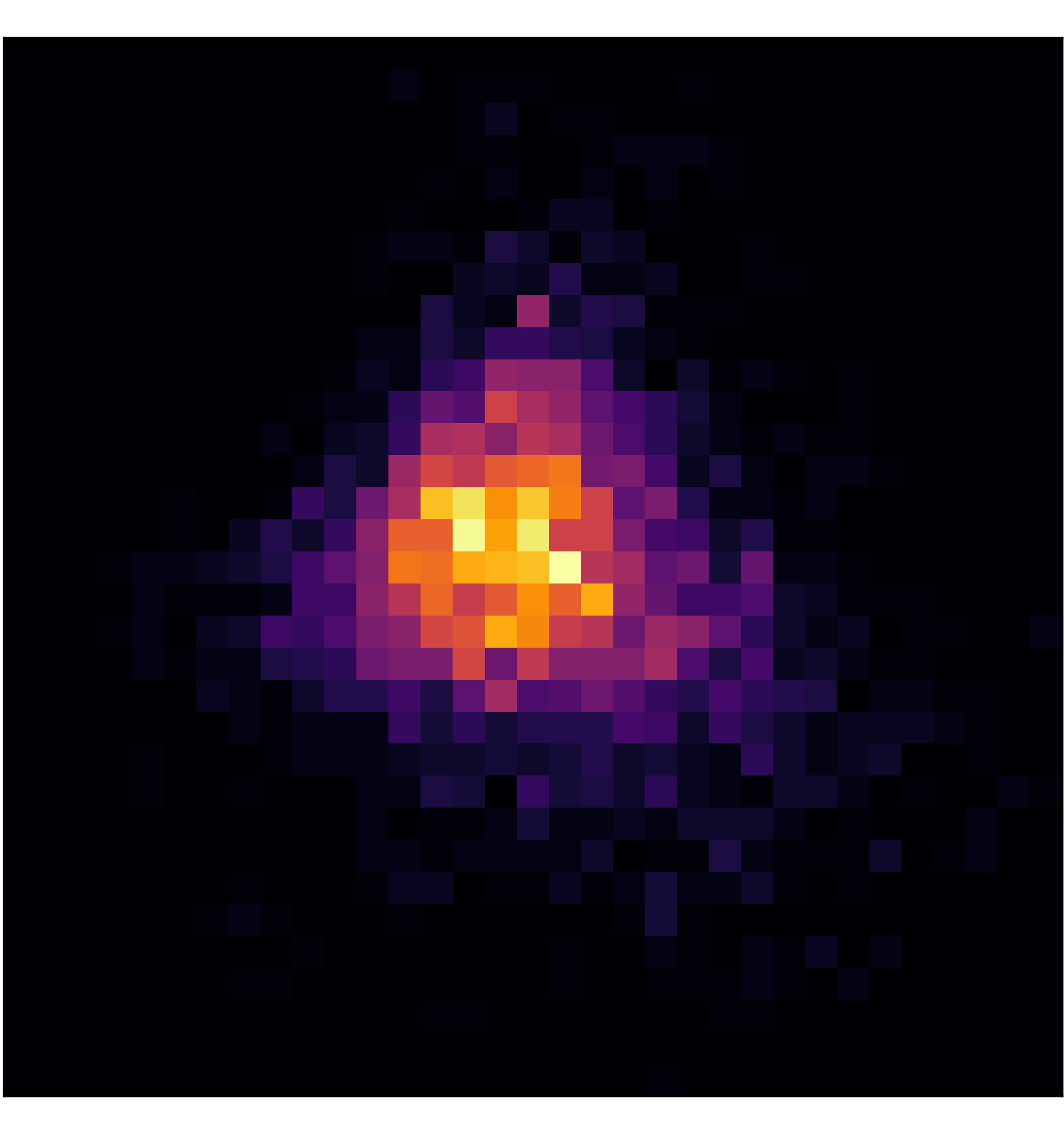}
	   	&
	   	\includegraphics[height=65pt]{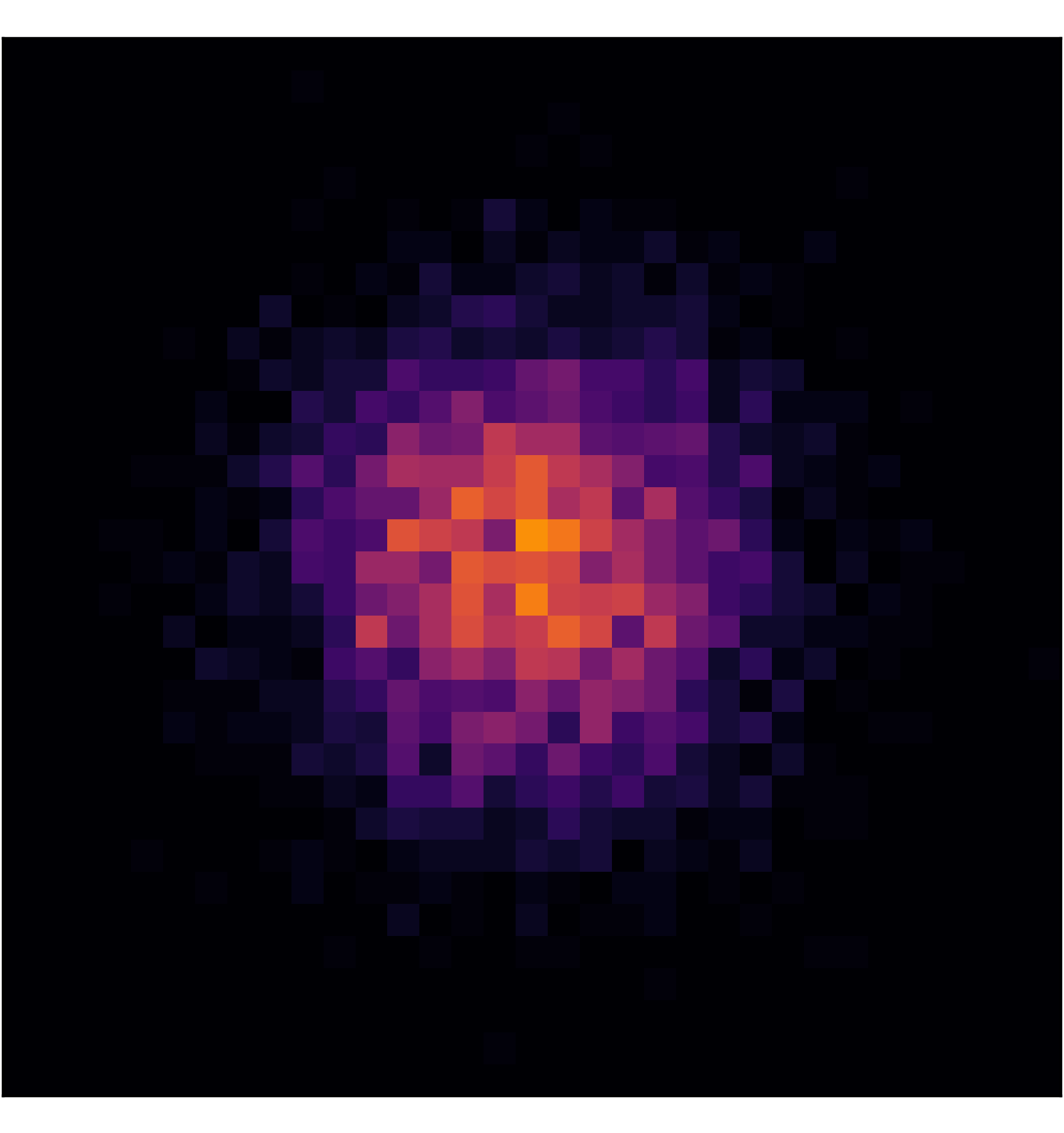}
	   	&
	   	\includegraphics[height=65pt]{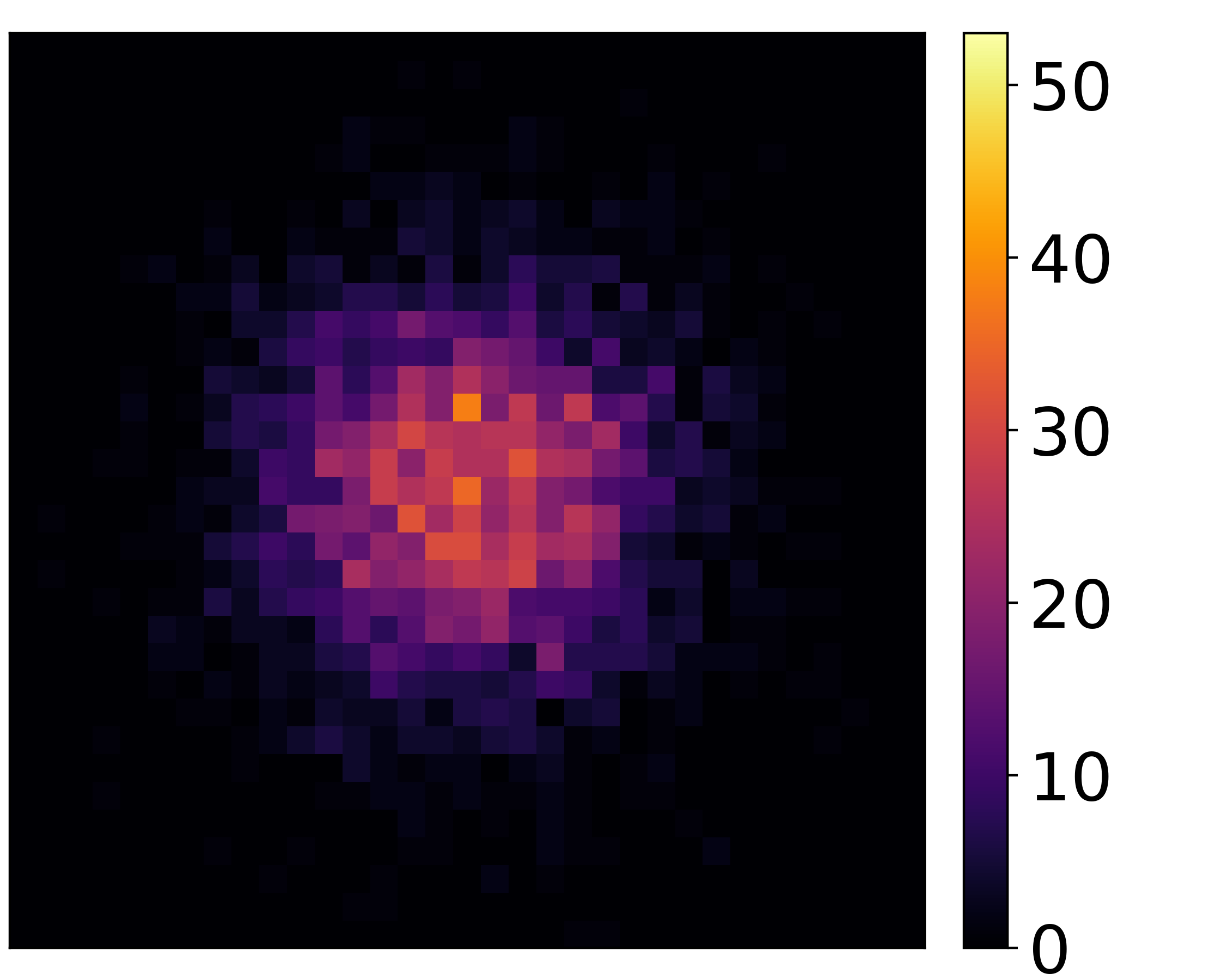} \\
	   	$\bfy \sim \rho_1(\bfy)$ & $f^{-1}(\bfy)$ & $f^{-1}(\bfy)$\hspace{16pt} \\ 
	   	\includegraphics[height=65pt]{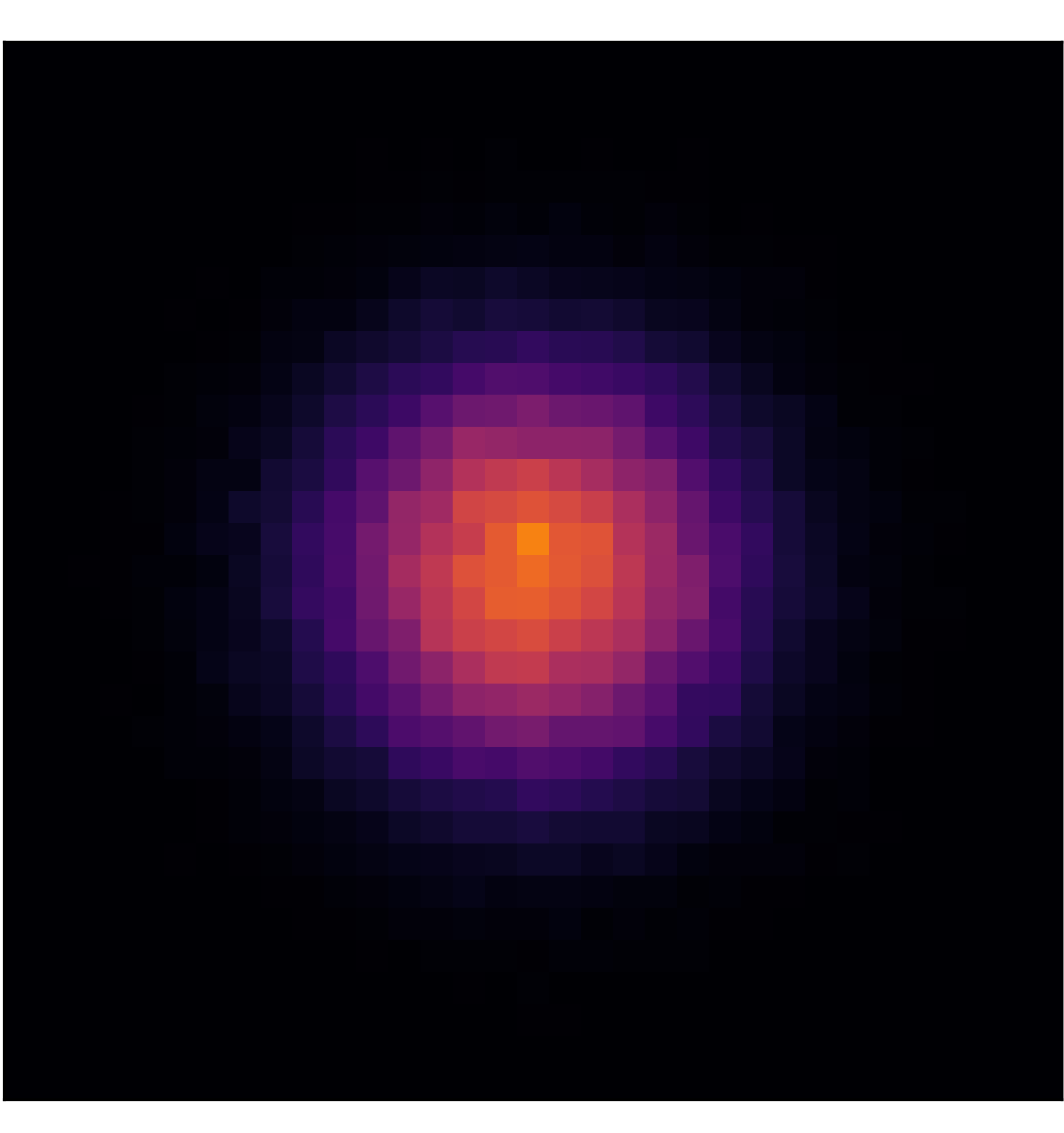}
	   	&
	   	\includegraphics[height=65pt]{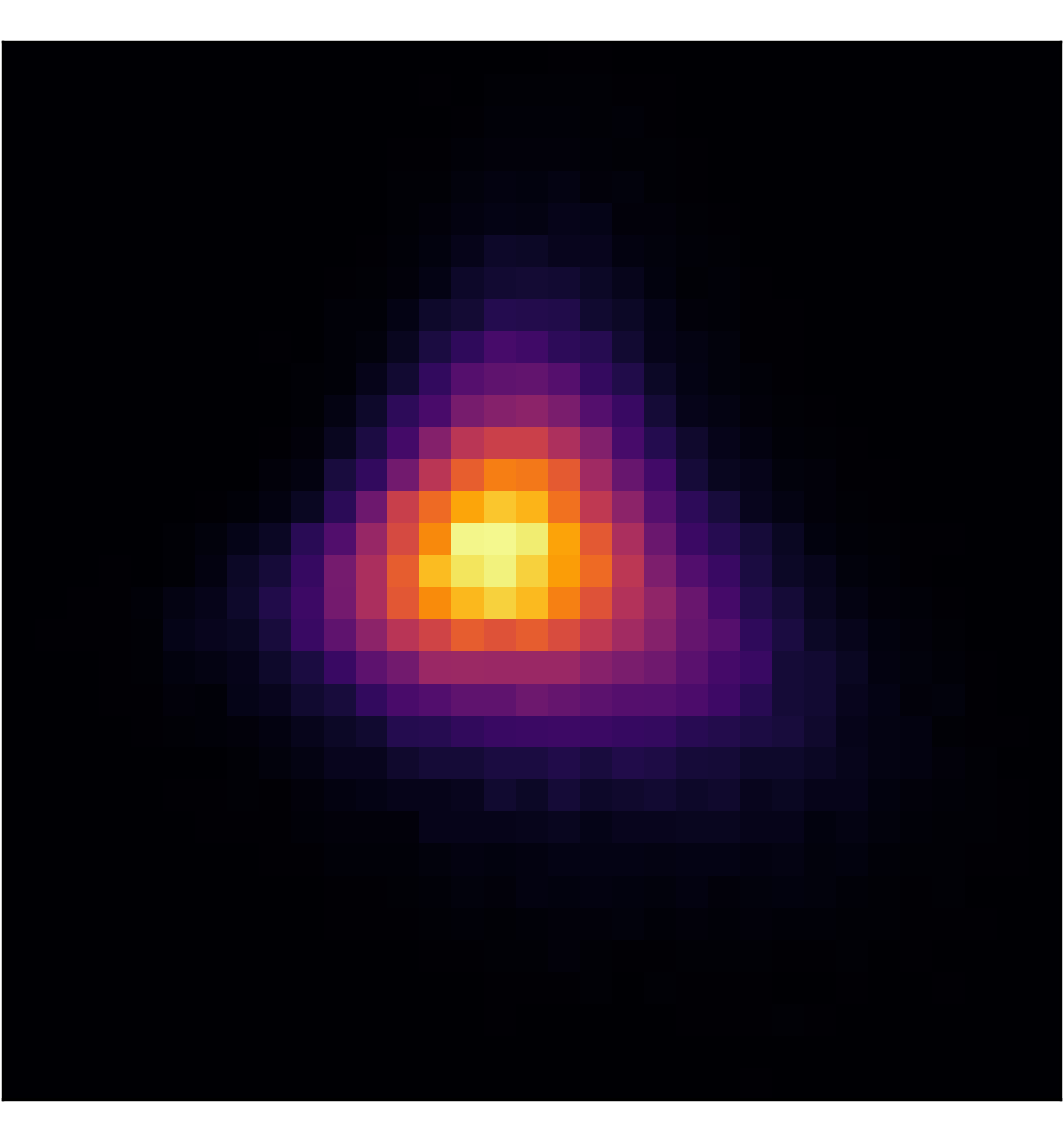}
	   	&
	   	\includegraphics[height=65pt]{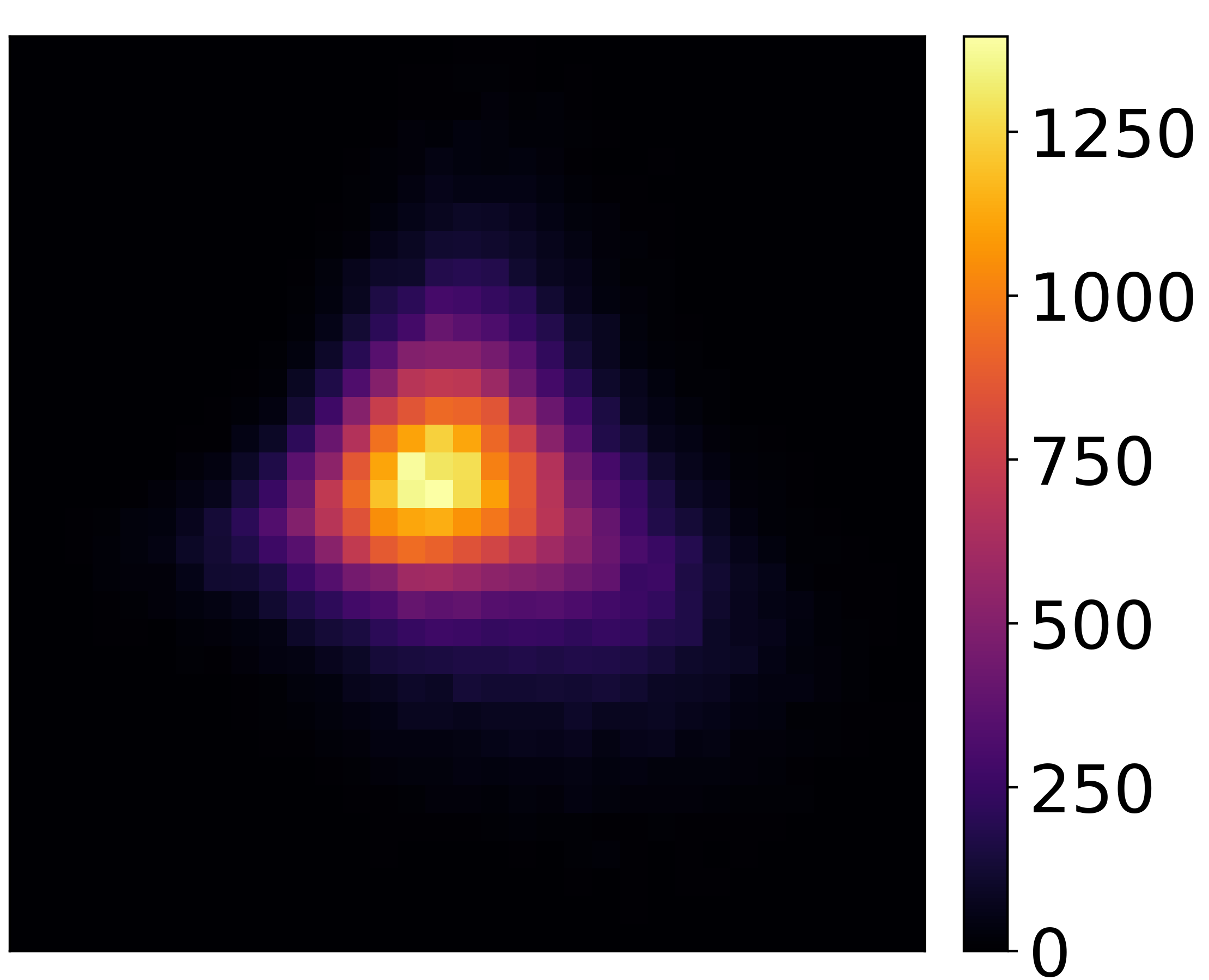}
	   	\end{tabular}%
	}\hspace{20pt}%
   	\subfloat[\miniboone{} dimension 28 vs 29\label{fig:miniboone2}]{
	   	\begin{tabular}{ccc}
	   	Samples & \model{} & FFJORD\hspace{16pt} \\
	   	$\bfx \sim \rho_0(\bfx)$ & $f(\bfx)$ & $f(\bfx)$\hspace{16pt} \\
	   	\includegraphics[height=65pt]{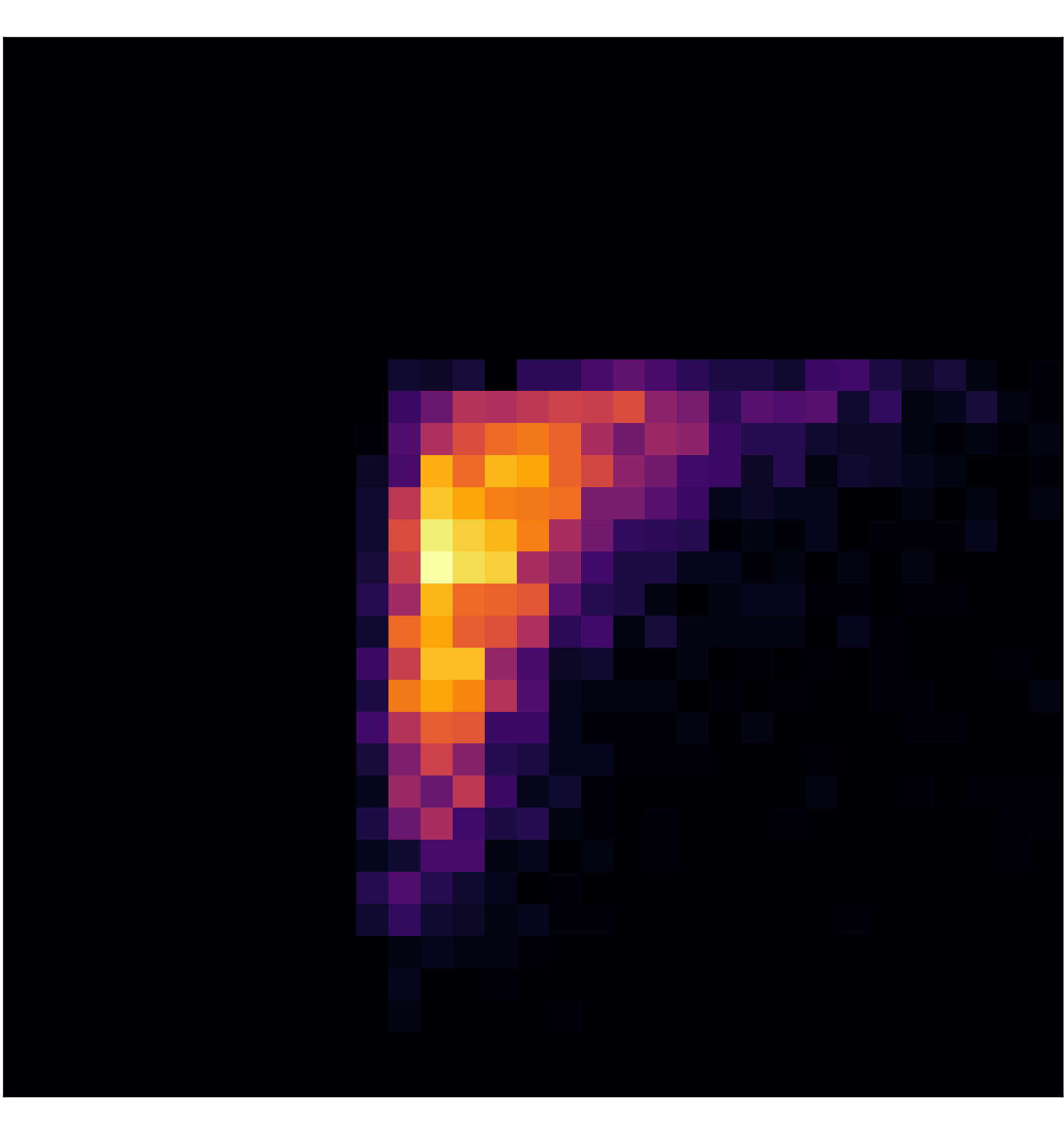}
	   	&
	   	\includegraphics[height=65pt]{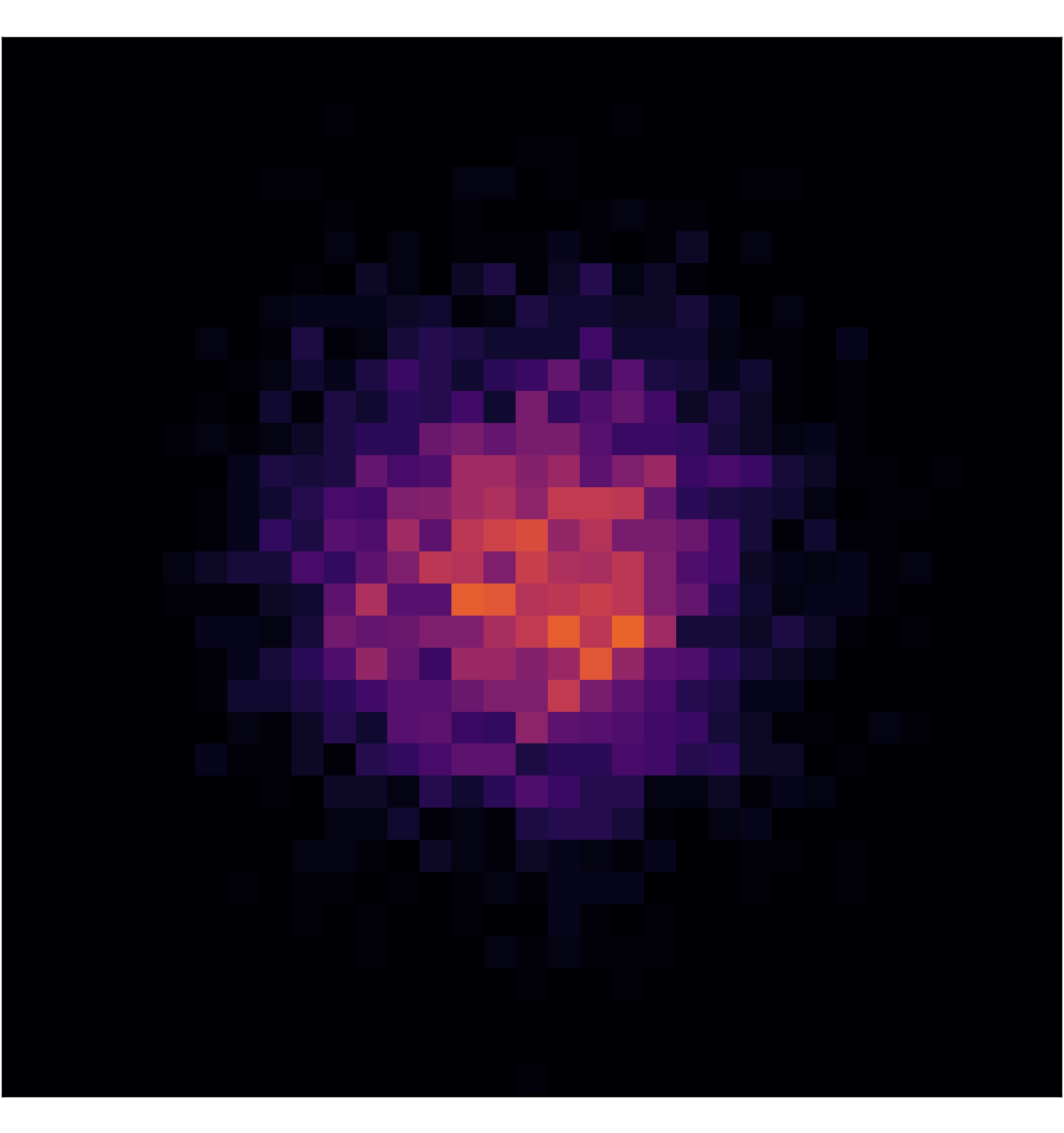}
	   	&
	   	\includegraphics[height=65pt]{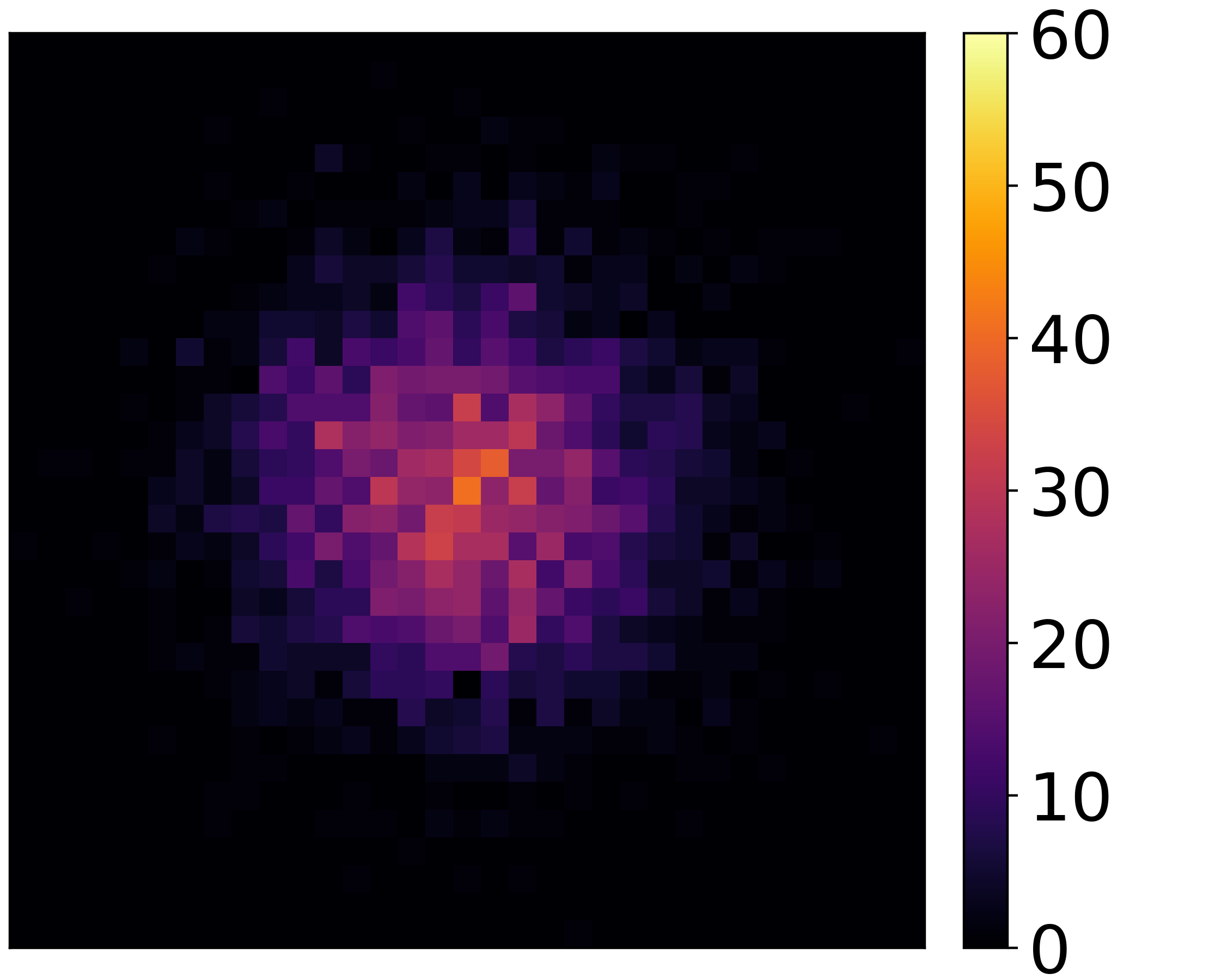} \\
	   	$\bfy \sim \rho_1(\bfy)$ & $f^{-1}(\bfy)$ & $f^{-1}(\bfy)$\hspace{16pt} \\ 
	   	\includegraphics[height=65pt]{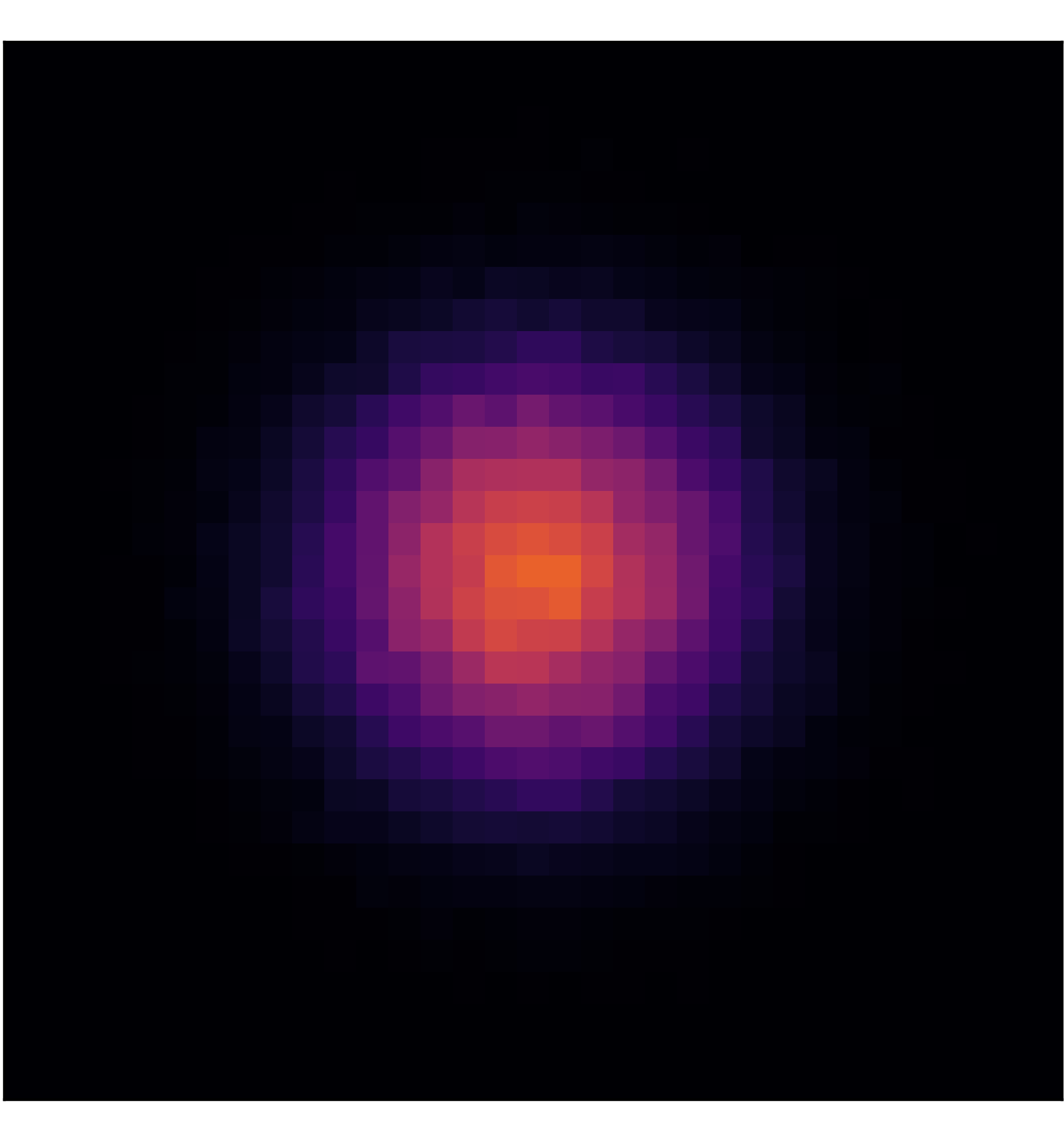}
	   	&
	   	\includegraphics[height=65pt]{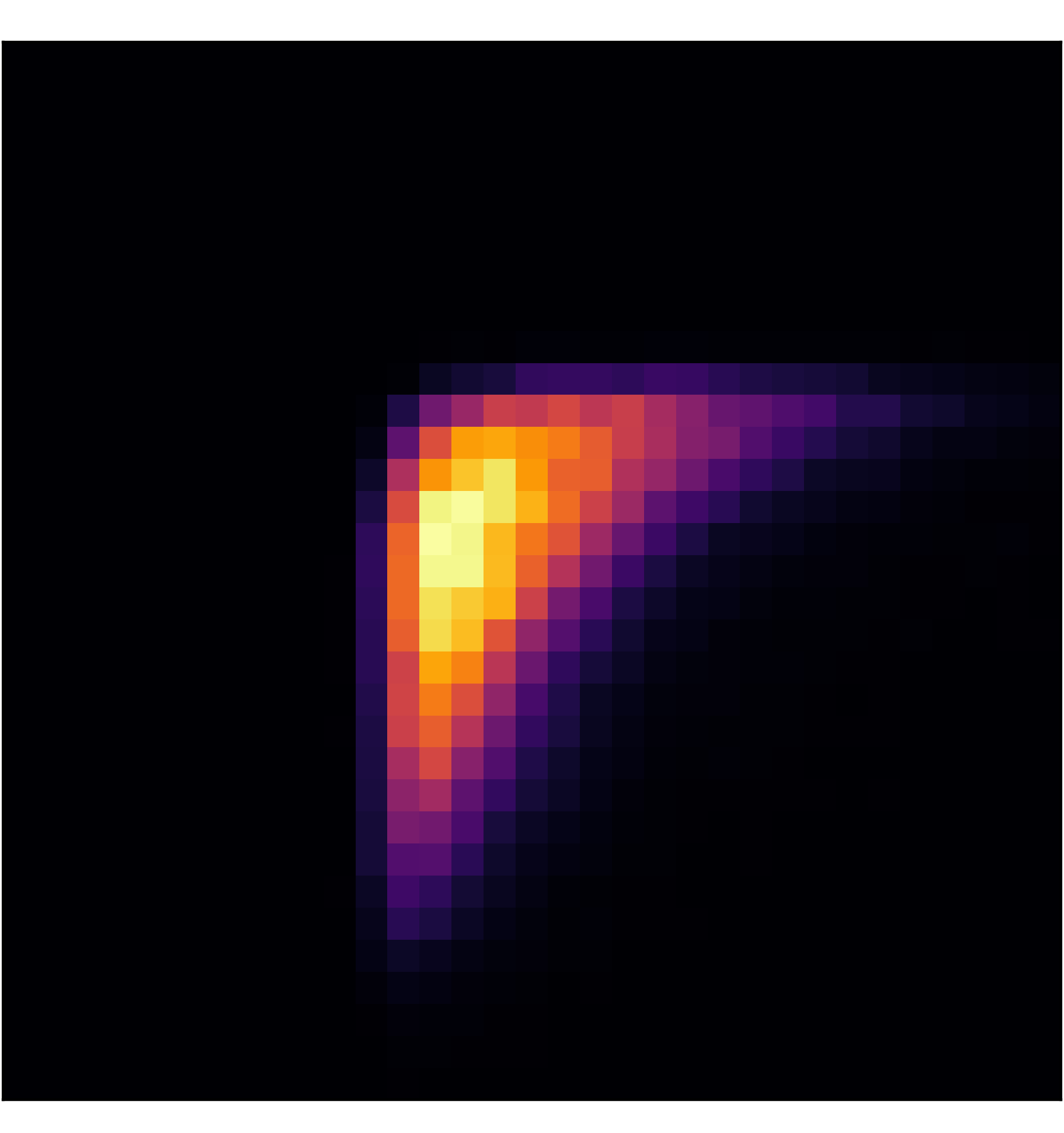}
	   	&
	   	\includegraphics[height=65pt]{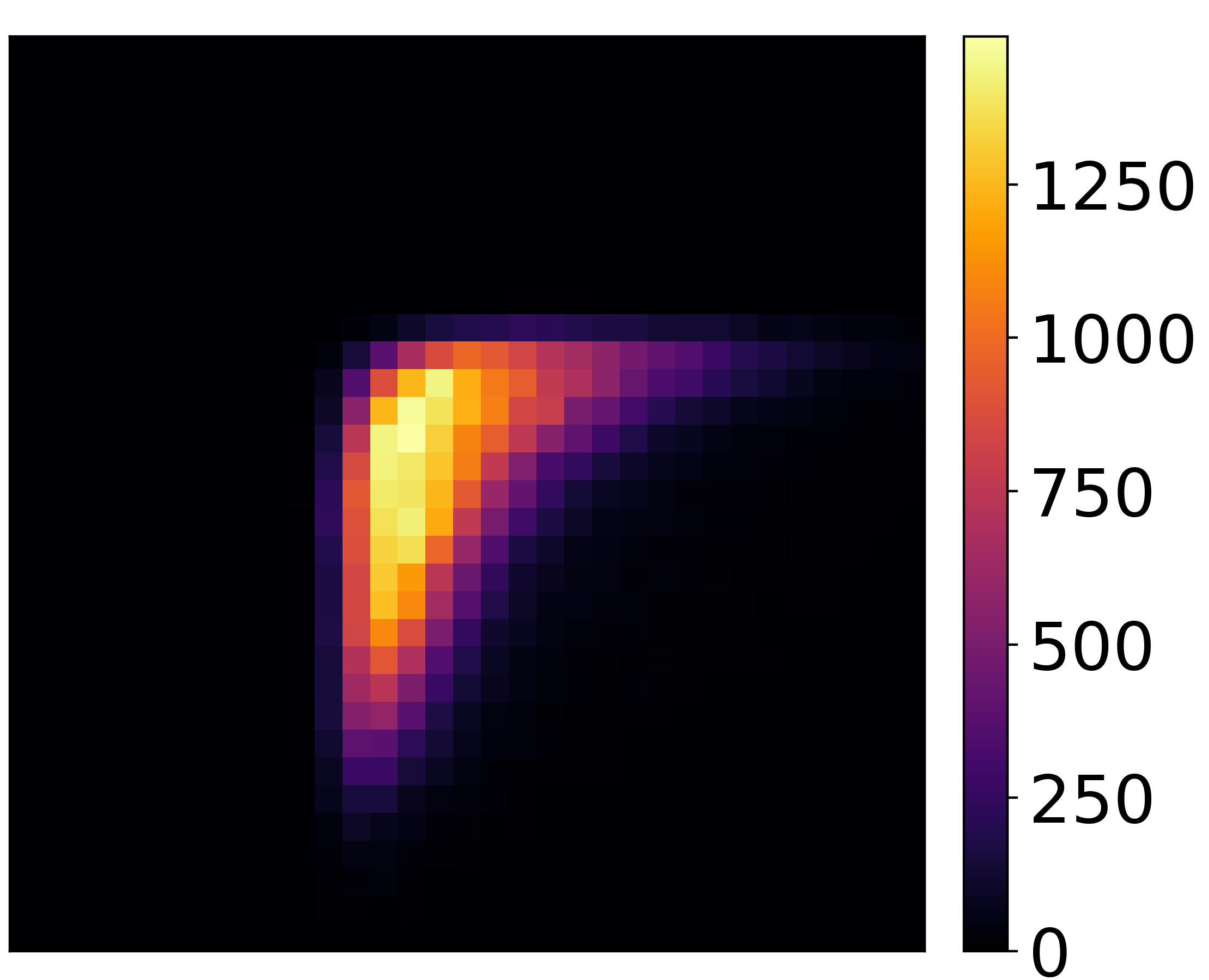}
	   	\end{tabular}%
	}%
    \caption{\miniboone{} density estimation.
    Two-dimensional slices using the 3,648 43-dimensional testing samples $\bfx \sim \rho_0(\bfx)$ and $10^5$ samples $\bfy$ from distribution $\rho_1$ (more visuals in Fig.~\ref{fig:miniboone_full}). 
    }
    \label{fig:miniboone}
\addtolength{\tabcolsep}{5pt} 
\end{figure*}

\paragraph{Metrics}

	In density estimation, the goal is to approximate $\rho_0$ using observed samples $\bfX=\{ \bfx_i \}_{i=1}^N$, where $\bfx_i$ are drawn from the distribution $\rho_0$.
	In real applications, we lack a ground-truth $\rho_0$, rendering proper evaluation of the density itself untenable. However, we can follow evaluation techniques applied to generative models. Drawing random points $\{ \bfy_i \}_{i=1}^M$ from $\rho_1$, we invert the flow to generate synthetic samples $\bfQ=\{ \bfq_i \}_{i=1}^M$, where $\bfq_i = f^{-1}(\bfy_i)$. We compare the known samples to the generated samples via maximum mean discrepancy (MMD)~\cite{gretton2012mmd,li2015generative,theis2016mmd,peyre2018computational}%
	\begin{equation}%
	\begin{split}%
	  	{\rm MMD}(\bfX,\bfQ) = \frac{1}{N^2} \sum_{i=1}^N \sum_{j=1}^N k(\bfx_i,\bfx_j) \\
				   + \frac{1}{M^2} \sum_{i=1}^M \sum_{j=1}^M k(\bfq_i,\bfq_j)
				   - \frac{2}{NM}  \sum_{i=1}^N \sum_{j=1}^M k(\bfx_i,\bfq_j) ,
	\end{split}%
	\end{equation}%
	for Gaussian kernel $k(\bfx_i,\bfq_j)= \exp( - \frac{1}{2} \| \bfx_i - \bfq_j \|^2)$. MMD tests the difference between two distributions ($\rho_0$ and our estimate of $\rho_0$) on the basis of samples drawn from each ($\bfX$ and $\bfQ$). A low MMD value means that the two sets of samples are likely to have been drawn from the same distribution~\cite{gretton2012mmd}. Since MMD is not used in the training, it provides an external, impartial metric to evaluate our model on the hold-out test set (Tab.~\ref{tab:large}).
	
	Many normalizing flows use $C$ for evaluation. The loss $C$ is used to train the forward flow to match $\rho_1$. Testing loss, i.e., $C$ evaluated on the testing set, should provide the same quantification on a hold-out set.
	However, in some cases, the testing loss can be low even when $f(\bfx)$ is poor and differs substantially from $\rho_1$ (Fig.~\ref{fig:8gauss},~\ref{fig:bad_power}).
	Furthermore, because the model's inverse contains error, accurately mapping to $\rho_1$ with the forward flow does not necessarily mean the inverse flow accurately maps to $\rho_0$.

	Testing loss varies drastically with the integration computation~\cite{theis2016mmd,wehenkel2019unconstrained,onken2020do}.
	It depends on $\ell$, which is computed along the characteristics via time integration of the trace (App.~\ref{app:test_loss}).
	Too few discretization points leads to an inaccurate integration computation and greater inverse error. Thus, a low inverse error implies an accurate integration computation because the flow closely models the ODE. An adaptive ODE solver alleviates this concern when provided a sufficiently small tolerance~\cite{grathwohl2019ffjord}. Similarly, we check that the flow models the continuous solution of the ODE by computing the inverse error
	\begin{equation}
		\label{eq:inverr}
	    \E_{\rho_0(\bfx)}\lVert f^{-1} \left( f(\bfx) \right) - \bfx \rVert_{2}	
	\end{equation}	
	on the testing set using a finer time discretization than used in training.
	We evaluate the expectation values in~\eqref{eq:MFGFlowObj} and~\eqref{eq:inverr} using the discrete samples $\bfX$, which we assume are randomly drawn from and representative of the initial distribution $\rho_0$.
	
\paragraph{Toy Problems}

	We train \model{} on several toy distributions that serve as standard benchmarks~\cite{grathwohl2019ffjord,wehenkel2019unconstrained}. Given random samples, we train \model{} then use it to estimate the density $\rho_0$ and generate samples (Fig.~\ref{fig:toyModels}).
	We present a thorough comparison with a state-of-the-art CNF on one of these (Fig.~\ref{fig:8gauss}) .

\paragraph{Density Estimation on Real Data Sets}

	We compare our model's performance on real data sets (\power{}, \gas{}, \hepmass{}, \miniboone{}) from the University of California Irvine (UCI) machine learning data repository and the \bsds{} data set containing natural image patches. The UCI data sets describe observations from Fermilab neutrino experiments, household power consumption, chemical sensors of ethylene and carbon monoxide gas mixtures, and particle collisions in high energy physics. Prepared by \citet{papamakarios2017masked}, the data sets 
	are commonly used in normalizing flows~\cite{dinh2016density,grathwohl2019ffjord,huang2018neural,wehenkel2019unconstrained}.
	The data sets vary in dimensionality (Tab.~\ref{tab:large}).
	
	For each data set, we compare \model{} with FFJORD~\cite{grathwohl2019ffjord} and RNODE~\cite{finlay2020train} (current state-of-the-art) in speed and performance.
	We compare speed both in training the models and when running the model on the testing set. To compare performance, we compute the MMD between the data set and $M{=}10^5$ generated samples $f^{-1}(\bfy)$ for each model; for a fair comparison, we use the same $\bfy$ for FFJORD and \model{} (Tab.~\ref{tab:large}). 
	We show visuals of the samples $\bfx\sim\rho_0(\bfx)$, $\bfy\sim\rho_1(\bfy)$, $f(\bfx)$, and $f^{-1}(\bfy)$ generated by \model{} and FFJORD  (Fig.~\ref{fig:miniboone}, App.~\ref{app:extra_figs}). We report the loss $C$ values (Tab.~\ref{tab:large}) to be comparable to other literature but reiterate the inherent flaws in using $C$ to compare models.
	
	The results demonstrate the computational efficiency of \model{} relative to the state-of-the-art (Tab.~\ref{tab:large}). With the exception of the \gas{} data set, \model{} achieves comparable MMD to the state-of-the-art with drastically reduced training time. 
	\model{} learns a slightly smoothed representation of the \gas{} data set (Fig.~\ref{fig:gas}).
	We attribute most of the training speedup to the efficiency from using our exact trace instead of the Hutchinson's trace estimation (Fig.~\ref{fig:trace_compare}, Fig.~\ref{fig:hutch}). 
	On the testing set, our exact trace leads to faster testing time than the state-of-the-art's exact trace computation via AD (Tab.~\ref{tab:comparison}, Tab.~\ref{tab:large}).
	To evaluate the testing data, we use more time steps than for training, effectively re-discretizing the ODE at different points. The inverse error shows that \model{} is numerically invertible and suggests that it approximates the true solution of the ODE.
    Ultimately, \model{}'s combination of OT-influenced regularization, reduced parameterization, DTO approach, and efficient exact trace computation results in fast and accurate training and testing.

\paragraph{MNIST}

	We demonstrate the generation quality of \model{} using an encoder-decoder structure.
	Consider encoder $B \colon \R^{784} \to \R^d$ and decoder $D \colon \R^d \to \R^{784}$ such that $D(B(\bfx)) \approx \bfx$.
	We train $d$-dimensional flows that map distribution $\rho_0(B(\bfx))$ to $\rho_1$. The encoder and decoder each use a single dense layer and activation function (ReLU for $B$ and sigmoid for $D$). We train the encoder-decoder separate from and prior to training the flows. The trained encoder-decoder, due to its simplicity, renders digits $D(B(\bfx))$ that are a couple pixels thicker than the supplied digit $\bfx$.  
	
\begin{figure}
    \centering
    \subfloat[Originals\label{fig:mnistOrig}]{\includegraphics[width=0.8\columnwidth]{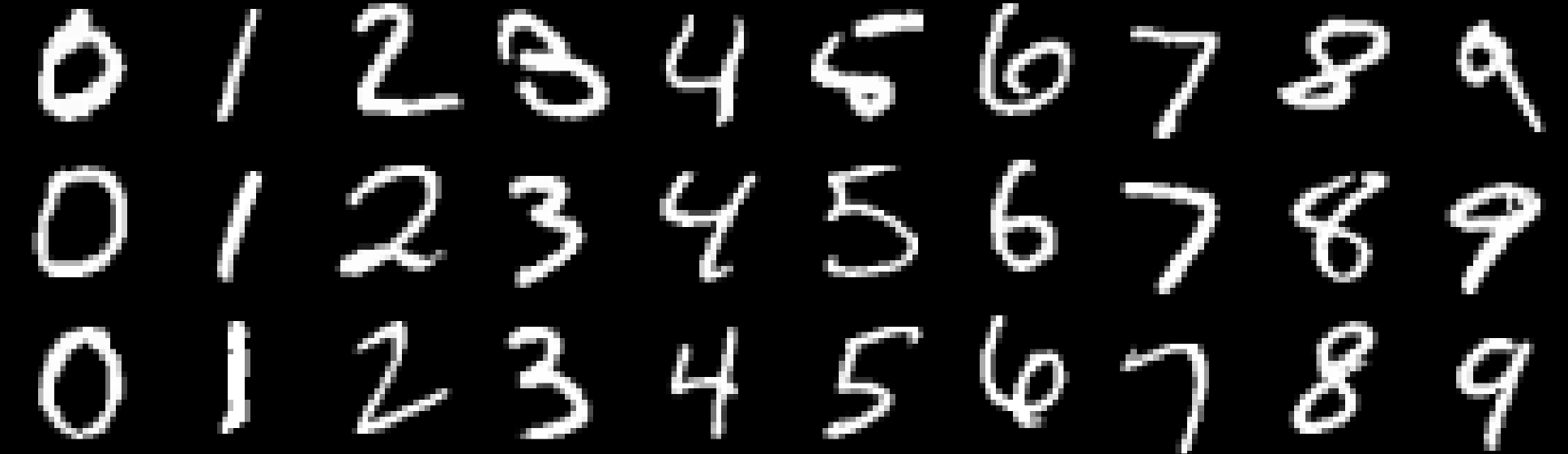}} \\
    \subfloat[Generations\label{fig:mnistGen}]{\includegraphics[width=0.8\columnwidth]{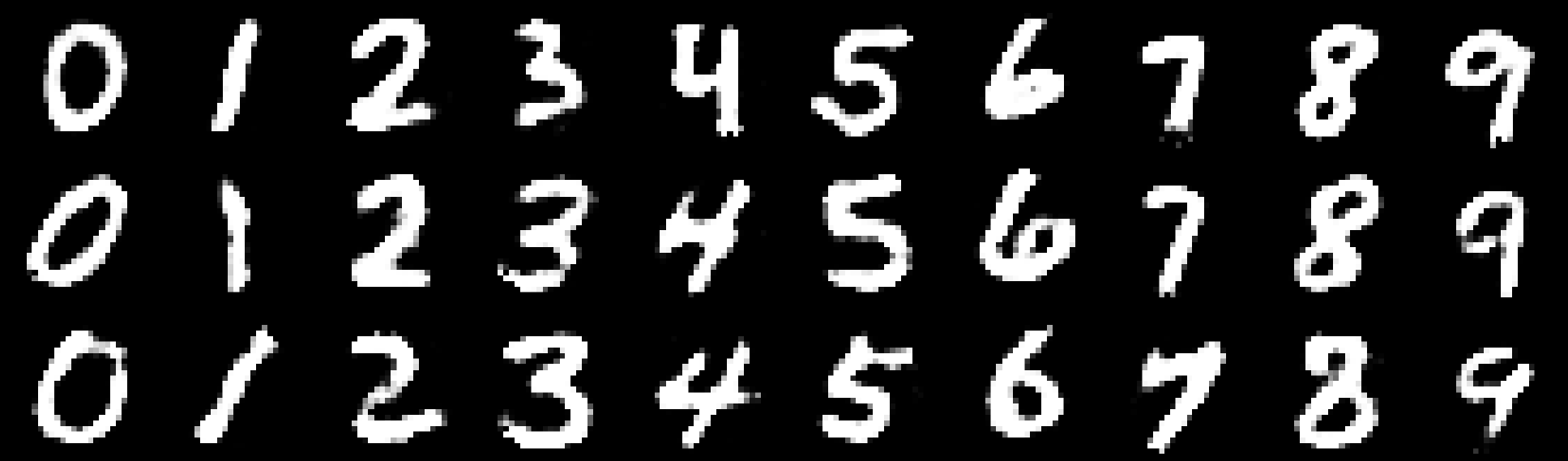}}
  \caption{MNIST generation conditioned by class.}
\end{figure}

\begin{figure}
    \centering
    \includegraphics[width=0.57\linewidth]{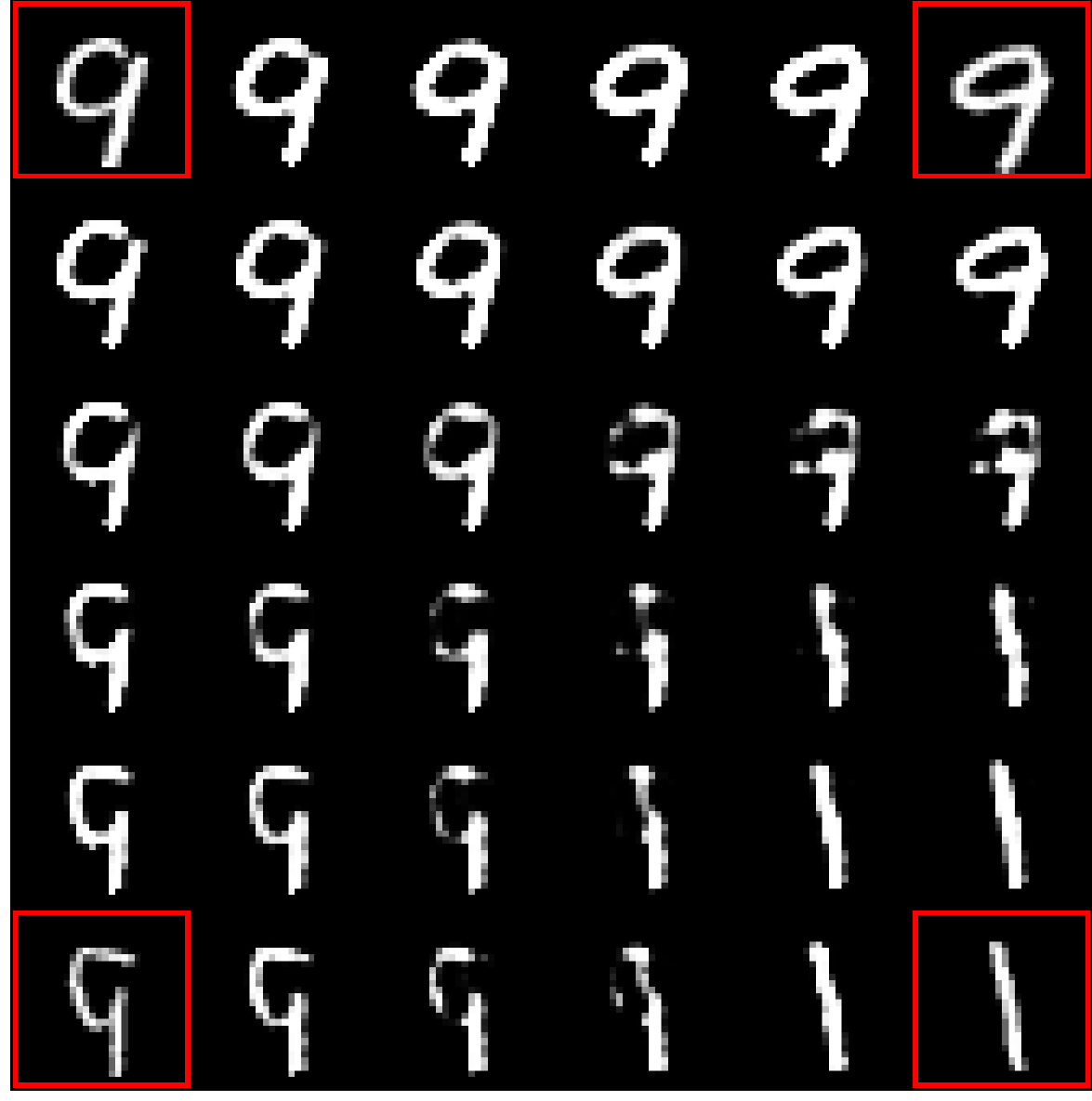}
  \caption{MNIST interpolation in the latent space. 
  Original images are boxed in red. }
  \label{fig:mnistVAE}
\end{figure}

	We generate new images via two methods. First, using $d{=}64$ and a flow conditioned on class, we sample a point $\bfy \sim \rho_1(\bfy)$ and map it back to the pixel space to create image $D(f^{-1}(\bfy))$ (Fig.~\ref{fig:mnistGen}). Second, using $d{=}128$ and an unconditioned flow, we interpolate between the latent representations $f(B(\bfx_1)),f(B(\bfx_2))$ of original images $\bfx_1,\bfx_2$. For interpolated latent vector $\bfy \in \R^d$, we invert the flow and decode back to the pixel space to create image $D(f^{-1}(\bfy))$ (Fig.~\ref{fig:mnistVAE}).

\section{Discussion}

    We present \model{}, a fast and accurate approach for training and performing inference with CNFs. 
    Our approach tackles two critical computational challenges in CNFs. 
    
    First, solving the neural ODEs in CNFs can require many time steps resulting in high computational cost.
    Leveraging OT theory, we include a transport cost and add an HJB regularizer.
    These additions help carry properties from the continuous problem to the discrete problem and allow \model{} to use few time steps without sacrificing performance.
	Second, computing the trace term in~\eqref{eq:neural_odes} is computationally expensive.
	OT-Flow features exact trace computation at time complexity and cost comparable to trace estimators used in existing state-of-the-art CNFs (Fig.~\ref{fig:trace_compare}). The exact trace provides better convergence than the estimator (Fig.~\ref{fig:hutch}). 
	Our analytic gradient and trace approach is not limited to the ResNet architectures, but expanding to other architectures requires further derivation.

\section*{Acknowledgments}
 	This research was supported by the NSF award DMS 1751636, Binational Science Foundation Grant 2018209, AFOSR Grants 20RT0237 and FA9550-18-1-0167, AFOSR MURI FA9550-18-1-0502, ONR Grant No. N00014-18-1-2527, a gift from UnitedHealth Group R\&D, and a GPU donation by NVIDIA Corporation. 
 	Important parts of this research were performed while LR was visiting the Institute for Pure and Applied Mathematics (IPAM), which is supported by the NSF Grant No. DMS 1440415.

\bibliography{main}

\clearpage

\onecolumn

\section*{\LARGE{Appendix}}
\appendix

 \setcounter{figure}{0} 
 \setcounter{table}{0} 
 \renewcommand{\thefigure}{A\arabic{figure}}
 \renewcommand{\thetable}{A\arabic{table}}

\section{Derivation of Loss $C$} 
\label{app:G}

	Let $\rho_0$ be the initial density of samples, and $\bfz$ be the trajectories that map samples from $\rho_0$ to $\rho_1$. The change in density as we flow a sample $\bfx \sim \rho_0$ at time $t$ is given by the change of variables formula
	\begin{equation}
	  \label{eq:jacobi}
	  \rho_0(\bfx) = \rho \big( \bfz(\bfx, t) \big) \det \big( \nabla \bfz(\bfx, t) \big),
	\end{equation}
	where $\bfz(\bfx,0) = \bfx$. In normalizing flows, the discrepancy between the flowed distribution at final time $T$, denoted $\rho(\bfx,T)$, and the normal distribution can be measured using the Kullback-Leibler (KL) divergence%
	\begin{equation}
	  \label{eq:appKL}
	    \D_{\rm KL} \big[ \, \rho(\bfx,T) \, || \, \rho_1(\bfx) \, \big] 
	    = \int_{\R^d} \log\left(\frac{\rho(\bfx,T)}{\rho_1(\bfx)}\right) \rho(\bfx,T) \, \du\bfx.
	\end{equation}
	
	Changing variables, and using~\eqref{eq:jacobi}, we can rewrite~\eqref{eq:appKL} as
	\begin{equation}
	  \label{eq:appKL_inZ}
	  \begin{split}
	  &\D_{\rm KL} \big[ \, \rho(\bfz(\bfx,T)) \, || \, \rho_1(\bfz(\bfx,T)) \, \big] 
	  \\
	  &= \int_{\R^d} \log\left(\frac{\rho(\bfz(\bfx,T))}{\rho_1(\bfz(\bfx,T))}\right) \rho(\bfz(\bfx,T)) \det\big(\nabla \bfz(\bfx, T)\big) \, \du \bfx,
	  \\
	  &= \int_{\R^d} \log\left(\frac{\rho_0(\bfx)}{\rho_1 \big(\bfz(\bfx,T)\big) \, \det \big(\nabla \bfz(\bfx,T)\big)}\right) \rho_0(\bfx)
	  \, \du \bfx,
	  \\
	  &= \int_{\R^d} \Big[ \log\big(\rho_0(\bfx)\big) - \log\big(\rho_1(\bfz(\bfx,T))\big) - \log\det\big(\nabla \bfz(\bfx,T)\big) \Big] \rho_0(\bfx) \, \du \bfx.
	  \end{split}
	\end{equation}

	For normalizing flows, we assume $\rho_1$ is the standard normal
	\begin{equation}
	  \rho_1(\bfx) = \dfrac{1}{\sqrt{(2\pi)^d}}  \exp{\left(\frac{-\| \bfx \|^2}{2} \right)},
	\end{equation}	
	which will reduce the term
	\begin{equation}
	  \label{eq:logRho1_inZ}
	  \log\big(\rho_1(\bfz(\bfx,T))\big) = - \frac{1}{2} \| \bfz(\bfx,T) \|^2 - \frac{d}{2}\log(2\pi).
	\end{equation}
	
	Substituting~\eqref{eq:logRho1_inZ} into~\eqref{eq:appKL_inZ}, we obtain 
	\begin{equation}
	  \begin{split}
	    \D_{\rm KL} &= \int_{\R^d} \Big[\log\big(\rho_0(\bfx)\big) - \log\det \big(\nabla \bfz(\bfx, T)\big) + \- \frac{1}{2} \| \bfz(\bfx,T) \|^2 + \frac{d}{2}\log(2\pi) \Big] \rho_0(\bfx) \, \du\bfx
	    \\
	    &= \int_{\R^d} \Big[\log\big(\rho_0(\bfx)\big) + C(\bfx,T)\Big] \rho_0(\bfx) \, \du \bfx
	    \\
	    &= \E_{\rho_0(\bfx)}  \left\{ \log\big(\rho_0(\bfx)\big) + C(\bfx,T) \right\},
	  \end{split}
	\end{equation}
	where $C(\bfx,T)$ is defined in~\eqref{eq:KLbasedLoss}.
	Density $\rho_0(\bfx)$ is unknown in normalizing flows. Thus, the term $\log(\rho_0(\bfx))$ is dropped, and normalizing flows minimize $C$ alone. Subtracting this constant does not affect the minimizer.

\newcommand{\rottext}[1]{\rotatebox{90}{\parbox{35mm}{\centering #1 }}}
\begin{figure}[t!]
  \centering
  \addtolength{\tabcolsep}{-3pt} 
  \begin{tabular}{cccc}
  & $\bfx \sim \rho_0(\bfx)$ & $f(\bfx) = \bfz(\bfx,T)$ & $f^{-1}(\bfy) \, , \, \bfy \sim \rho_1(\bfy)$
  \\
  \rottext{\textbf{No} HJB \\ 2 Time Steps}
  &
  \includegraphics[width=0.25\textwidth]{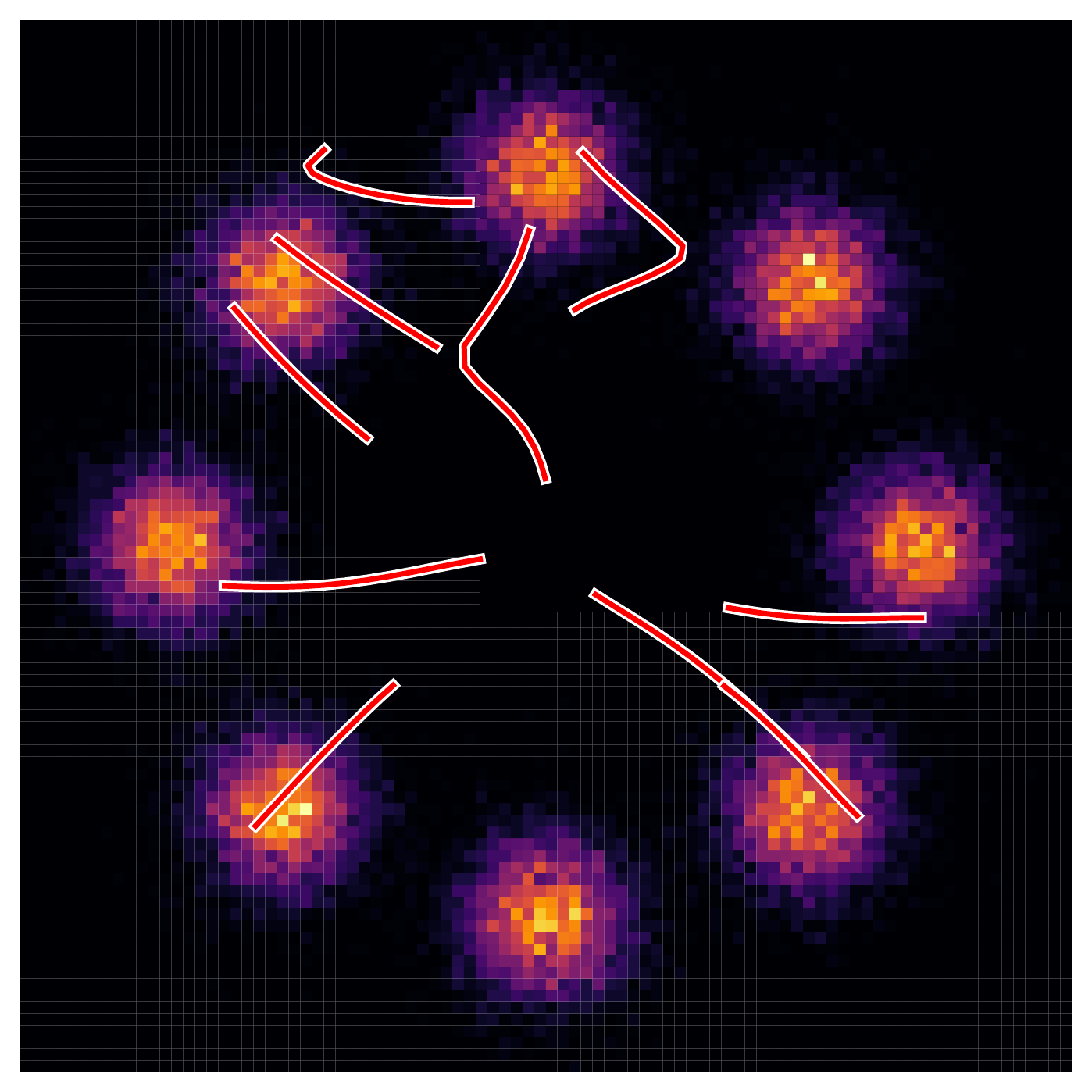}
  &
  \includegraphics[width=0.25\textwidth]{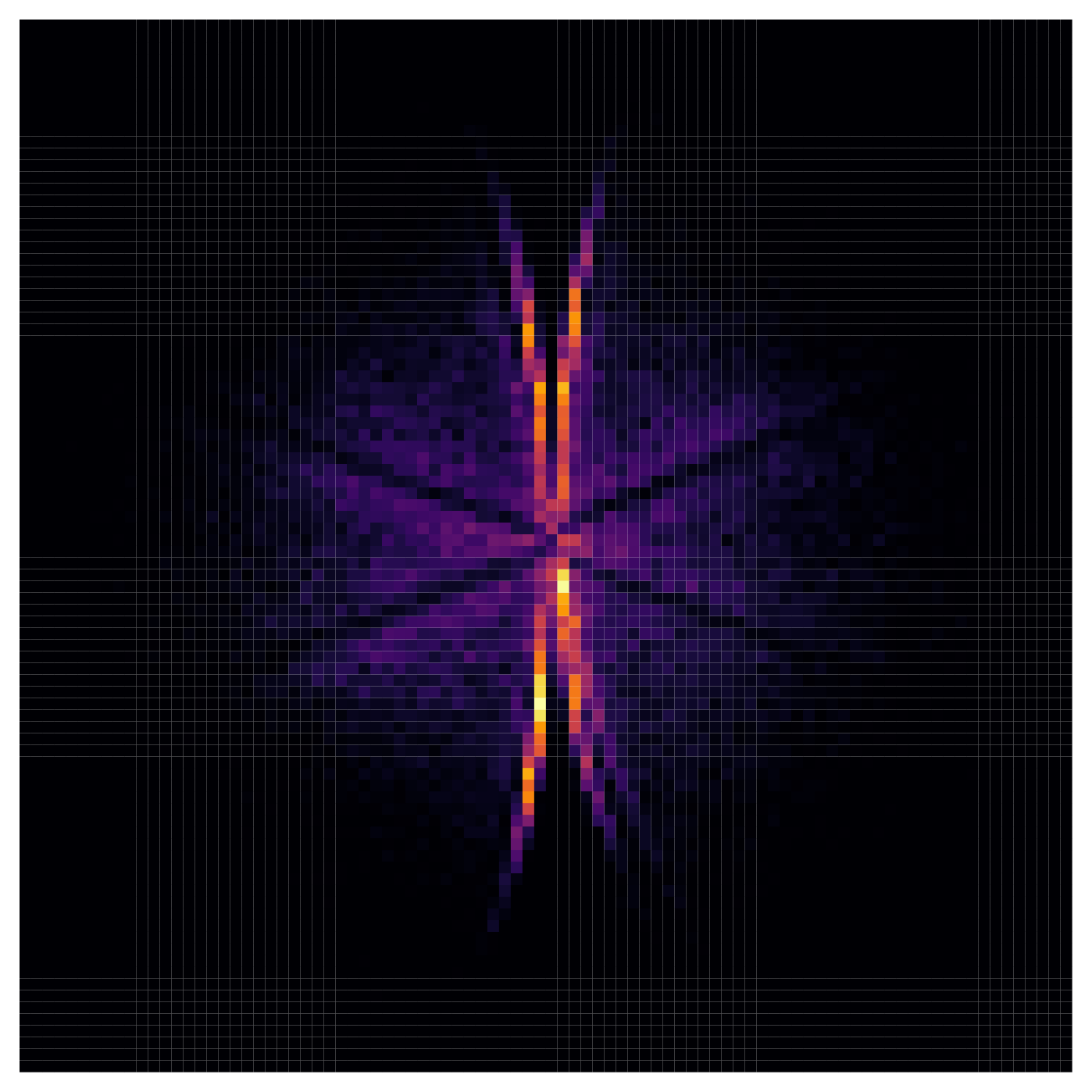}
  &
  \includegraphics[width=0.25\textwidth]{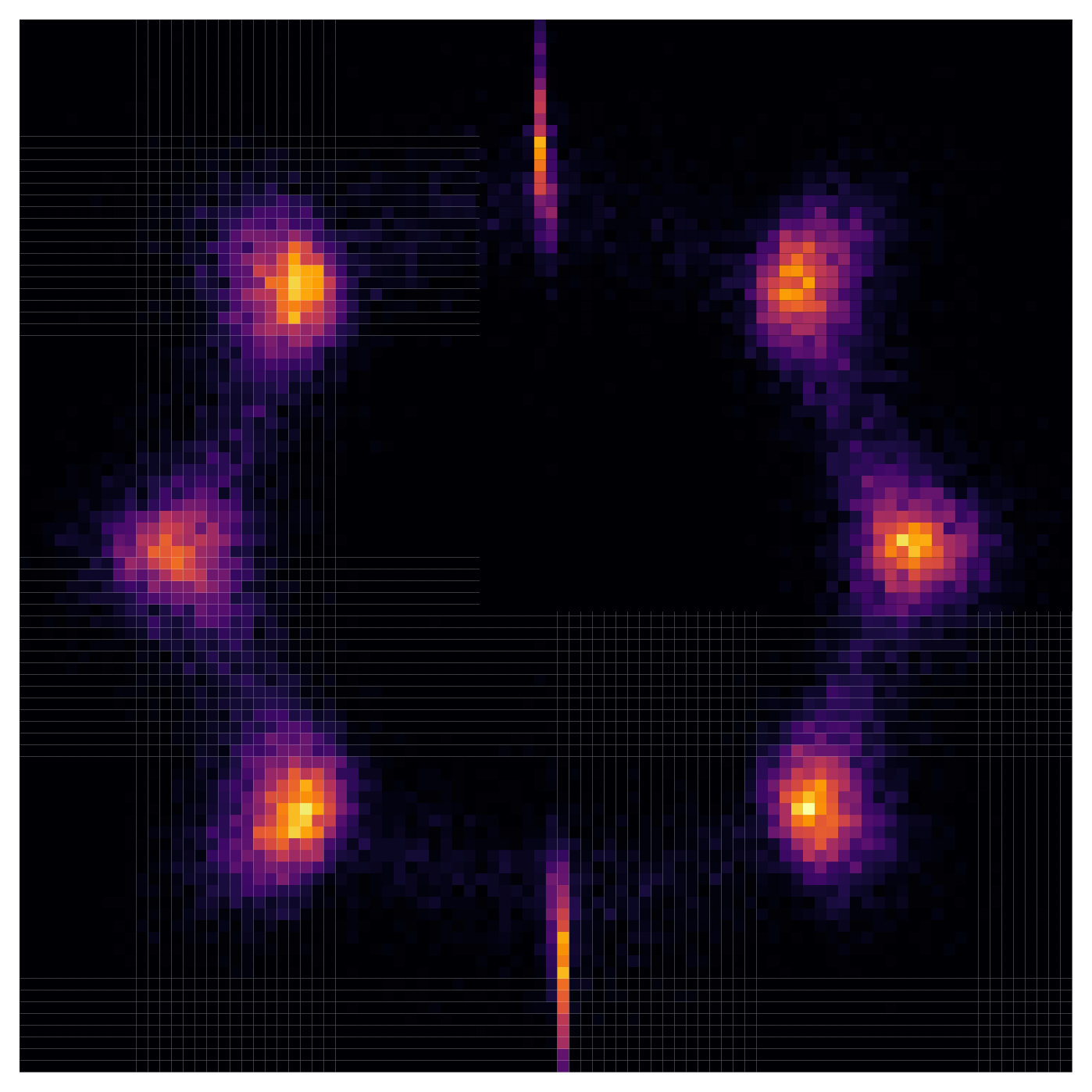}
  \\ 
  \rottext{\textbf{No} HJB \\ 8 Time Steps} 
  &
  \includegraphics[width=0.25\textwidth]{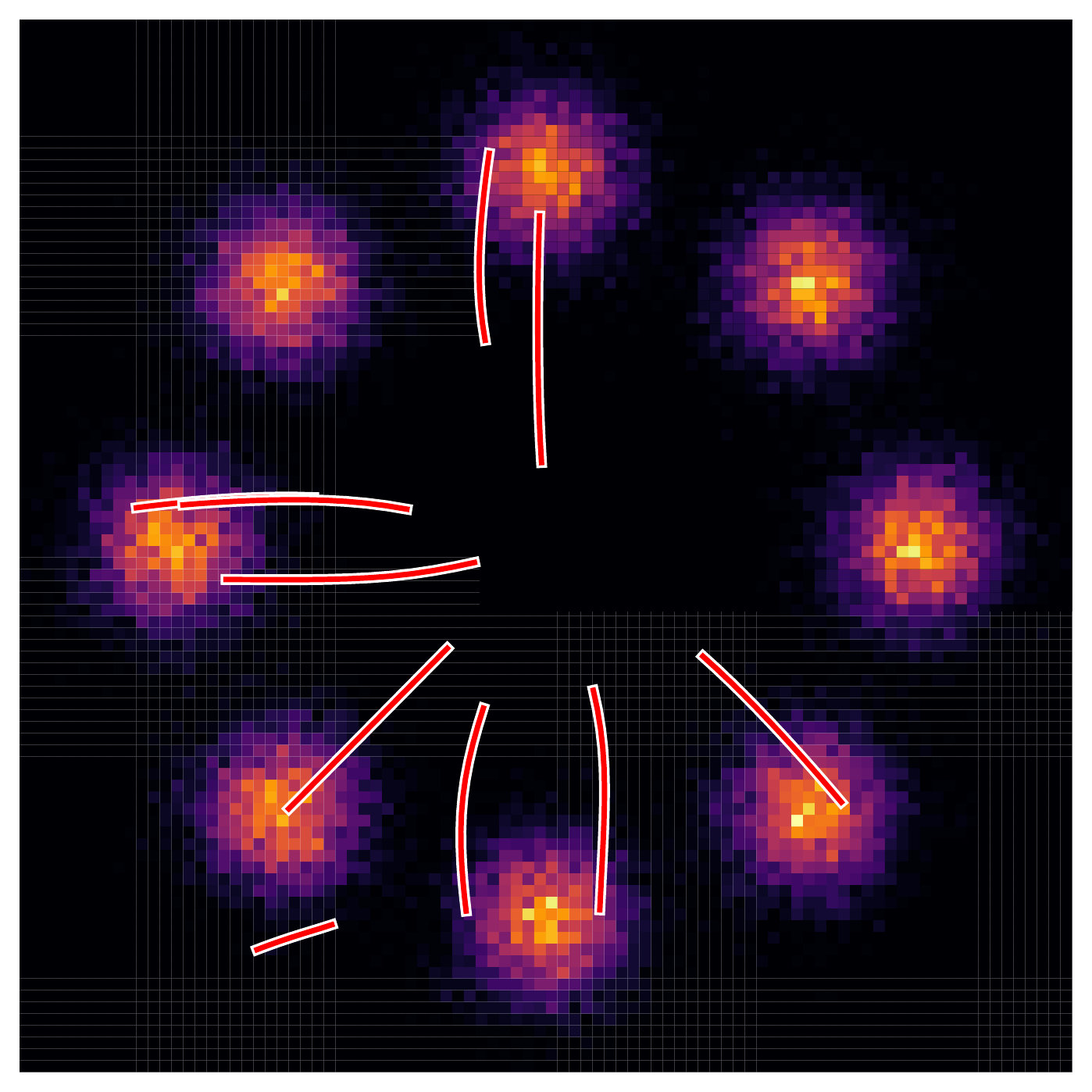}
  &
  \includegraphics[width=0.25\textwidth]{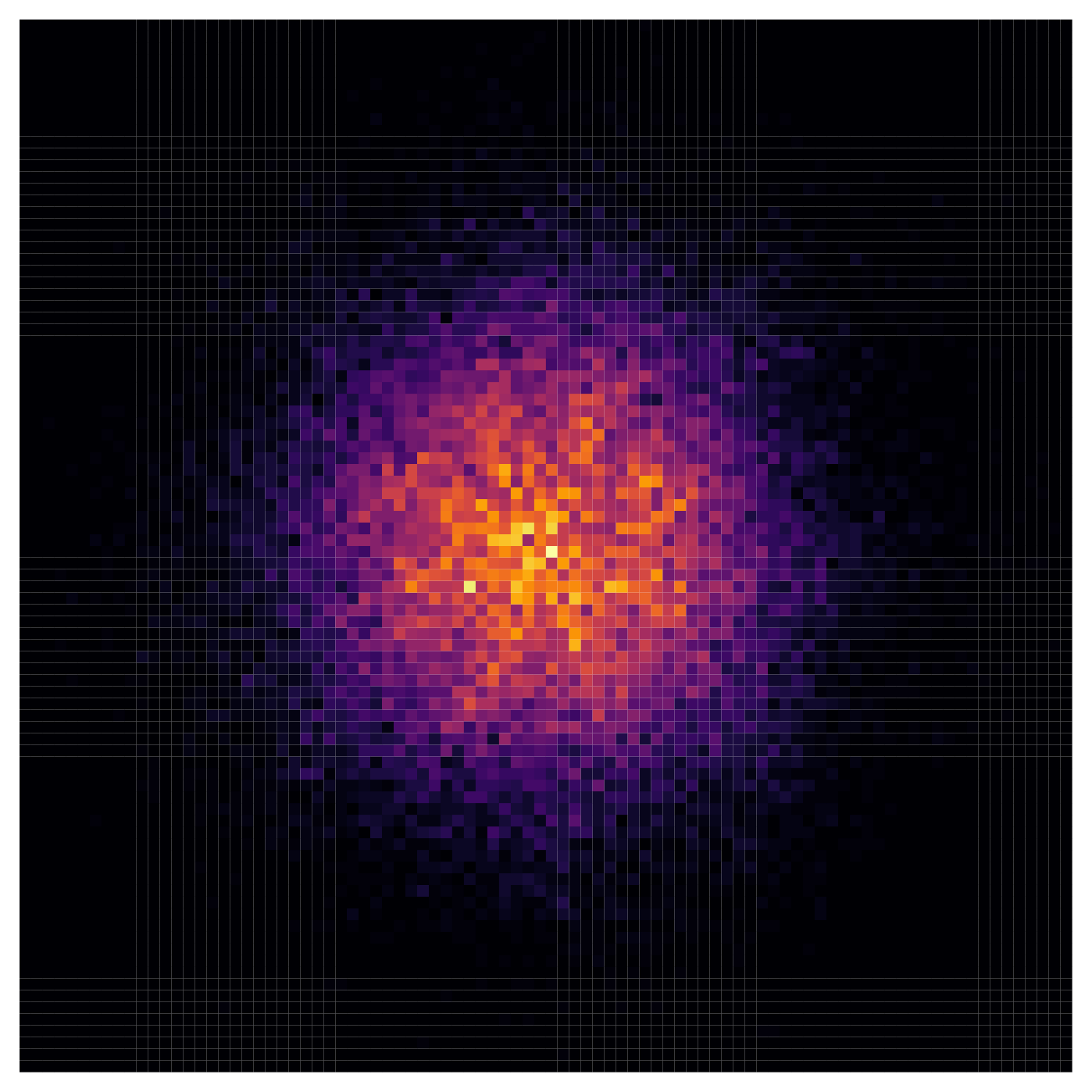}
  &
  \includegraphics[width=0.25\textwidth]{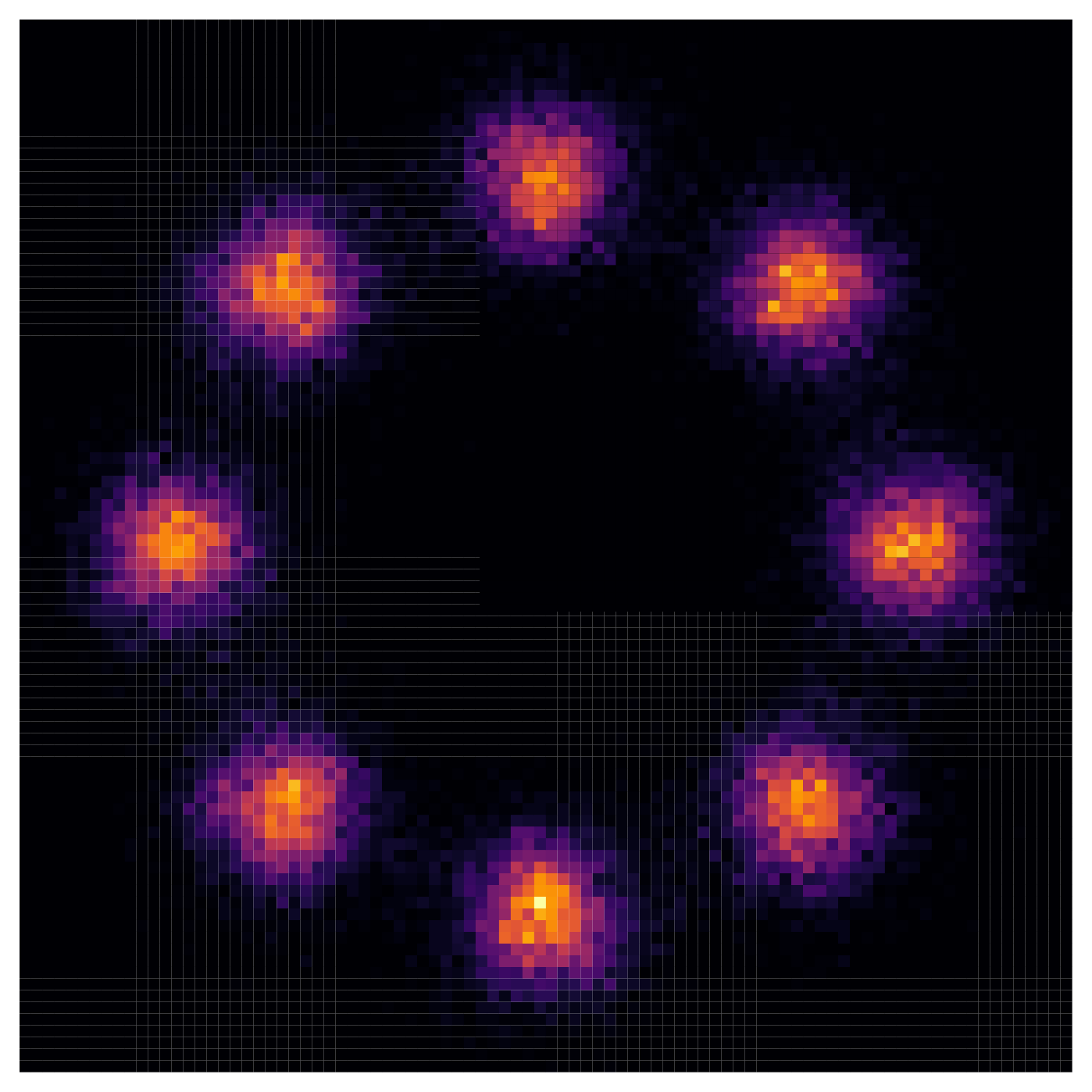}
  \\ 
  \rottext{\textbf{With} HJB \\ 2 Time Steps} 
  &
  \includegraphics[width=0.25\textwidth]{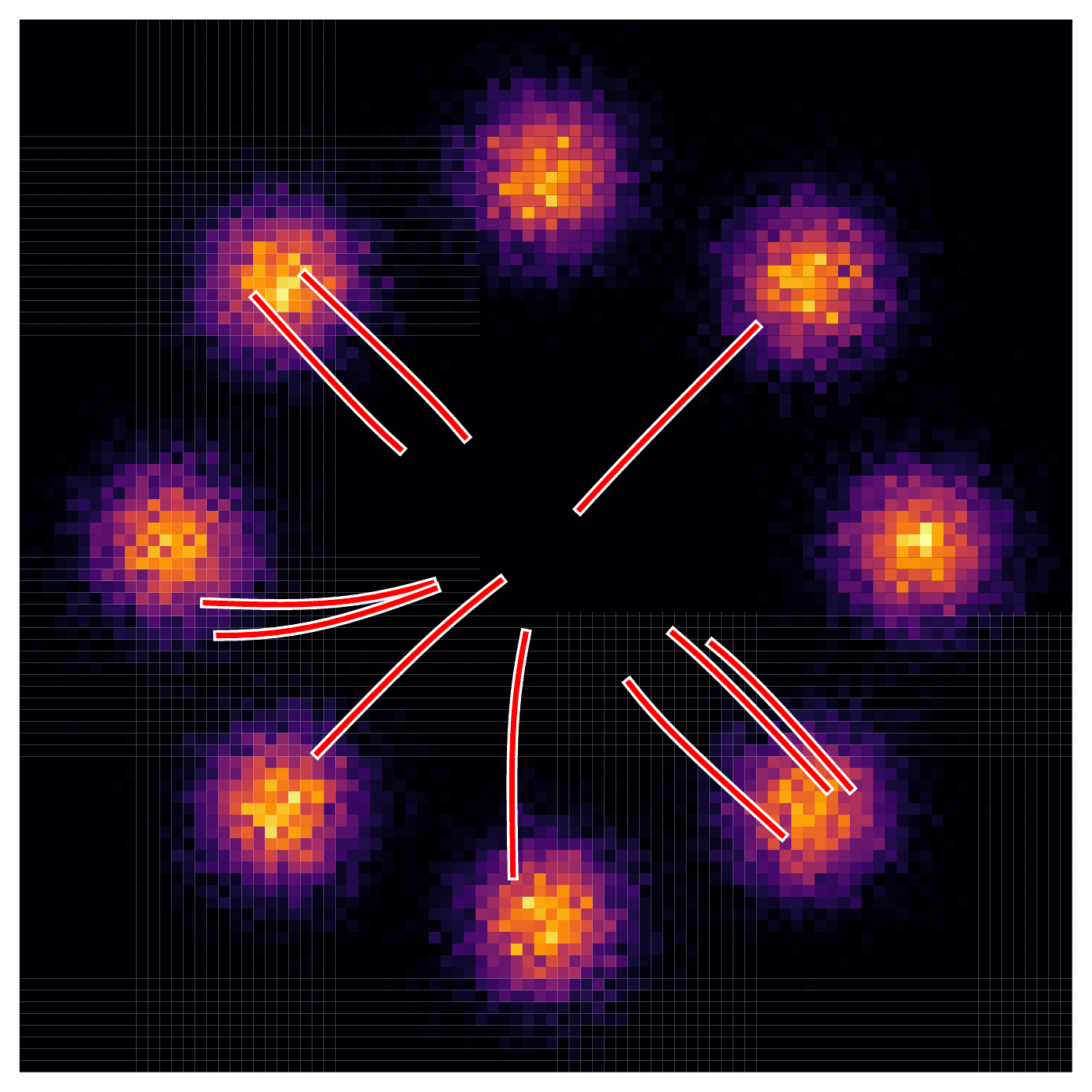}
  &
  \includegraphics[width=0.25\textwidth]{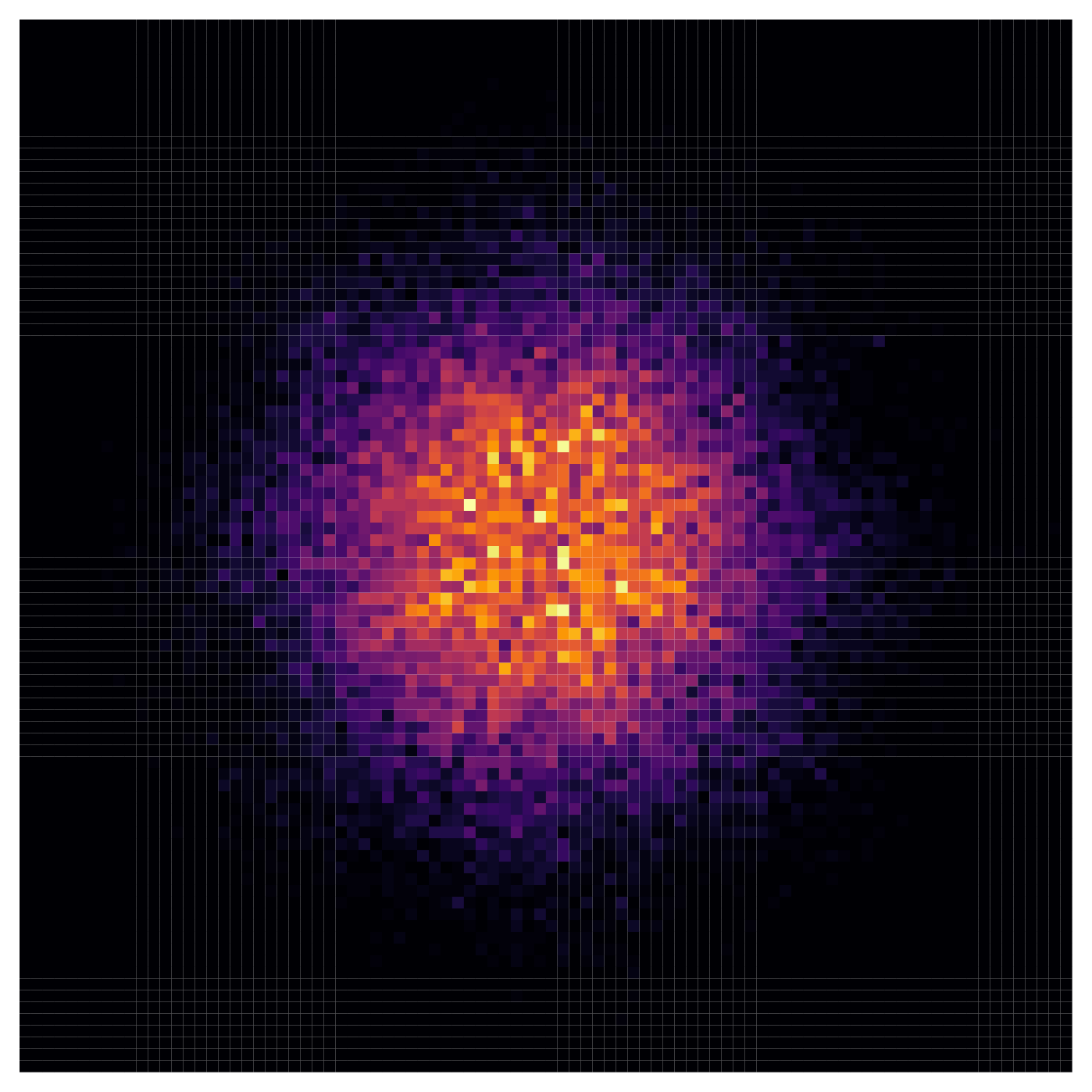}
  &
  \includegraphics[width=0.25\textwidth]{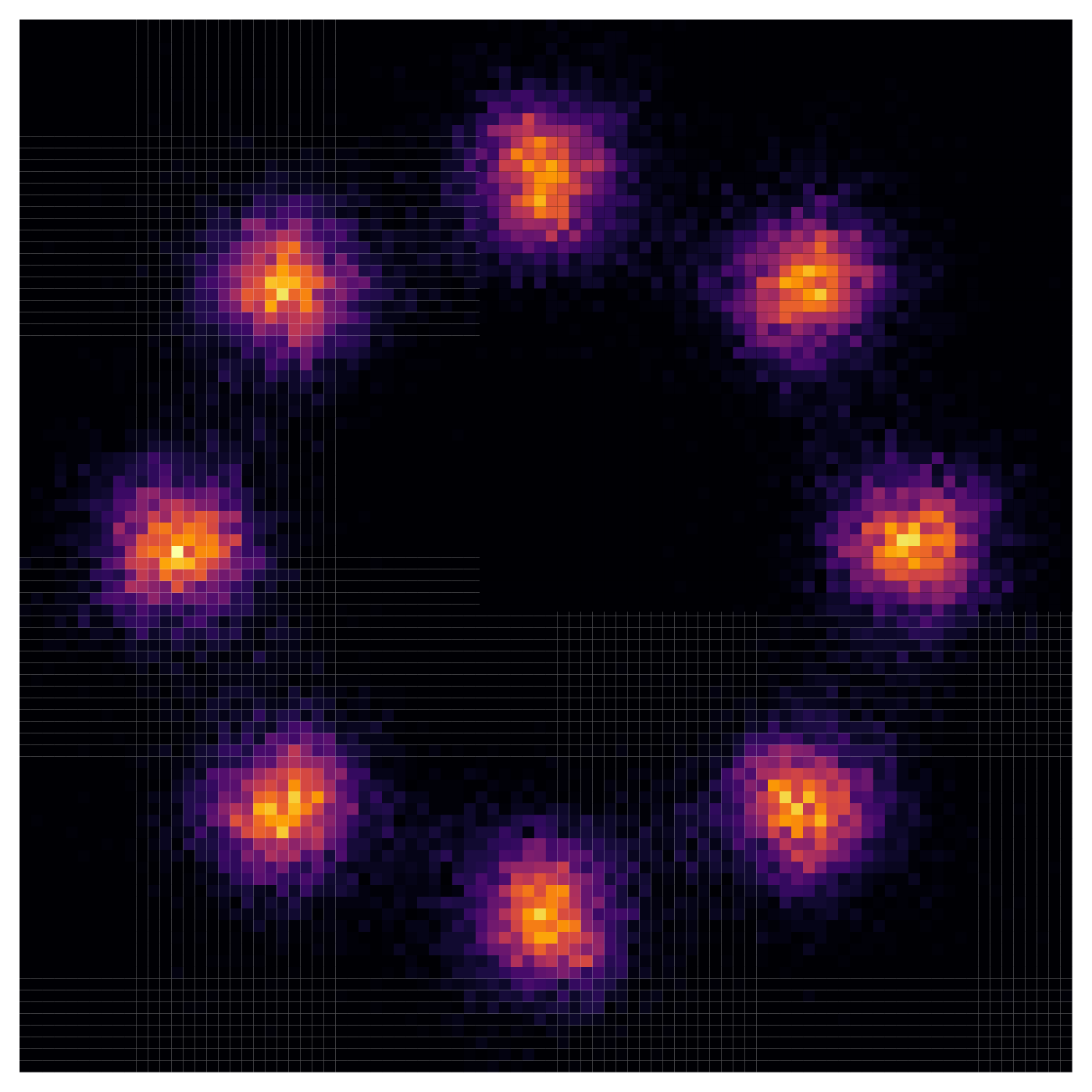}
  \end{tabular}
  \caption{Effect of adding an HJB regularizer during training. 
  The first row presents a flow trained using two RK4 time steps without an HJB regularizer.
  The second row presents a flow trained using eight RK4 time steps without an HJB regularizer.
  The third row presents a flow trained using two RK4 time steps \emph{with} an HJB regularizer.
  For each flow, we show initial, forward mapping, and generation.
  The HJB regularizer allows for training a flow with one-fourth the number of time steps, leading to a drastic reduction in computational and memory costs. White trajectories display the forward flow $f$ for several random samples; red trajectories display the inverse flow $f^{-1}$.}
  \label{fig:effectOfHJBReg}
  \addtolength{\tabcolsep}{3pt} 
\end{figure}

\section{The HJB Regularizer}
\label{app:HJBReg}

	\paragraph{Theory}
	The optimality conditions of~\eqref{eq:LregODE} imply that the potential $\Phi$ satisfies the Hamilton-Jacobi-Bellman (HJB) equations~\cite{evans1983introduction}
	\begin{equation}
		\label{eq:HJB}
		-\partial_t \Phi(\bfx,t) + \frac{1}{2}\|\nabla \Phi(\bfz(\bfx,t),t)\|^2 = 0, \quad \Phi(\bfx,T)=G(\bfx).
	\end{equation}
	where $G$ is the terminal condition of the partial differential equation (PDE).
	Consider the KL divergence in~\eqref{eq:appKL_inZ} after the change of variables is performed.
	OT theory~\cite{villani2008optimal,benamou2017variational} states that the HJB terminal condition is given by
	\begin{equation}
	\label{eq:varderiv}
		\begin{split}
			G(\bfz(\bfx,T)) &\coloneqq \frac{\delta}{\delta \rho_0}  \D_{\rm KL} \big[ \, \rho(\bfz(\bfx,T)) \, || \, \rho_1(\bfz(\bfx,T)) \, \big]
			\\
			&= \frac{\delta}{\delta \rho_0} \int_{\R^d} \Big[ \log\big(\rho_0(\bfx)\big) - \log\big(\rho_1(\bfz(\bfx,T))\big) - \log\det\big(\nabla \bfz(\bfx,T)\big) \Big] \rho_0(\bfx) \, \du \bfx
			\\
			&= 1 + \log\big(\rho_0(\bfx)\big) - \log\big(\rho_1(\bfz(\bfx,T))\big) - \log\det\big(\nabla \bfz(\bfx,T)\big),
		\end{split}
	\end{equation}
	where $\frac{\delta}{\delta \rho_0}$ is the variational derivative with respect to $\rho_0$. 	
	
	While solving~\eqref{eq:HJB} in high-dimensional spaces is notoriously difficult, penalizing its violations along the trajectories is inexpensive. Therefore, we include the value $R(x,T)$ in the objective function, which we accumulate during the ODE solve (Sec.~\ref{sec:OMT-Flow}). The density $\rho_0$, which is required to evaluate $G$, is unknown in our problems. Similar to~\citet{yang2019}, we do not enforce the HJB terminal condition but do enforce the HJB equations for $t \in (0,T)$ via regularizer $R$.

	\paragraph{Effect of Added Regularizer}
	
	For the demonstration (Fig.~\ref{fig:effectOfHJBReg}), we compare three models: two RK4 time steps with no HJB regularizer, eight RK4 time steps with no HJB regularizer, and two RK4 time steps with the HJB regularizer. For several starting points, we plot the forward flow trajectories $f$ in white and the inverse flow $f^{-1}$ trajectories in red. The last two models have straight trajectories, which the first model lacks. All three models are invertible since their forward and inverse trajectories align.

\section{Error Bounds}
\label{app:errors}

	For the timing comparison between our exact trace and the Hutchinson's estimator (Fig.~\ref{fig:trace_compare}), we run 20 replications and compute error bounds via bootstrapping. From the 20 runs, we sample with replacement 4,000 times with size 16. We compute the means for each of these 4,000 samplings and compute the 0.5 and 99.5 percentiles; we present these 99\% confidence intervals in Fig.~\ref{fig:trace_compare} as a shaded area.
	
	For the density estimation on the real data sets, we train three instances for each model and data set. We present the means of these three instance in Tab.~\ref{tab:large} and the standard deviation for each set of three in Tab.~\ref{tab:error_bounds}.

\section{Implementation Details}
\label{app:implementation}

	We incorporate the accumulation of the regularizers in the ODE. The full optimization problem is
	\begin{equation}
		\min_{\bfth} \;\; \E_{\rho_0(\bfx)} \; \Big\{ \alpha_1 C(\bfx, T) + L(\bfx,T) + \alpha_2 R(\bfx,T) \Big\}
	\end{equation}
	subject to
	\begin{equation*}
		\renewcommand*{\arraystretch}{1.2} 
		\begin{split}
		\partial_t\begin{pmatrix}
		\bfz(\bfx,t)\\
		\ell(\bfx,t)\\
		L(\bfx,t) \\
		R(\bfx,t) \\
		\end{pmatrix}=\begin{pmatrix}
		- \nabla \Phi(\bfz(\bfx,t),t; \bfth)\\
		- \tr ( \nabla^2 \Phi(\bfz(\bfx,t),t; \bfth))\\
		\frac{1}{2}\| \nabla \Phi(\bfz(\bfx,t),t; \bfth) \|^2\\
		\left| \partial_t \Phi(\bfz(\bfx,t),t; \bfth) - \frac{1}{2}\|\nabla \Phi(\bfz(\bfx,t),t; \bfth)\|^2 \right|\\
		\end{pmatrix}, \quad 
		\begin{pmatrix}
			\bfz(\bfx,0)\\
			\ell(\bfx,0)\\
			L(\bfx,0) \\
			R(\bfx,0) \\
		\end{pmatrix} = 
		\begin{pmatrix}
			\bfx\\
			0\\
			0 \\
			0 \\
		\end{pmatrix},
		\end{split}
	\end{equation*} 
	where we optimize the weights $\bfth$, defined in~\eqref{eq:NNArchitecture}, that parameterize $\Phi$.
	We include two hyperparameters $\alpha_1,\alpha_2$ to assist the optimization. Specially selected hyperparameters can improve the convergence and performance of the model. Other hyperparameters include the hidden space size $m$, the number of time steps used by the Runge-Kutta 4 solver $n_t$, the number of ResNet layers for which we use 2 for all experiments, and various settings for the ADAM optimizer.

\section{Exact Trace computation} 
\label{app:trace}
	
	We expand on the trace computation formulae presented in Sec.~\ref{sec:trace} for a ResNet with $M+1$ layers.
	
	\paragraph{Gradient Computation}
	To compute the gradient, first note that for an ($M+1$)-layer residual network and given inputs $\bfs = (\bfx,t)$, we obtain $N(\bfs; \bfth_N) = \bfu_M$ by forward propagation 
	\begin{equation}
	    \begin{split}
	    \bfu_0 & = \sigma(\bfK_0 \bfs + \bfb_0) \\ 
	    \bfu_1 & = \bfu_0 + h \, \sigma(\bfK_1 \bfu_0 + \bfb_1)\\
	    \vdots & \qquad\qquad\vdots\\  
	    \bfu_M &= \bfu_{M-1} + h \, \sigma(\bfK_M \bfu_{M-1} + \bfb_M),\\  
	    \end{split}
	\end{equation}
	where $h>0$ is a fixed step size, and the network's weights are $\bfK_0 \in \R^{m \times (d+1)}$, $\bfK_1,\ldots, \bfK_M \in \R^{m\times m}$, and $\bfb_0, \ldots, \bfb_M \in \R^m$.

	The gradient of the neural network is computed using backpropagation as follows
	\begin{equation} \label{eq:app_backprop}
	\begin{split}
	 	\bfz_{M+1} &= \bfw \\
	    \bfz_M     &= \bfz_{M+1} + h \, \bfK_M^\top \, {\rm diag}\big(\sigma'(\bfK_M \bfu_{M-1} + \bfb_M)\big) \bfz_{M+1},\\  
	    \vdots & \qquad \qquad\vdots\\
	    \bfz_{1} &= \bfz_2 + h \, \bfK_{1}^\top {\rm diag}\big(\sigma'(\bfK_{1} \bfu_{0} + \bfb_{1})\big) \bfz_2,\\  
	    \bfz_{0} &= \bfK_{0}^\top {\rm diag}\big(\sigma'(\bfK_{0} \bfs + \bfb_{0})\big) \bfz_1,\\  
 	\end{split}
	\end{equation}
	which gives $\nabla_{\bfs} N(\bfs;\bfth_N) \bfw = \bfz_0$.

\paragraph{Exact Trace Computation} 
	
	Using~\eqref{eq:laplace} and the same $\bfE$, we compute the trace in one forward pass through the layers.
	The trace of the first ResNet layer is
	\begin{equation}
	\begin{split}
	    t_0  & = \tr \Big( \bfE^\top \, \nabla_{\bfs} \big( \bfK_0^\top {\rm diag}(\sigma'(\bfK_0 \bfs + \bfb_0)) \bfz_1\big) \, \bfE \Big)\\
	         & = \tr \left(\bfE^\top \bfK_0^\top \, {\rm diag} \big(\sigma''(\bfK_0 \bfs + \bfb_0) \odot \bfz_1 \big) \, \bfK_0 \bfE \right)\\
	         & = \big(\sigma''(\bfK_0 \bfs + \bfb_0) \odot \bfz_1\big)^\top \big((\bfK_0 \bfE)\odot(\bfK_0 \bfE)\big) \mathbf{1},
	\end{split}
	\end{equation}
	using the same notation as~\eqref{eq:trace}. For the last step, we used the diagonality of the middle matrix.
	Computing $t_0$ requires $\mathcal{O}(m\cdot d)$ FLOPS when first squaring the elements in the first $d$ columns of $\bfK_0$, then summing those columns, and finally one inner product. 
	To compute the trace of the entire ResNet, we continue with the remaining rows in~\eqref{eq:app_backprop} in reverse order to obtain
	\begin{equation}
	    \tr \left( \bfE^\top \, \nabla_{\bfs}^2 ( N(\bfs;\bfth_N) \bfw) \, \bfE \right) = t_0 + h \sum_{i=1}^M t_i,
	\end{equation}
	where $t_i$ is computed as
	 \begin{equation*}
	    \begin{split}
	        t_i & = \tr \left(\bfJ_{i-1}^\top \nabla_{\bfs} \big( \bfK_i^\top {\rm diag}(\sigma'(\bfK_i \bfu_{i-1}(\bfs) + \bfb_i)) \bfz_{i+1} \big) \bfJ_{i-1}\right) \\
	           & = \tr \left(\bfJ_{i-1}^\top \bfK_i^\top \,{\rm diag} \big(\sigma''(\bfK_i \bfu_{i-1} + \bfb_i) \odot \bfz_{i+1} \big) \, \bfK_i \bfJ_{i-1} \right)\\
	           & = \big(\sigma''(\bfK_i \bfu_{i-1} + \bfb_i) \odot \bfz_{i+1}\big)^\top \big((\bfK_i \bfJ_{i-1})\odot(\bfK_i \bfJ_{i-1})\big) \mathbf{1}.
	    \end{split}
	\end{equation*}
	Here, $\bfJ_{i-1} = \nabla \bfu_{i-1}^\top \in \R^{m\times d}$ is a Jacobian matrix, which can be updated and over-written in the forward pass at a computational cost of $\bigO(m^2 \cdot d)$ FLOPS.
	The $\bfJ$ update follows:
	\begin{equation}
	  \begin{aligned}
	      \nabla \bfu_i^\top &= \nabla \bfu_{i-1}^\top + {\rm diag} \big( h \, \sigma' ( \bfK_i \bfu_{i-1} + \bfb_i) \big) \bfK_i  \nabla \bfu_{i-1}^\top \\
	      \bfJ &\leftarrow \bfJ + {\rm diag} \big( h \, \sigma' ( \bfK_i \bfu_{i-1} + \bfb_i) \big) \, \bfK_i \bfJ
	  \end{aligned}
	\end{equation}

    Since we parameterize the potential $\Phi$ instead of the $\bfv$, the Jacobian of the dynamics $\nabla \bfv$ is given by the Hessian of $\Phi$ in~\eqref{eq:neural_odes}. We note that Hessians are \emph{symmetric} matrices. We use the exact trace; however, if we wanted to use a trace estimate, a plethora of estimators perform better in accuracy and speed on symmetric matrices than on nonsymmetric matrices~\cite{hutchinson1990stochastic,avron2011randomized,ubaru2017fast}.

\begin{figure*}[t]
	\centering
	\begin{tabular}{llrrrrr} 
	\toprule
	\multirow{2}{*}{Data Set} & \multirow{2}{*}{Model} & \multirow{2}{*}{\# Param} & Training & Testing & Inverse & \multirow{2}{*}{MMD} \\
	&              &                     & Time (s) & Loss    & Error & \\
	\midrule
	\multirow{2}{*}{\textbf{Gaussian Mixture}} 
	& \model{} &  637 & 189  & 2.88 & 1.28\eu-8 & 6.38\eu-4 \\
	& FFJORD   & 9225 & 7882 & 2.85 & 7.22\eu-8 & 6.54\eu-4 \\ 
	\bottomrule
	\end{tabular}
	\includegraphics[clip, trim=2.3cm 2.3cm 2.3cm 2.3cm, width=0.6\linewidth]{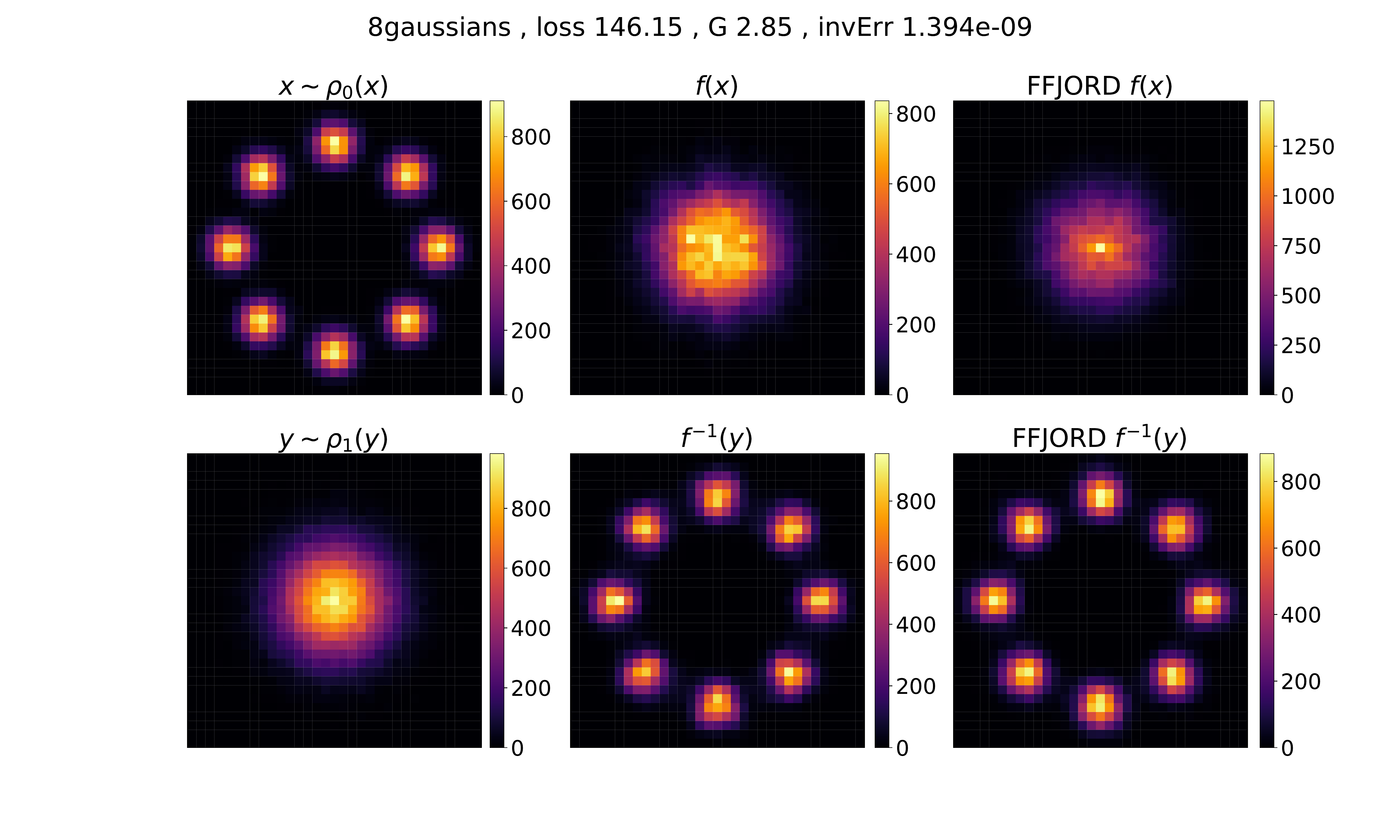}
 	\caption{CNF performance heatmaps for the toy Gaussian mixture problem. (top row) $10^5$ samples $\bfx$ from the pretended unknown $\rho_0$, the forward propagations of our flow $f(\bfx)$ and the FFJORD flow.
 	(bottom row) $10^5$ samples $\bfy$ drawn from the known $\rho_1$, our model's generation $f^{-1}(\bfy)$ using the inverse flow on normal samples and FFJORD's inverse flow on the same normal samples $\bfy$. }
 	\label{fig:8gauss}
	\end{figure*}

	\begin{figure*}
	    \centering
		\begin{minipage}{0.5\linewidth}
			\centering
			\begin{tabular}{cccc}
				\multicolumn{4}{c}{Discrete normalizing flow trained on \power{}} \\
				\midrule
				$d$ & \textbf{Testing Loss} & Inv Error & MMD \\
				\midrule
				6 & $\mathbf{-0.64}$ & 2.34\eu-3 & 1.94\eu-2 \\
			\end{tabular}
		\end{minipage}\\
		\includegraphics[clip, trim=4cm 2cm 0.5cm 0cm, width=0.28\linewidth]{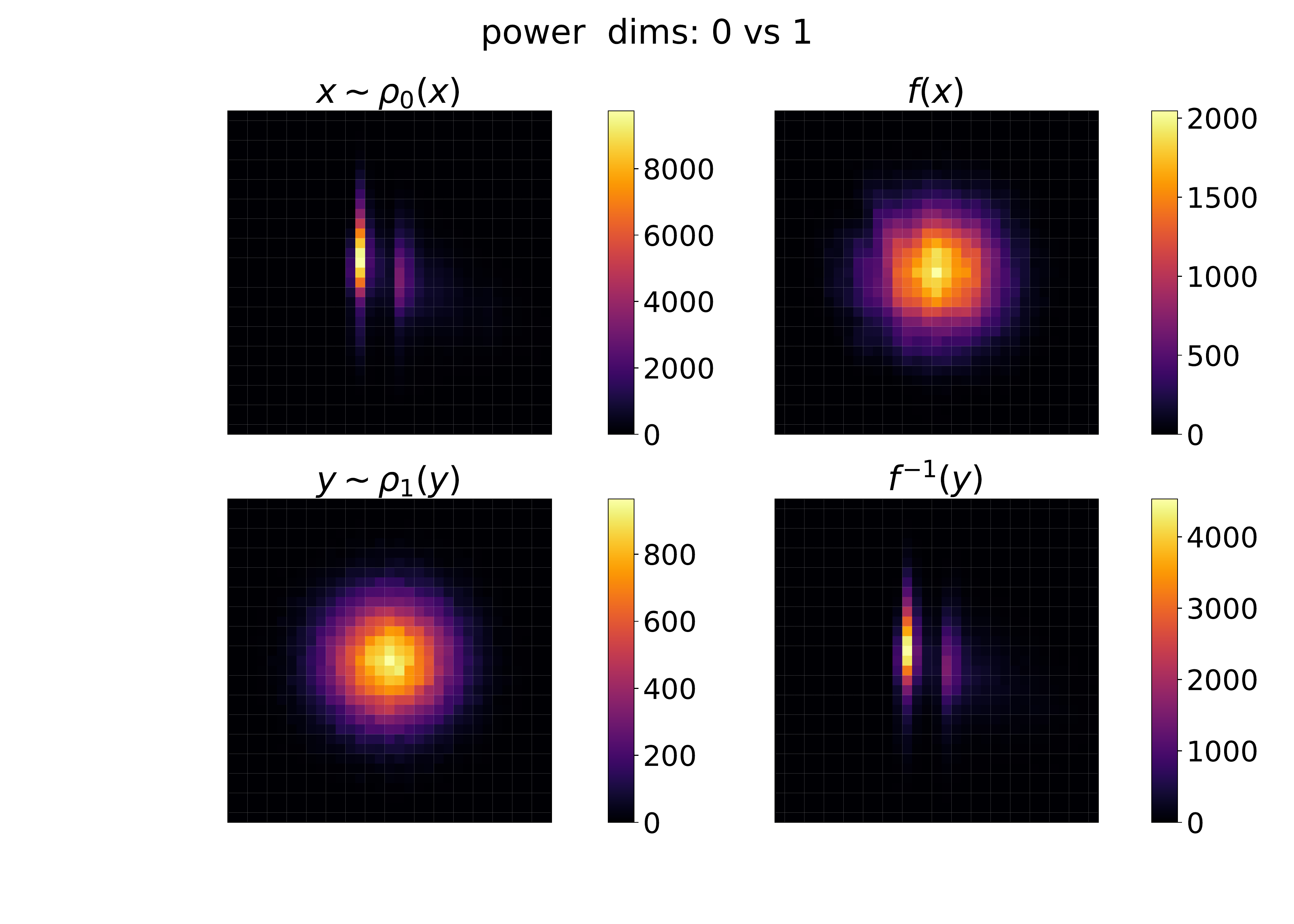}
		\includegraphics[clip, trim=4cm 2cm 0.5cm 0cm, width=0.28\linewidth]{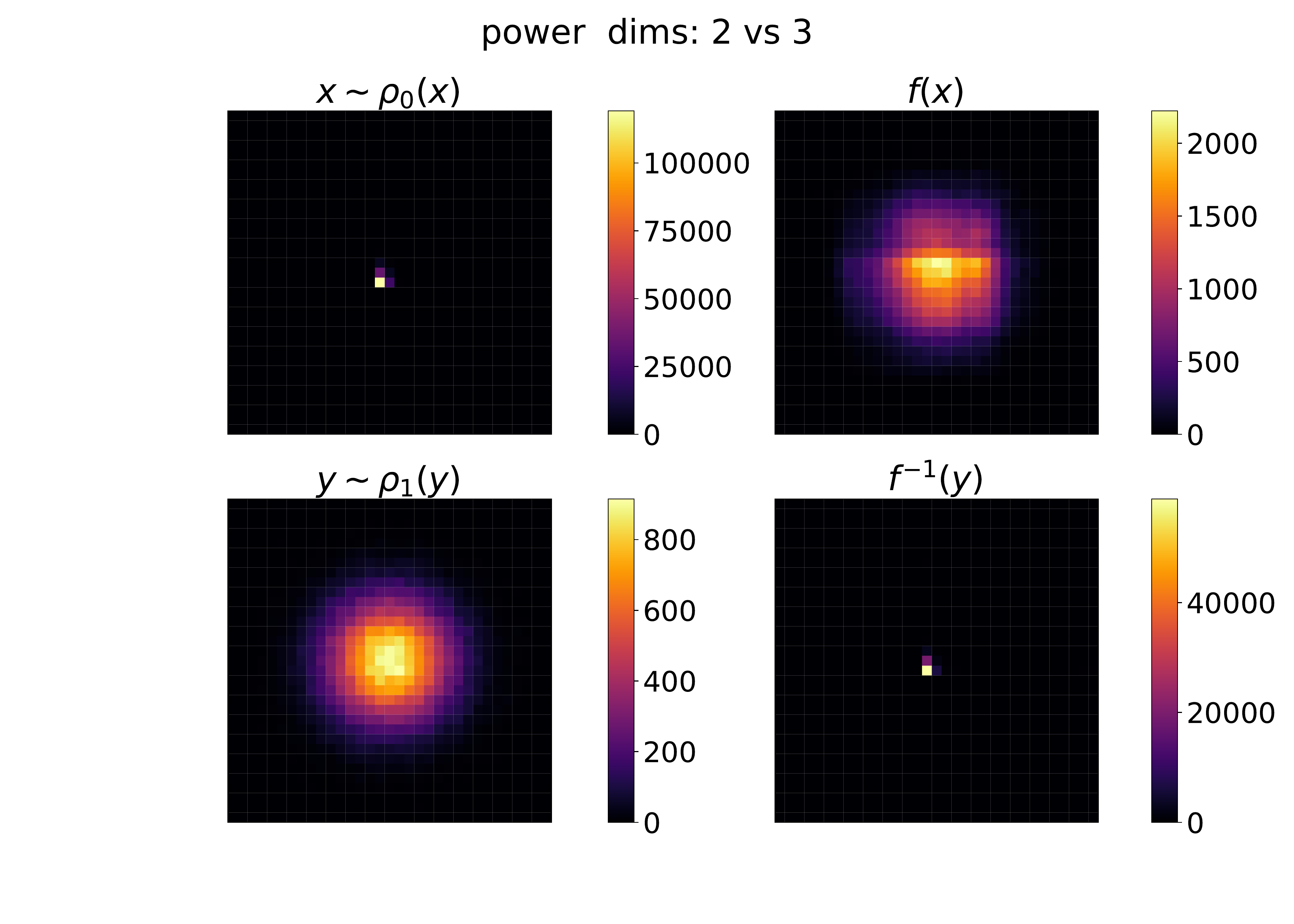}
		\includegraphics[clip, trim=4cm 2cm 0.5cm 0cm, width=0.28\linewidth]{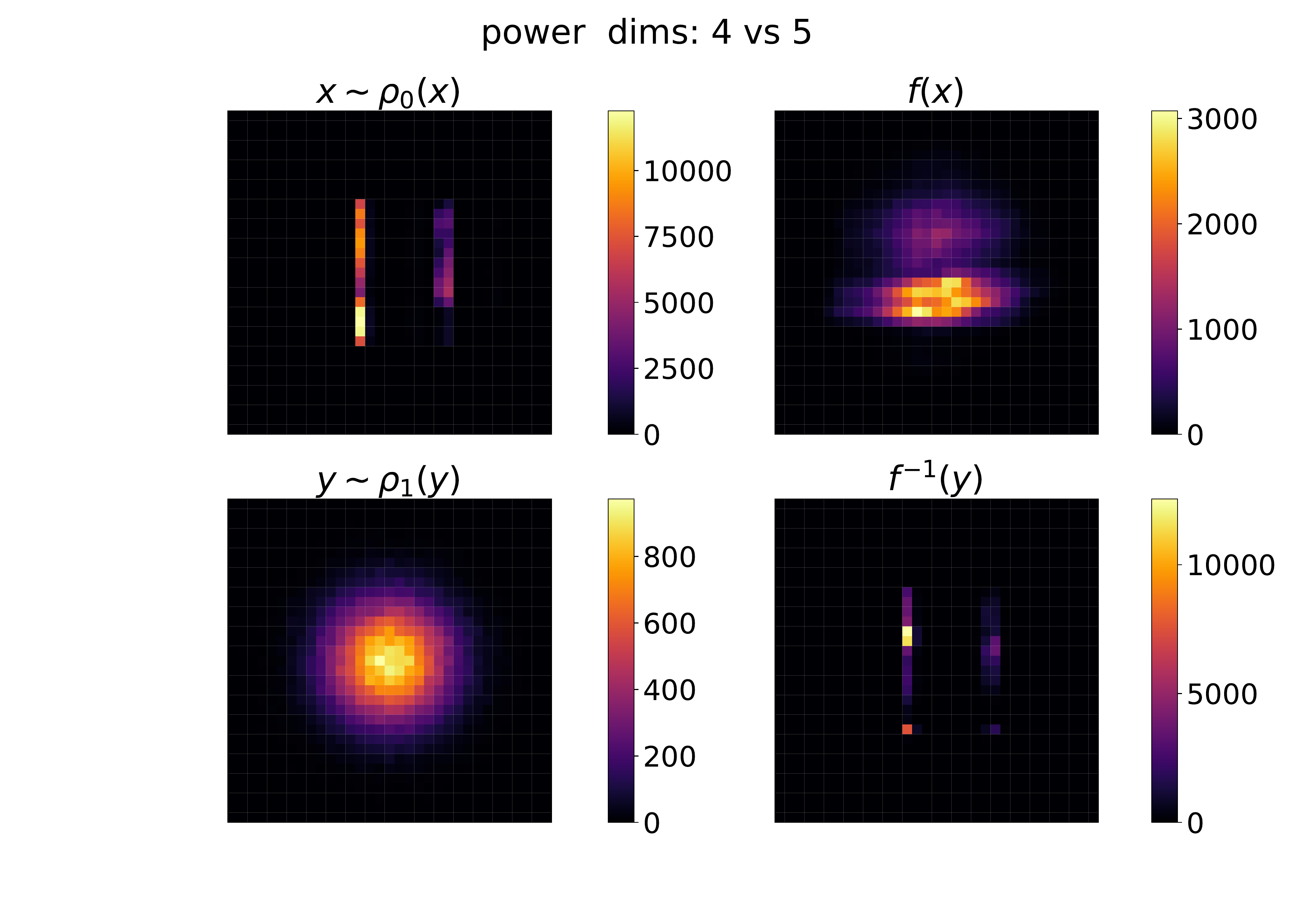}\\
	    \caption{\power{} density estimation for some discrete normalizing flow. By the testing loss metric, this model is considered very competitive. However, the model itself performs poorly, as clear in the visualization of the last two dimensions. The MMD shows that the generation is poor. The inverse error shows that the testing loss uses an integration scheme that is too coarse, as addressed in~\citet{wehenkel2019unconstrained} and ~\citet{onken2020do}.}
	    \label{fig:bad_power}
	\end{figure*}

	\begin{table*}
		\centering
		 \begin{tabular}{lccccc}                                          
		 \toprule
		                             & \power{} & \gas{} & \hepmass{} & \miniboone{} & \bsds{} \\
		 \midrule
		 \model{} (Ours) 			& \hphantom{0}18K & 127K & \hphantom{.0}72K & \hphantom{.0}78K & \hphantom{.}297K \\
			FFJORD \& RNODE					& \hphantom{0}43K & 279K & \hphantom{.}547K & \hphantom{.}821K & \hphantom{0}6.7M \\
			\midrule
			NAF~\cite{huang2018neural}		     & 414K & 402K & 9.27M & 7.49M & 36.8M\\
			UMNN~\cite{wehenkel2019unconstrained}	 & 509K & 815K & 3.62M  & 3.46M  & 15.6M \\      
		 \bottomrule                               
		 \end{tabular}
		\caption{Number of parameters comparison with discrete normalizing flows.}
		\label{tab:params}
	\end{table*}

	\begin{table*}
		\centering 
	    \begin{tabular}{lccccc}  
	    \toprule
	                            & \power{} & \gas{} & \hepmass{} & \miniboone{} & \bsds{} \\
	    \midrule
	    \model{} (Ours)  & -0.30 & \hphantom{0}-9.20 & 17.32  & 10.55 & -154.20\hphantom{$^\dagger$}  \\
	    RNODE trained by us  & -0.39 & -11.10 &  16.37 &  10.65  & -129.75$^\dagger$\\
		FFJORD trained by us & -0.37 & -10.69 &  16.13 &  10.57  & -133.94$^\dagger$ \\
		\midrule
		MADE~\cite{germain15}    & \hphantom{-}3.08 & \hphantom{0}-3.56 & 20.98 & 15.59 & -148.85\hphantom{$^\dagger$} \\
		RealNVP~\cite{dinh2016density}	& -0.17 & \hphantom{0}-8.33  & 18.71 & 13.55 & -153.28\hphantom{$^\dagger$} \\
		Glow~\cite{kingma2018glow}			& -0.17 & \hphantom{0}-8.15  & 18.92 & 11.35 & -155.07\hphantom{$^\dagger$} \\
		MAF~\cite{papamakarios2017masked}   & -0.24 & -10.08 & 17.70 & 11.75 & -155.69\hphantom{$^\dagger$} \\
		NAF~\cite{huang2018neural}          & -0.62 & -11.96 & 15.09 & \hphantom{0}8.86 & -157.73\hphantom{$^\dagger$} \\
		UMNN~\cite{wehenkel2019unconstrained} & -0.63 & -10.89 & 13.99 & \hphantom{0}9.67  & -157.98\hphantom{$^\dagger$} \\      
	    \bottomrule
	    \multicolumn{6}{l}{\small{$^\dagger$Training terminated before convergence.}}                                 
	    \end{tabular}
    	\caption{Testing Loss $C$ comparison with other models.}
    	\label{tab:test_loss}
	\end{table*}

\begin{table*}
\centering
    \begin{tabular}{clrrrrrrrr}
    \toprule
     \multirow{2}{*}{\vspace{-5pt}Data Set} & \multirow{2}{*}{\vspace{-5pt}Model}  &  \multicolumn{4}{c}{Training} & \multicolumn{4}{c}{Testing} \\
    \cmidrule(lr){3-6} \cmidrule(lr){7-10} 
     &  &  Time (h) & \# Iter & $\frac{\text{Time}}{\text{Iter}}$(s) & NFE & Time (s) & Inv Err & MMD & \multicolumn{1}{c}{$C$}\\
    \midrule
    \multirow{3}{*}{\shortstack[*]{\textbf{\power{}}\\$d=$ 6}} &
    	\model{}  &  $\pm$0.7 & $\pm$5.8K & $\pm$0.06 & $\pm$0     & $\pm$0.1  & $\pm$1.5e-6  & $\pm$3.26e-6 & $\pm$0.02\\
        & RNODE   &  $\pm$3.9 & $\pm$4.8K & $\pm$0.03 & $\pm$0     & $\pm$3.5  & $\pm$6.1e-7  & $\pm$1.34e-5  & $\pm$0.02\\
        & FFJORD  &  $\pm$8.1 & $\pm$5.6K & $\pm$0.76 & $\pm$51.7  & $\pm$2.7  & $\pm$1.4e-6  & $\pm$6.74e-6 & $\pm$0.06\\
    \midrule
    \multirow{3}{*}{\shortstack[*]{\textbf{\gas{}}\\$d=$ 8}} &
    	\model{}  & $\pm$0.2  & $\pm$4.0K & $\pm$0.04 & $\pm$0 & $\pm$0.07  & $\pm$5.9e-5  & $\pm$3.02e-5 & $\pm$0.02\\
        & RNODE   & $\pm$9.4  & $\pm$15K  & $\pm$0.01 & $\pm$0 & $\pm$67.1 & $\pm$1.1e-5  & $\pm$2.24e-5 & $\pm$0.30\\
        & FFJORD  & $\pm$13.1 & $\pm$6.8K & $\pm$0.33 & $\pm$36.1 & $\pm$58.6 & $\pm$8.4e-6  & $\pm$3.28e-5 & $\pm$0.15\\
    \midrule
    \multirow{3}{*}{\shortstack[*]{\textbf{\hepmass{}}\\$d=$ 21}}  &
    	\model{} &  $\pm$1.1 & $\pm$7.2K & $\pm$0.01 & $\pm$0    & $\pm$0.06  & $\pm$8.3e-8  & $\pm$1.20e-8 & $\pm$0.18\\
        & RNODE  &  $\pm$0.5 & $\pm$0.3K & $\pm$0.02 & $\pm$0    & $\pm$76.1  & $\pm$3.0e-6 & $\pm$1.07e-8 & $\pm$0.25\\ 
        & FFJORD &  $\pm$8.0 & $\pm$3.8K & $\pm$0.13 & $\pm$11.0 & $\pm$69.2  & $\pm$2.2e-6 & $\pm$1.00e-8 & $\pm$0.40\\ 
    \midrule                    
    \multirow{3}{*}{\shortstack[*]{\textbf{\miniboone{}}\\$d=$ 43}}  & 
       	\model{} &  $\pm$0.10 & $\pm$0.9K & $\pm$5.9e-3 & $\pm$0  & $\pm$0.01 & $\pm$2.5e-7  & $\pm$8.96e-9 & $\pm$0.04\\
       	& RNODE  &  $\pm$0.09 & $\pm$1.0K & $\pm$9.0e-4 & $\pm$0  & $\pm$2.4  & $\pm$1.6e-6  & $\pm$3.68e-9 & $\pm$0.10\\
       	& FFJORD &  $\pm$0.96 & $\pm$1.6K & $\pm$0.05  & $\pm$2.6 & $\pm$2.8  & $\pm$1.2e-6  & $\pm$4.45e-8 & $\pm$0.07\\
    \midrule
    \multirow{3}{*}{\shortstack[*]{\textbf{\bsds{}}\\$d=$ 63}} & 
    	\model{} & $\pm$1.3  & $\pm$6.4K & $\pm$0.01 & $\pm$0  & $\pm$0.61  & $\pm$3.4e-5 & $\pm$1.70e-4 & $\pm$0.23\\
        & RNODE  & $\pm$3.7  & $\pm$0.6K & $\pm$0.1  & $\pm$0  & $\pm$639.0 & $\pm$3.2e-7 & $\pm$5.64e-3 & $\pm$1.14\\  
        & FFJORD & $\pm$27.7 & $\pm$1.8K & $\pm$2.0  & $\pm$20.0  & $\pm$1197.5 & $\pm$6.6e-7 & $\pm$4.40e-3 & $\pm$6.40\\      
    \bottomrule                            
    \end{tabular}
    \caption{Error bounds for density estimation on real data sets. We provide the standard deviations computed for three runs.} 
    \label{tab:error_bounds}  
\end{table*}%

\section{Number of Parameters}
\label{app:parameters}

	CNFs use fewer parameters than many discrete normalizing flows~\cite{chen2018neural}. We observe that \model{} further reduces the number of parameters necessary to solve many CNF problems (Tab.~\ref{tab:params}). We attribute this to the OT formulation, which encapsulates the inherent physics of the dynamics. As a result, \model{} requires fewer parameters to fit. In our experiments, we observed that training FFJORD and RNODE with the same number of parameters as \model{} resulted in models that lacked sufficient expressibility. These models with reduced parameterization converged to poor MMD and loss values.

\section{Loss Metric}
\label{app:test_loss}
	
	The testing loss metric $C$ depends on the $\ell$ computation in~\eqref{eq:neural_odes}, which is the integration of the trace along the computed trajectory $\bfz$. 
	Different integration schemes have various error when integrating the trace~\cite{wehenkel2019unconstrained,onken2020do}. Too coarse of a time discretization can result in a low $C$ value while sacrificing invertibility. Furthermore, a low $C$ value does not imply good quality generation~\cite{theis2016mmd}.
	As a result, testing loss is unreliable for comparative evaluation of models' performances, so we use MMD.
	
	Visualizations present the best evaluation of a flow's performance. We motivate this with a thorough comparison of \model{} against FFJORD (Fig.~\ref{fig:8gauss}). Even though the testing losses are similar, the FFJORD flow pushes too many points to the origin and does not map well to a Gaussian. We can see this flaw when using $10^5$ samples.
		
	In high-dimensions, visualizations become difficult, which is why reliance on a loss function is appealing. We visualize two-dimensional slices of these high-dimensional point clouds using binned heatmaps (App.~\ref{app:extra_figs}). We then get a sense for which dimensions are or are not mapping to $\rho_1$. For instance, we present an arbitrary finite normalizing flow trained on the \power{} data set in which the testing loss looks competitive, but other metrics and the visualization demonstrate the model's flaws (Fig.~\ref{fig:bad_power}). The last two dimensions show that the forward flow $f(\bfx)$ noticeably differs from $\rho_1$ in these two dimensions. The associated MMD is poor for this model on the \power{} data set (comparable MMDs in Tab.~\ref{tab:large}), and the high inverse error suggests that the integration is not trustworthy. However, the model achieves a testing loss of $-0.64$ which outperforms numerous state-of-the-art models (Tab.~\ref{tab:test_loss}). Motivated by these demonstrations, we argue against using the testing loss metric to evaluate flows.

\section{Visualizations of High-Dimensional Data Sets} \label{app:extra_figs}

 \begin{figure*}[h]
     \centering
      \includegraphics[clip, trim=2cm 2cm 1.6cm 0cm, width=0.48\linewidth]{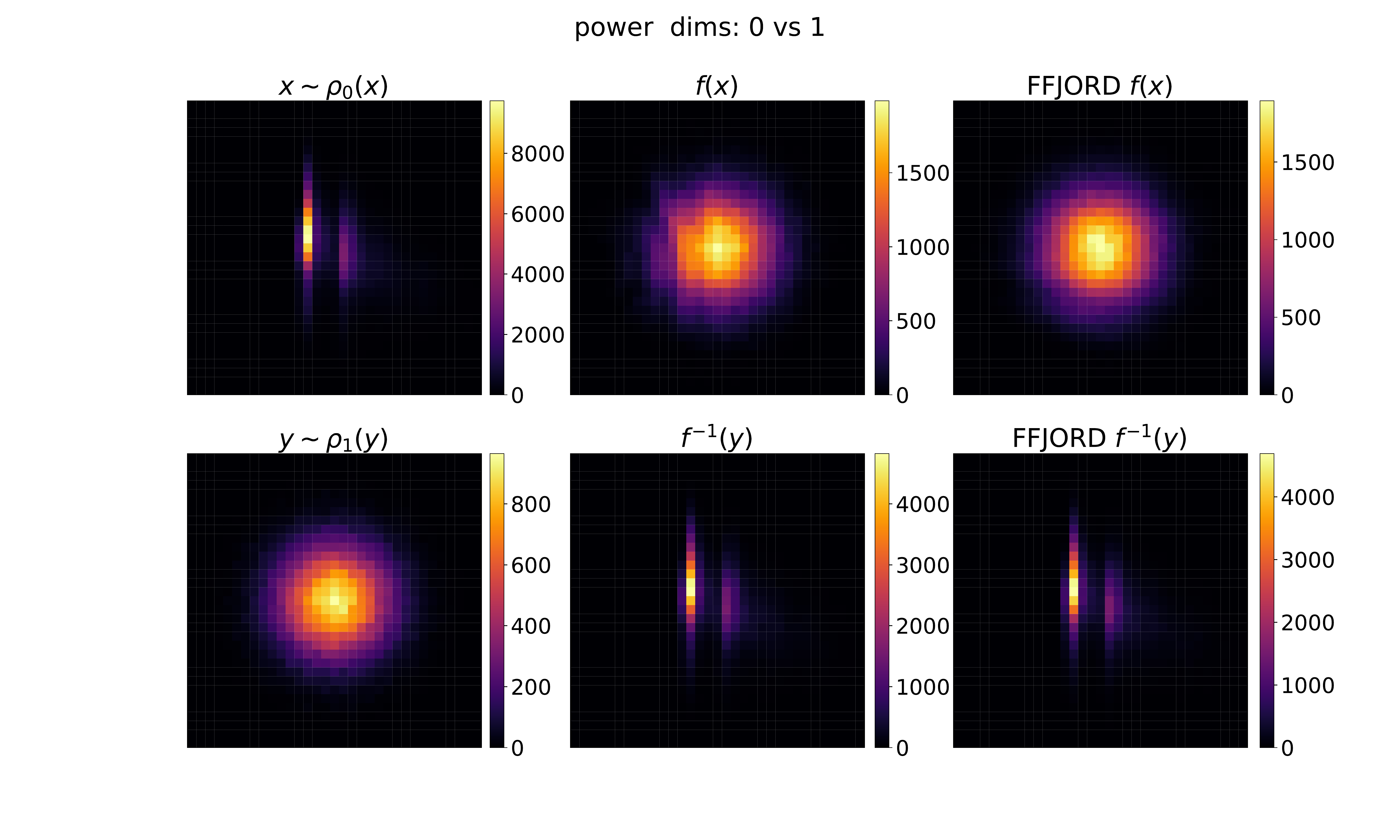}
      \includegraphics[clip, trim=2cm 2cm 1.6cm 0cm, width=0.48\linewidth]{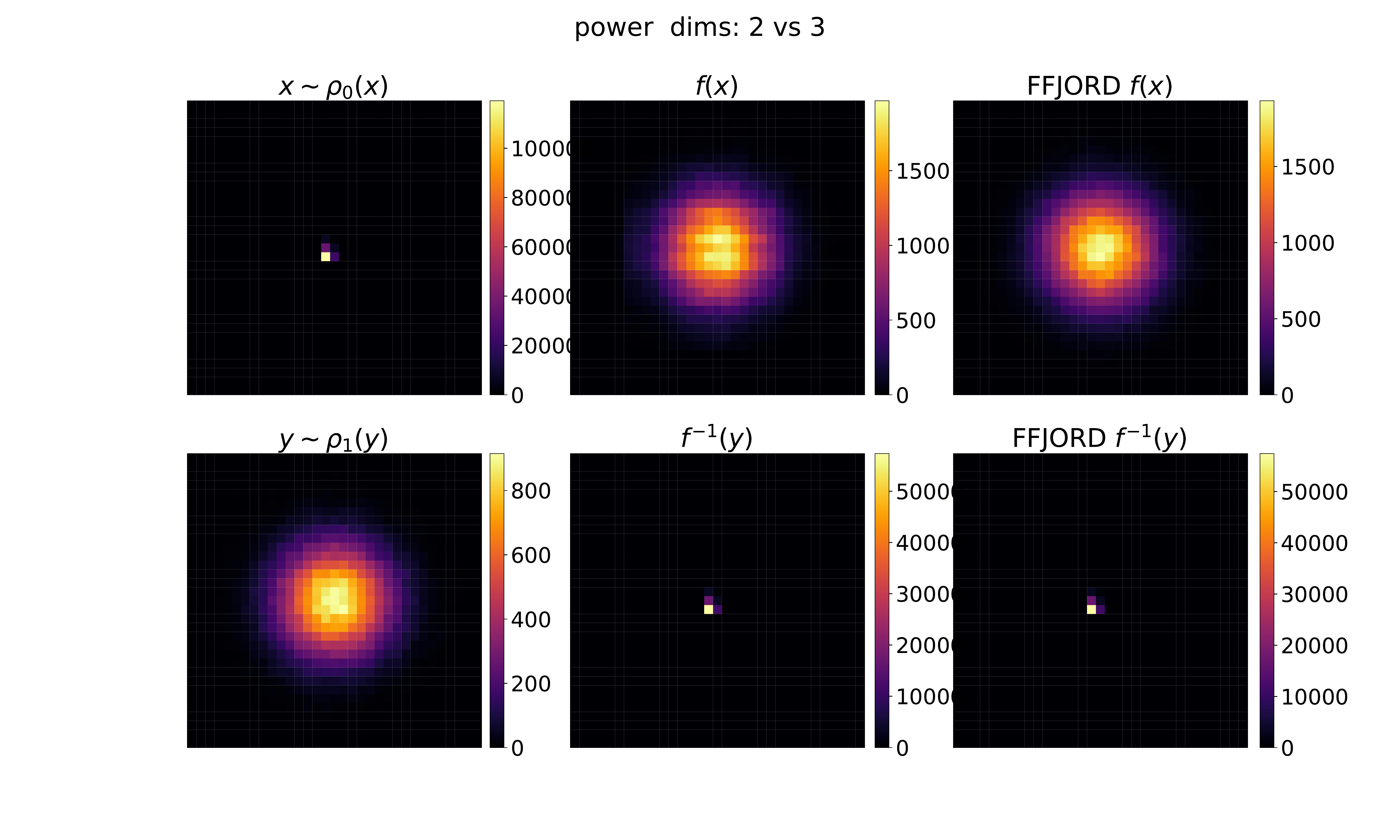} \\
      \includegraphics[clip, trim=2cm 2cm 1.6cm 0cm, width=0.48\linewidth]{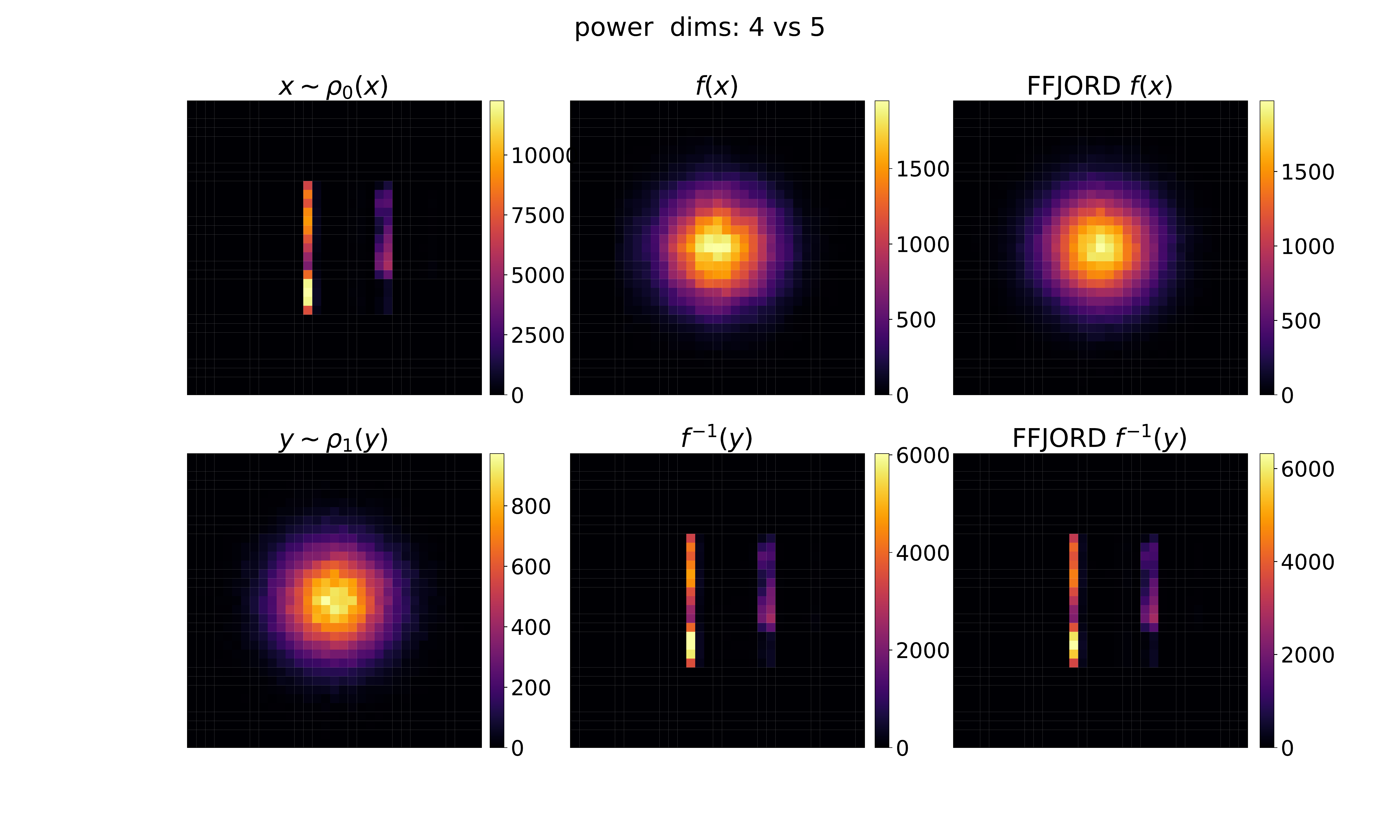}
     \caption{Model performance on \power{} test data.}
     \label{fig:power}
 \end{figure*}

 \begin{figure*}
     \centering
      \includegraphics[clip, trim=2cm 2cm 1.6cm 0cm, width=0.48\linewidth]{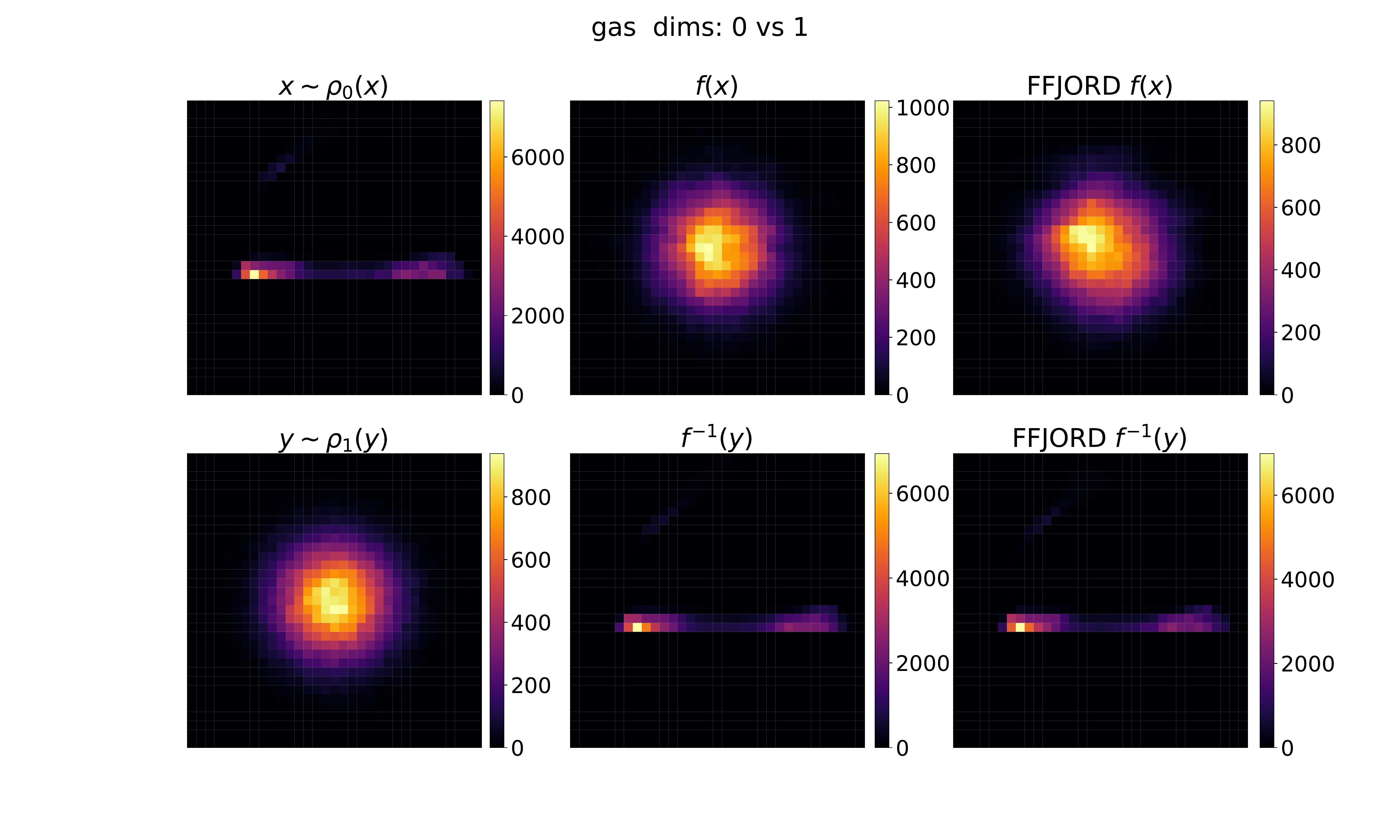}
      \includegraphics[clip, trim=2cm 2cm 1.6cm 0cm, width=0.48\linewidth]{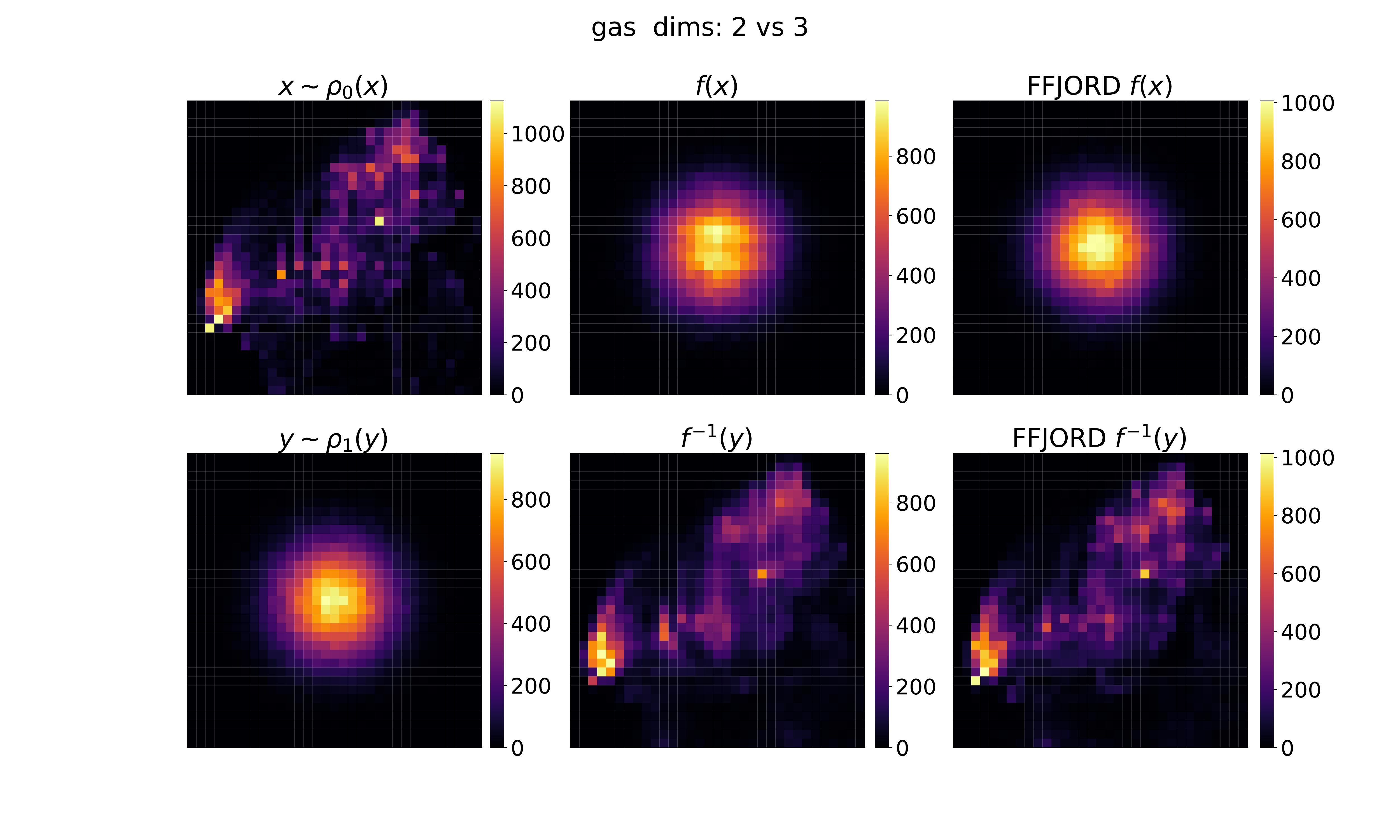} \\
      \includegraphics[clip, trim=2cm 2cm 1.6cm 0cm, width=0.48\linewidth]{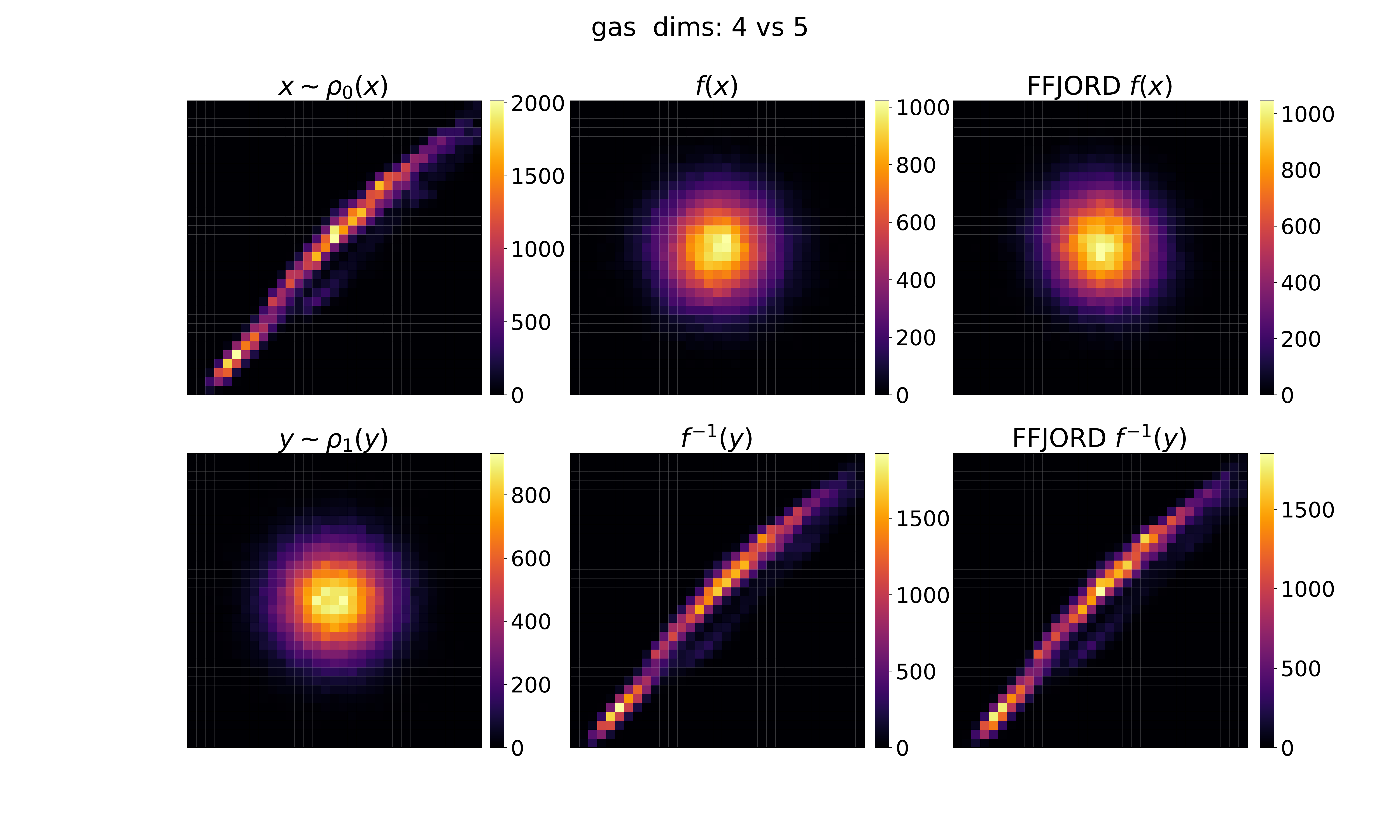}
      \includegraphics[clip, trim=2cm 2cm 1.6cm 0cm, width=0.48\linewidth]{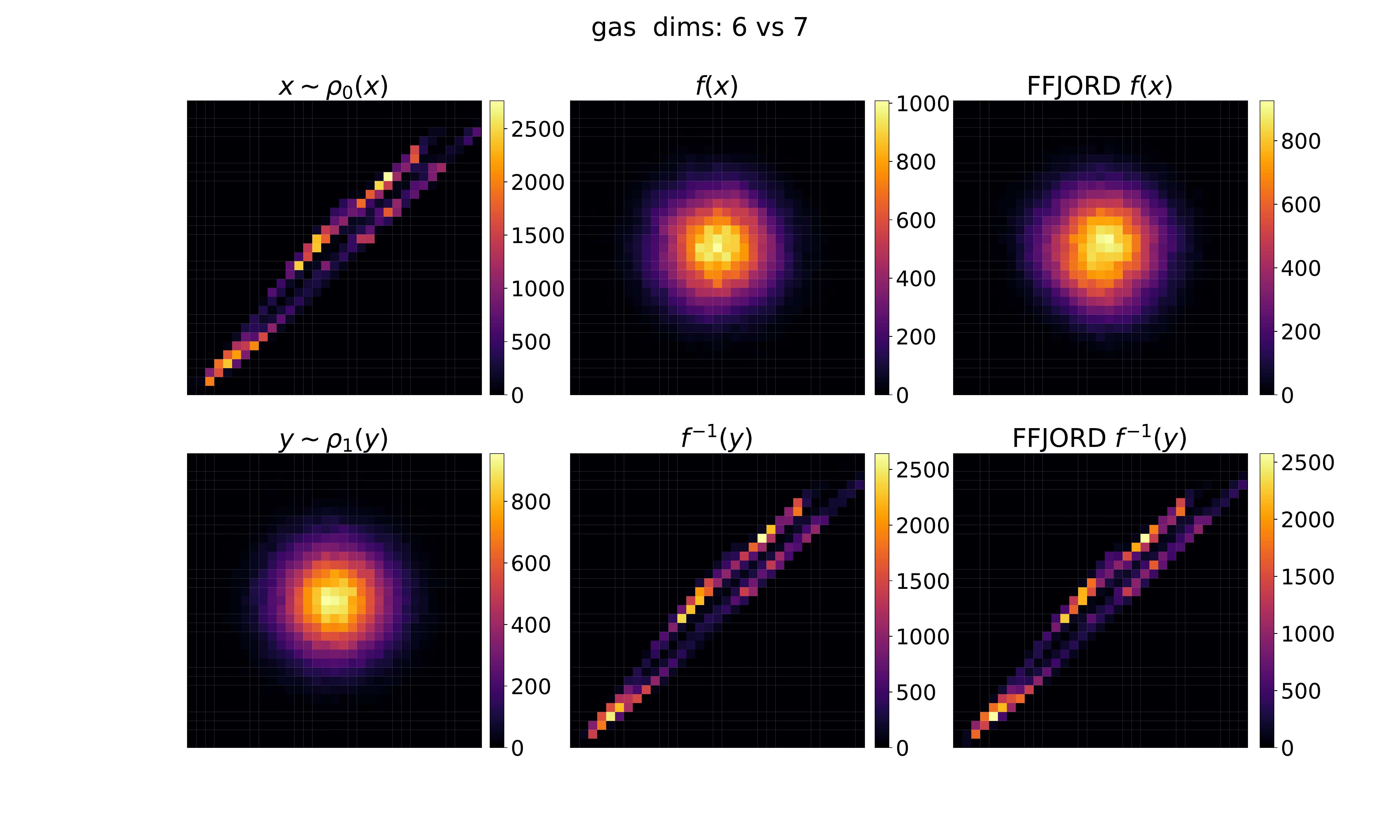}
     \caption{Model performance on \gas{} test data.}
     \label{fig:gas}
 \end{figure*}

 \begin{figure*}
     \centering
     \includegraphics[clip, trim=2cm 2.0cm 2cm 0.5cm, width=0.3\linewidth]
     	{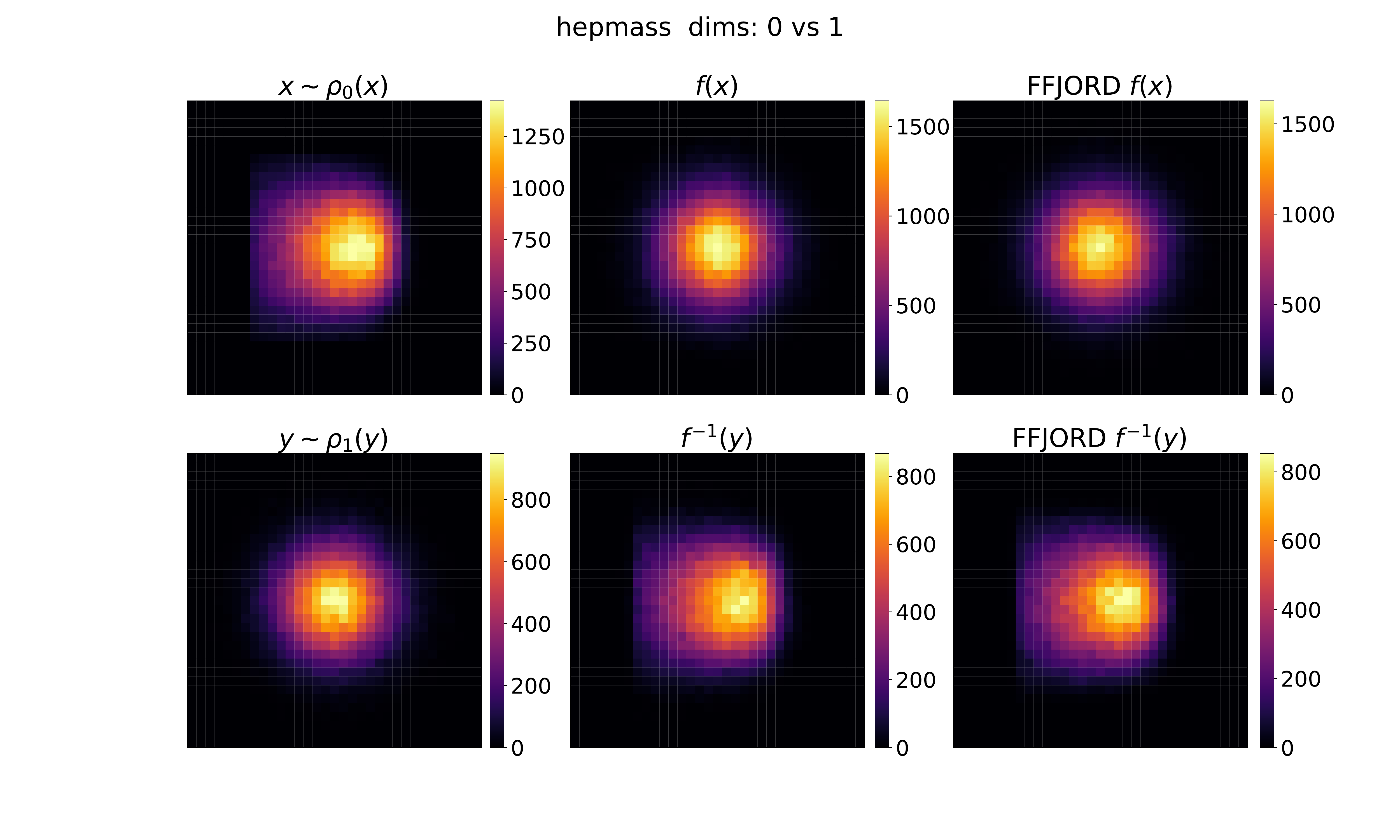}
     \includegraphics[clip, trim=2cm 2.0cm 2cm 0.5cm, width=0.3\linewidth]
     	{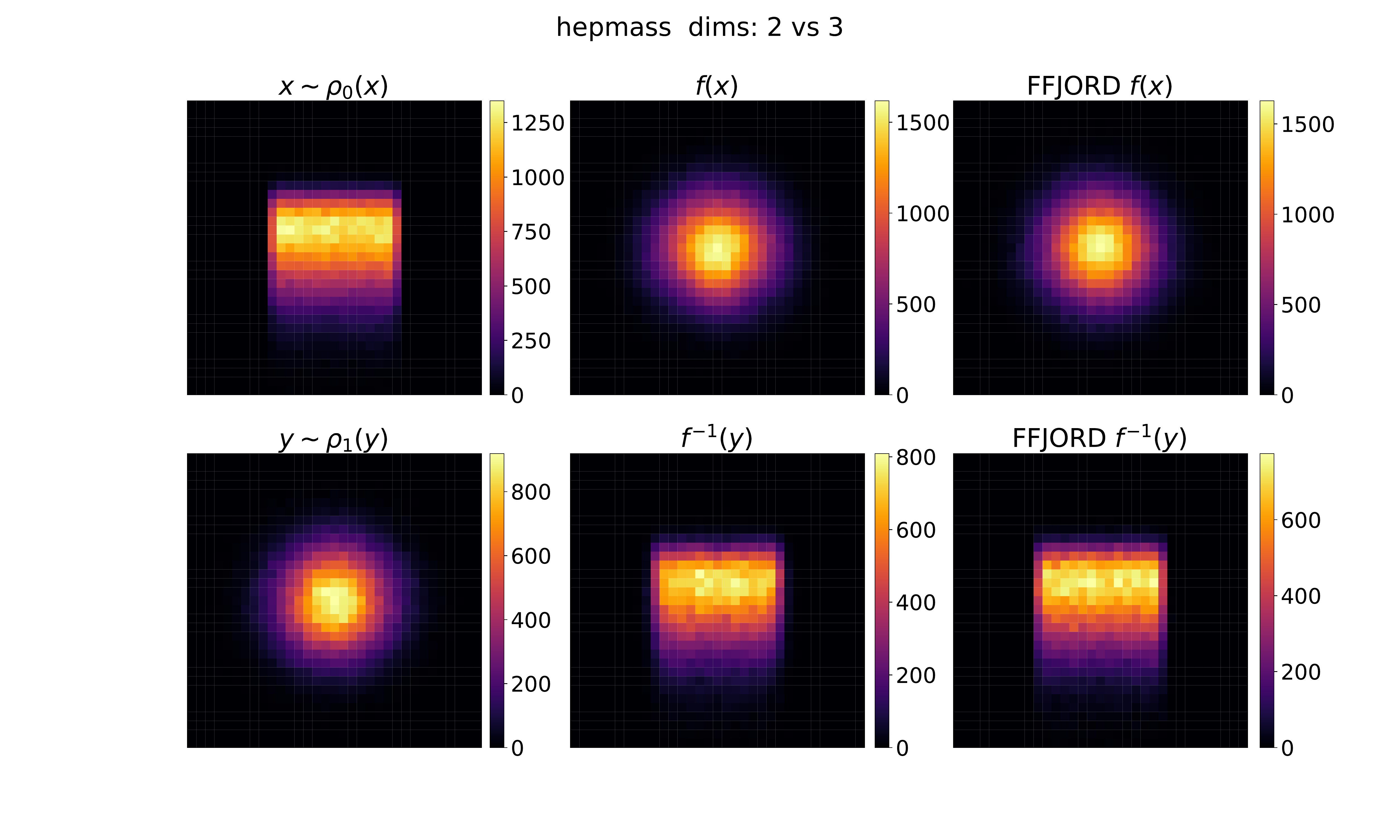}
     \includegraphics[clip, trim=2cm 2.0cm 2cm 0.5cm, width=0.3\linewidth]
     	{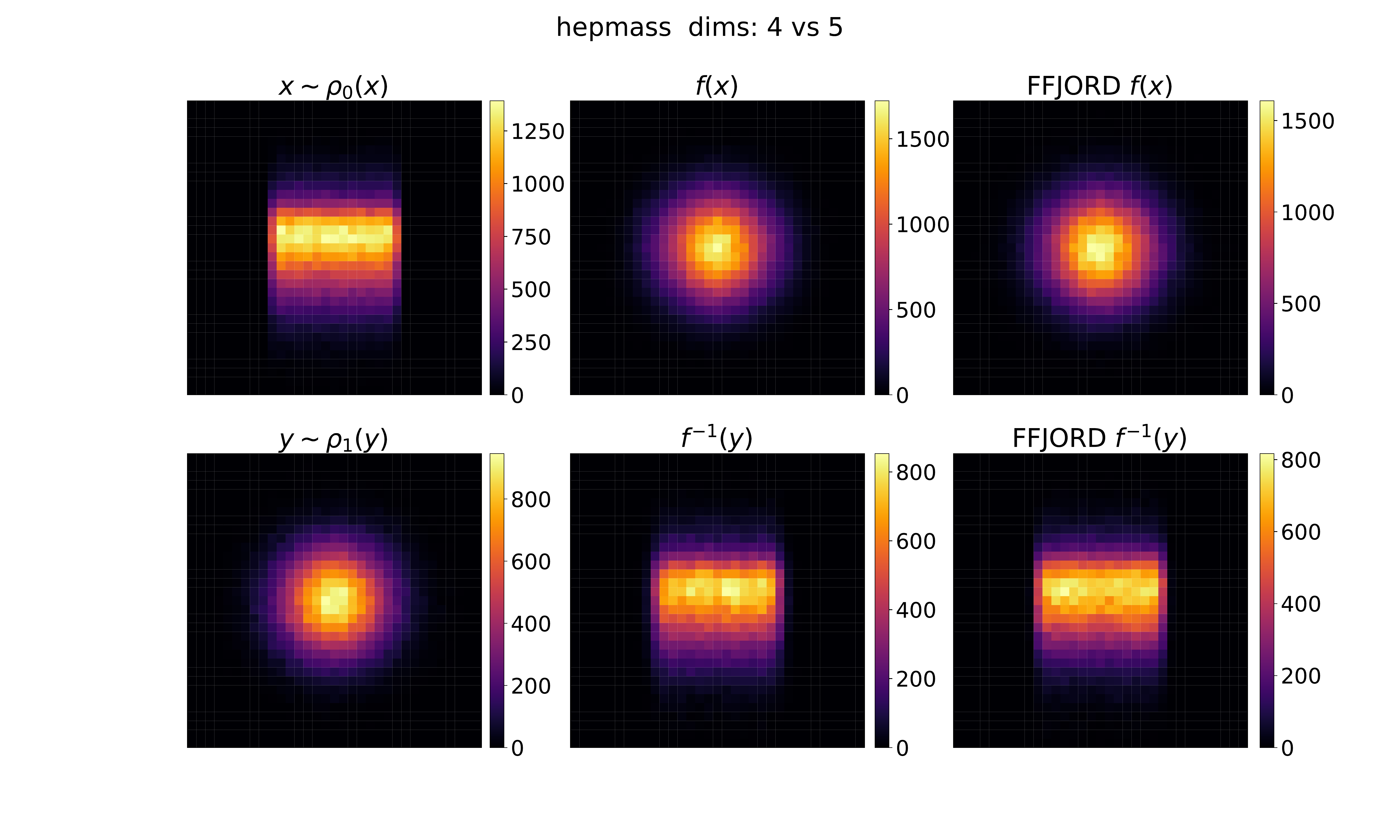}\\
     \includegraphics[clip, trim=2cm 2.0cm 2cm 0.5cm, width=0.3\linewidth]
         {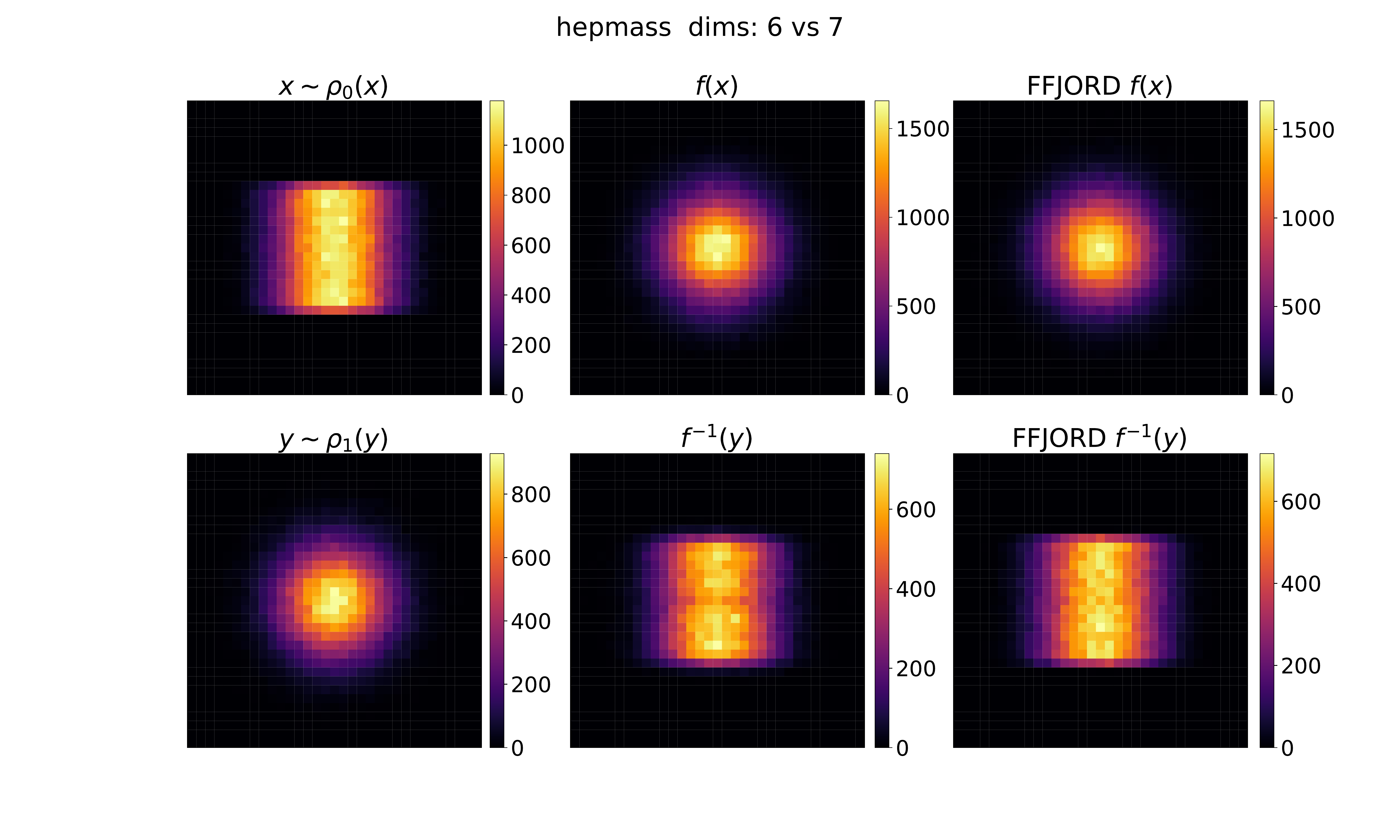}
     \includegraphics[clip, trim=2cm 2.0cm 2cm 0.5cm, width=0.3\linewidth]
         {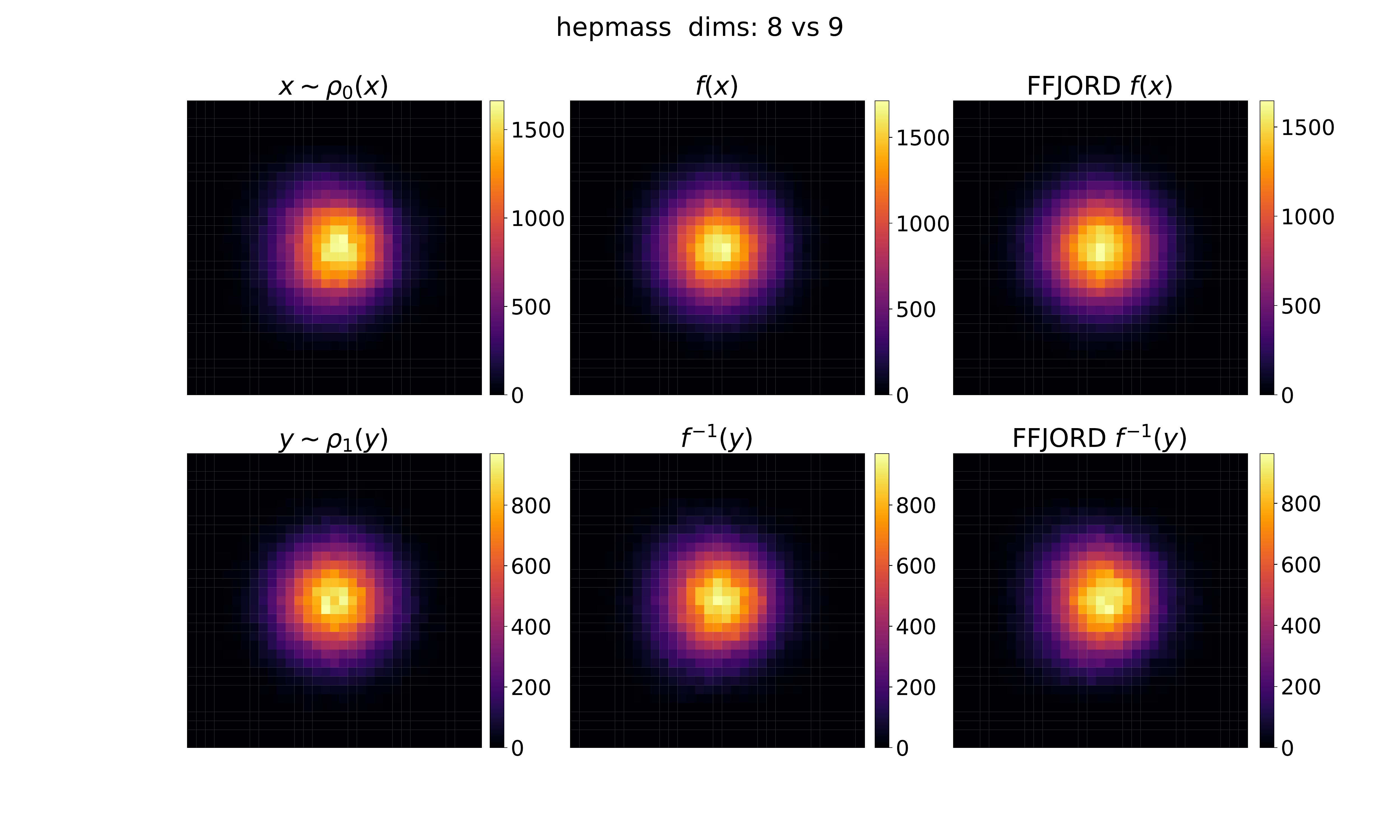}
     \includegraphics[clip, trim=2cm 2.0cm 2cm 0.5cm, width=0.3\linewidth]
         {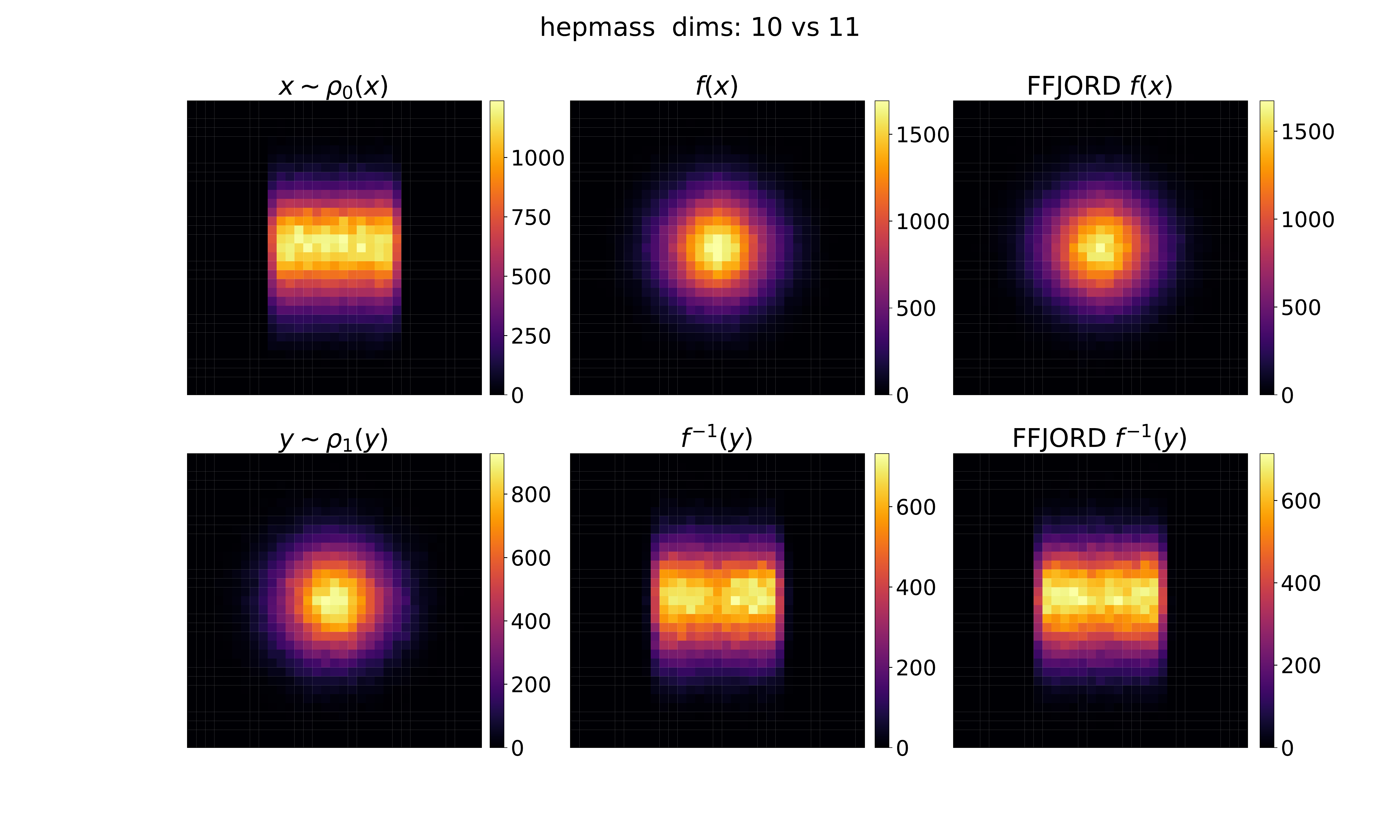}\\
     \includegraphics[clip, trim=2cm 2.0cm 2cm 0.5cm, width=0.3\linewidth]
         {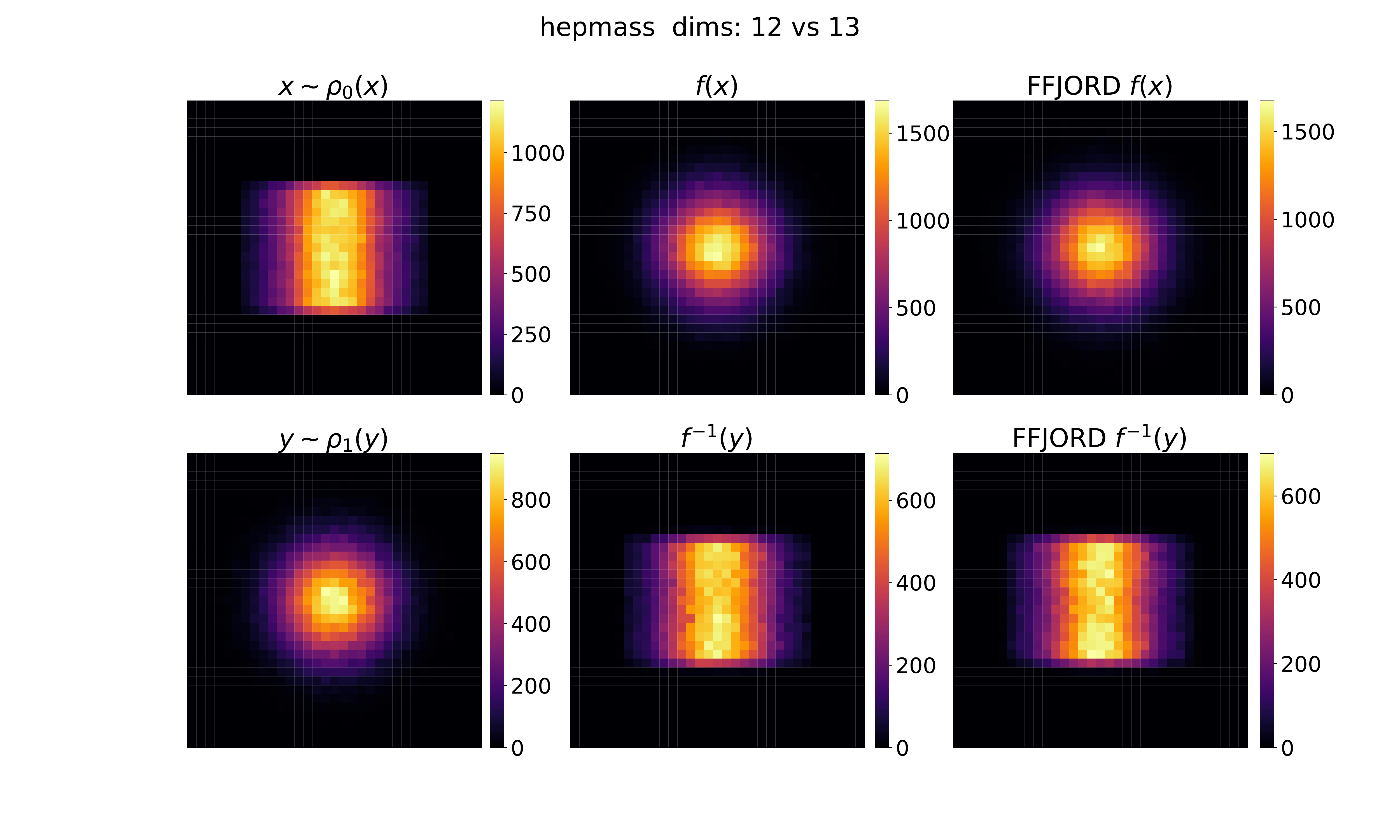}
     \includegraphics[clip, trim=2cm 2.0cm 2cm 0.5cm, width=0.3\linewidth]
         {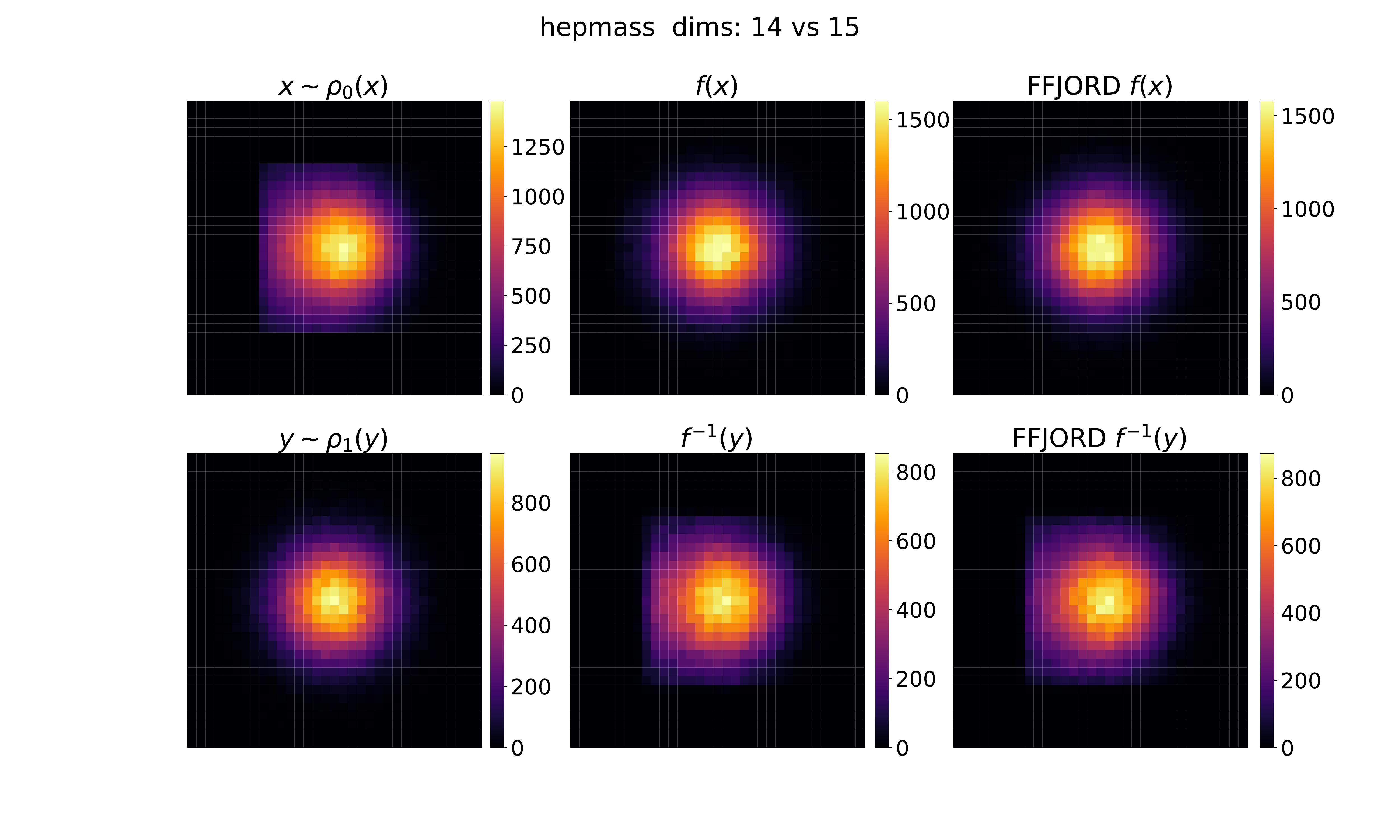}
     \includegraphics[clip, trim=2cm 2.0cm 2cm 0.5cm, width=0.3\linewidth]
         {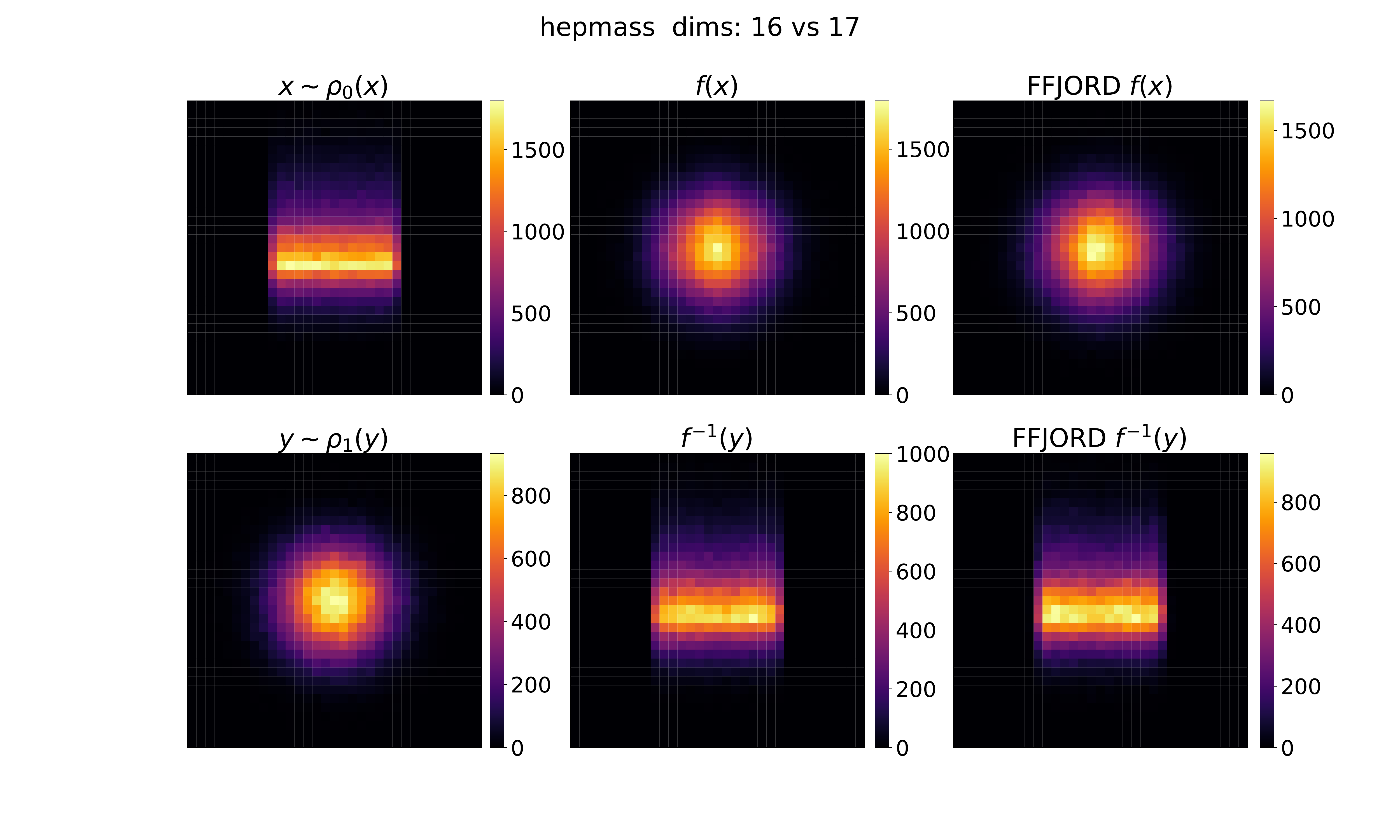}\\
     \includegraphics[clip, trim=2cm 2.0cm 2cm 0.5cm, width=0.3\linewidth]
         {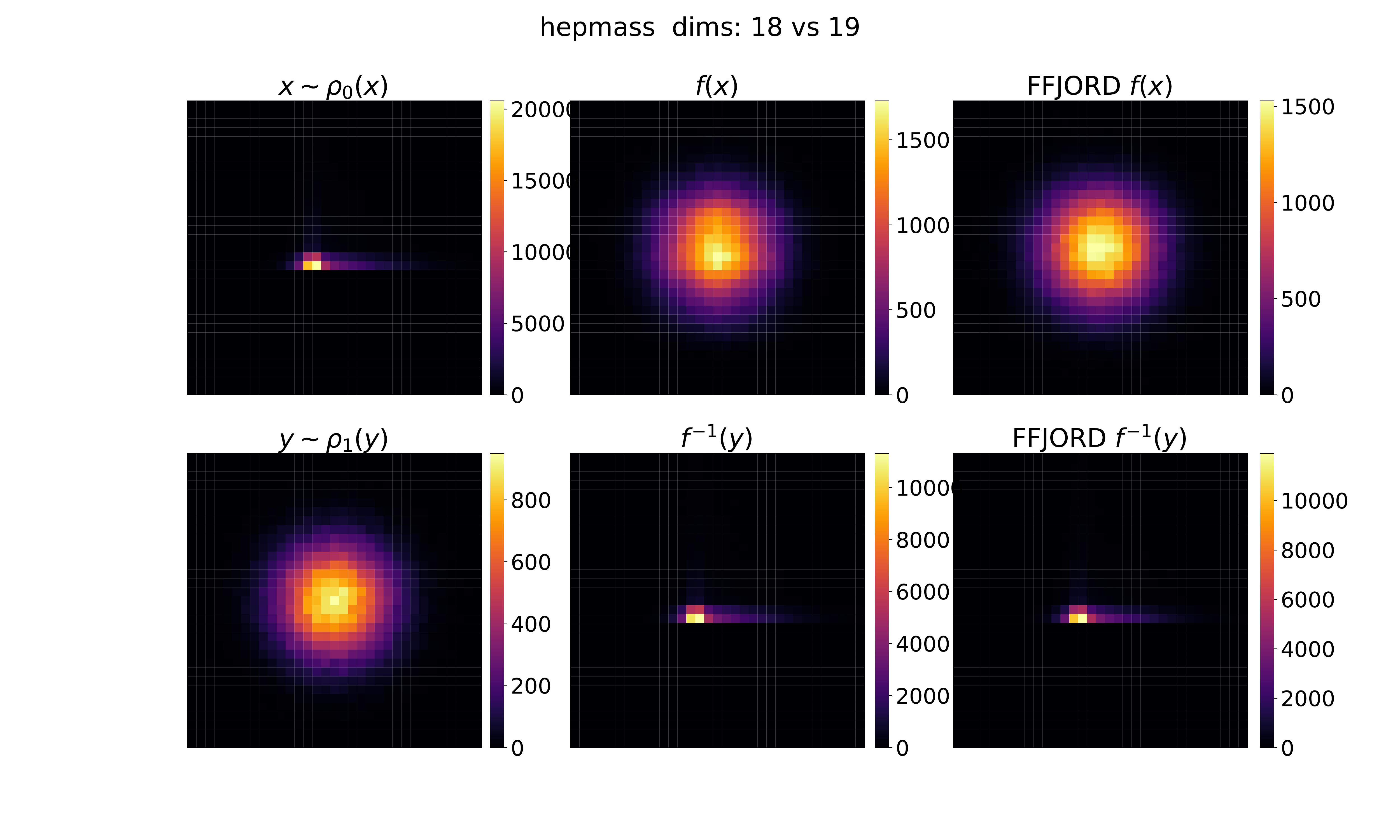}
     \caption{Model performance on the \hepmass{} test data.}
     \label{fig:hepmass}
 \end{figure*}

 \begin{figure*}
     \centering
     \includegraphics[clip, trim=2cm 2.0cm 2cm 0.5cm, width=0.3\linewidth]
     	{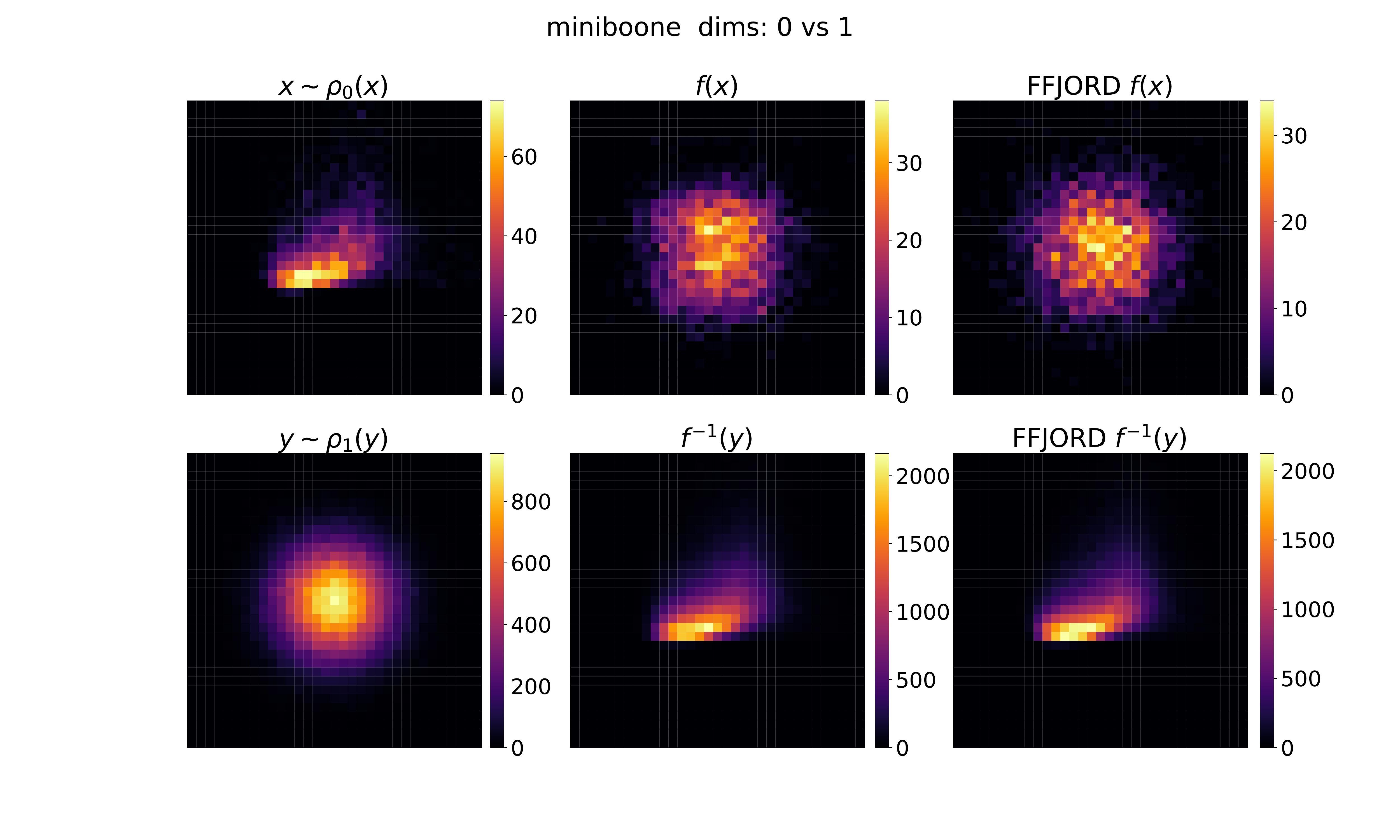} 
     \includegraphics[clip, trim=2cm 2.0cm 2cm 0.5cm, width=0.3\linewidth]
     	{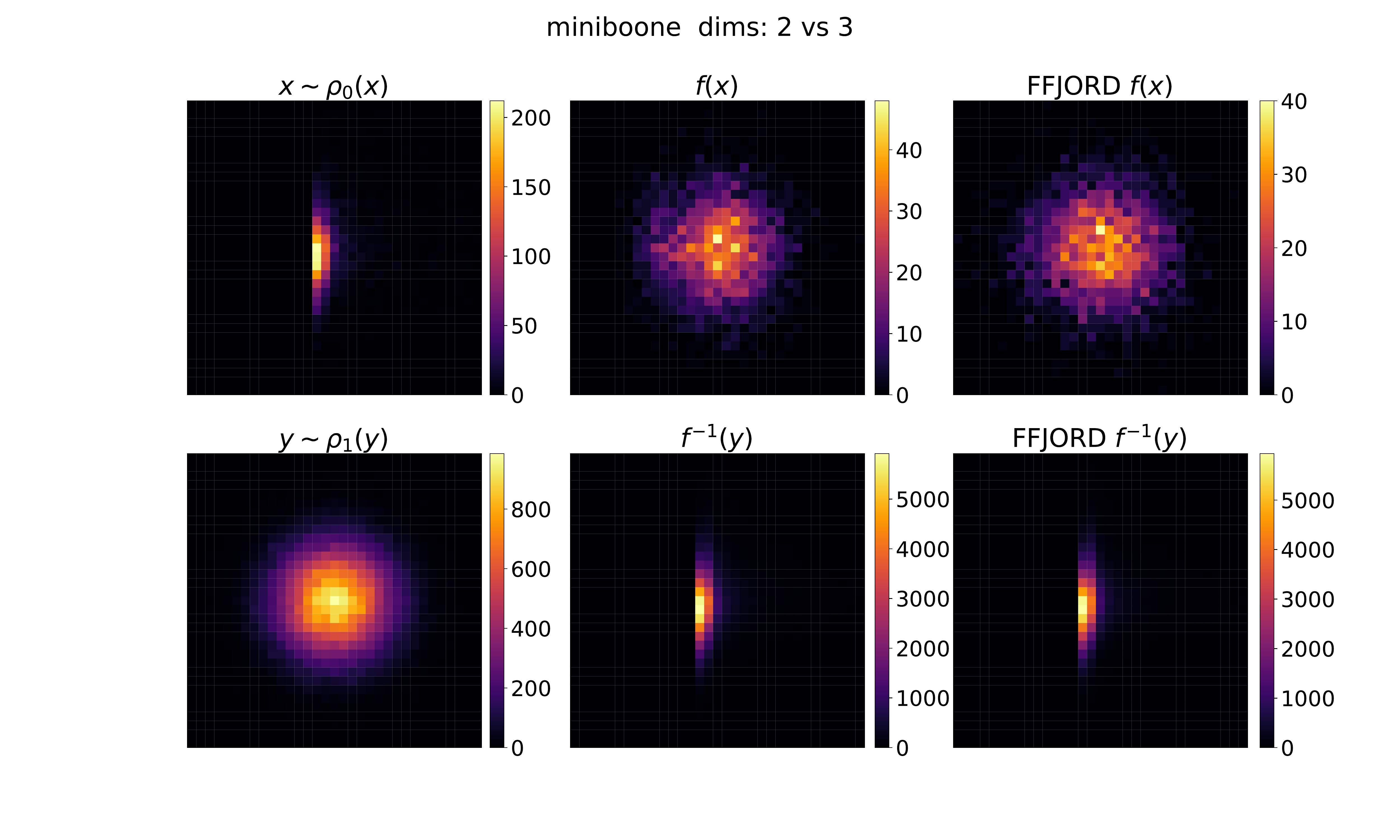}
     \includegraphics[clip, trim=2cm 2.0cm 2cm 0.5cm, width=0.3\linewidth]
     	{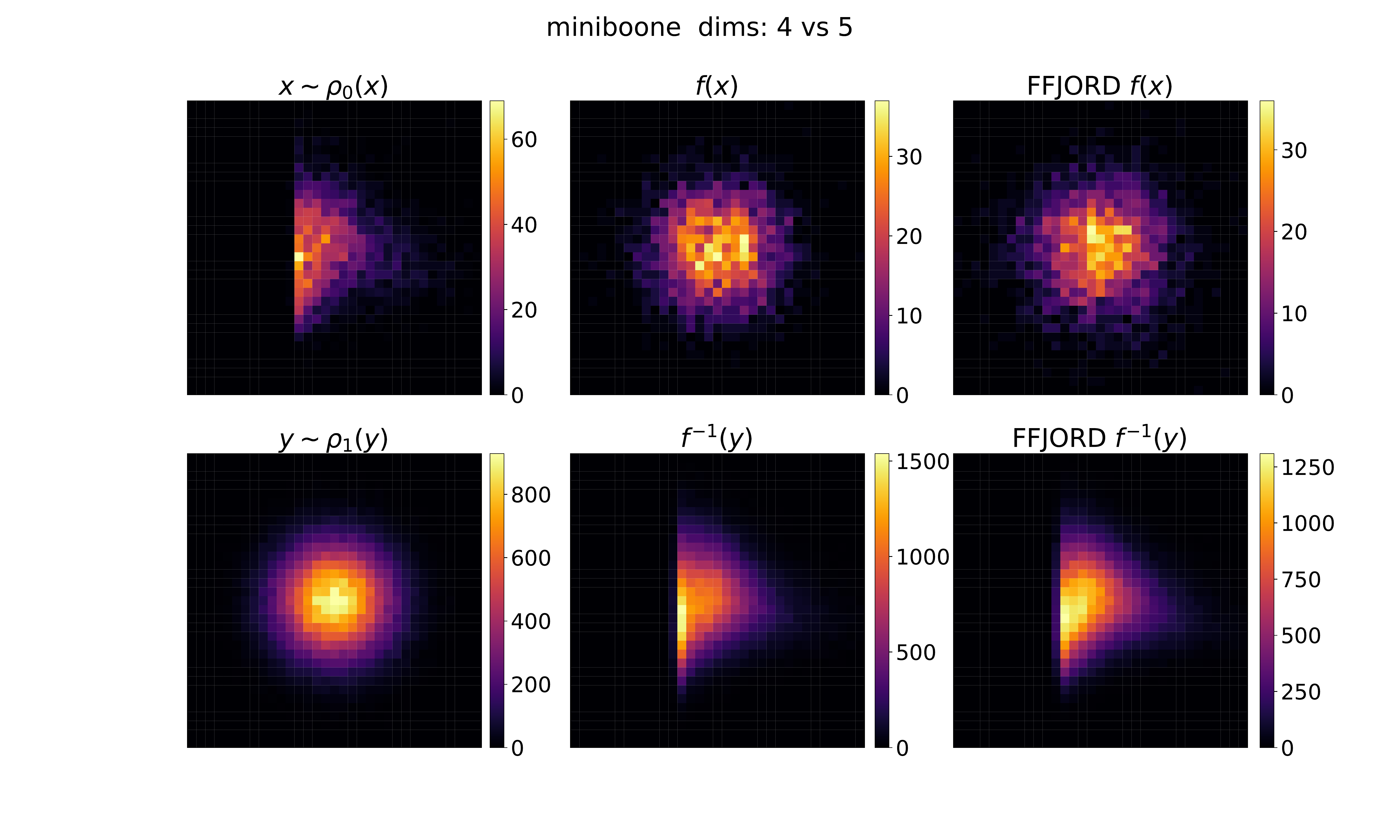}\\
     \includegraphics[clip, trim=2cm 2.0cm 2cm 0.5cm, width=0.3\linewidth]
         {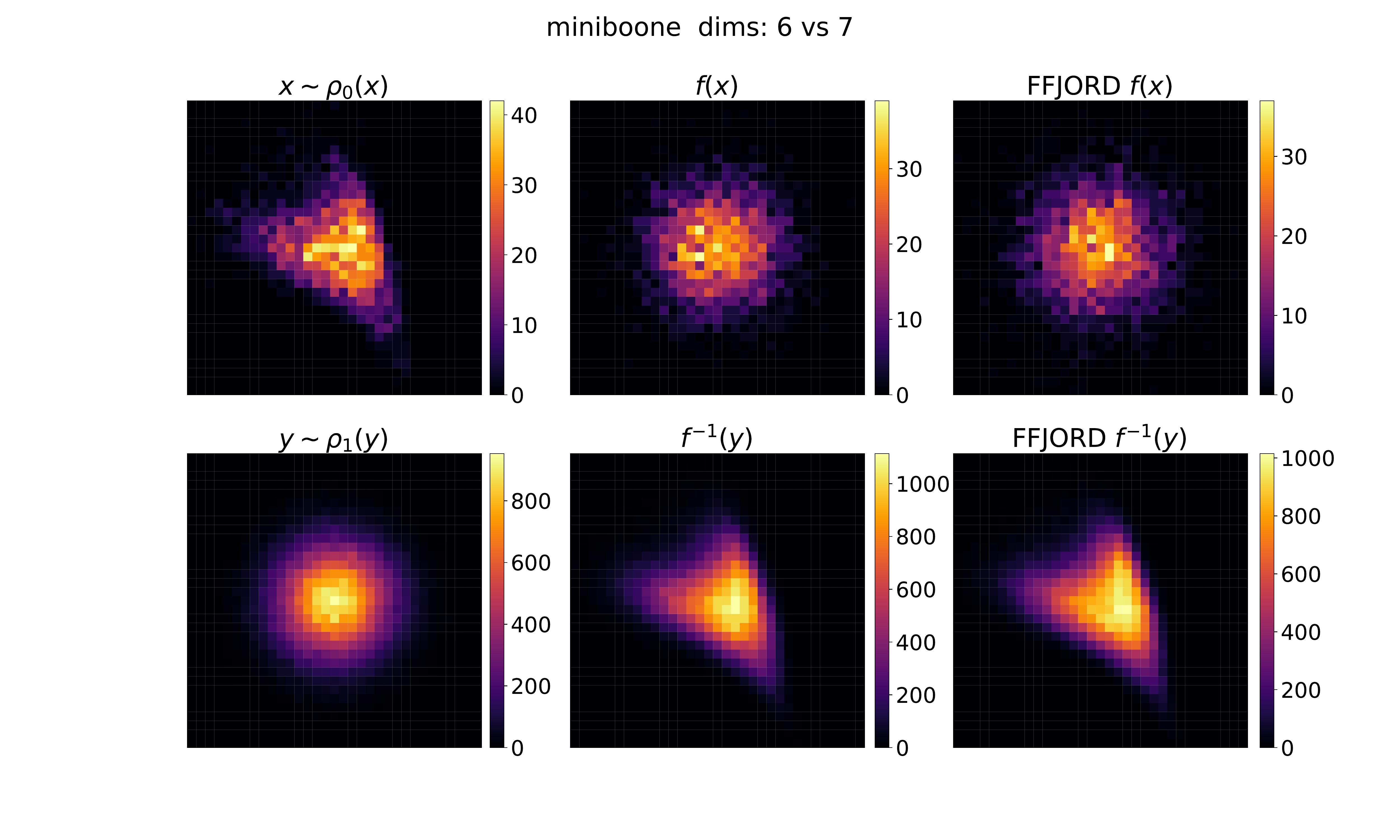}
     \includegraphics[clip, trim=2cm 2.0cm 2cm 0.5cm, width=0.3\linewidth]
         {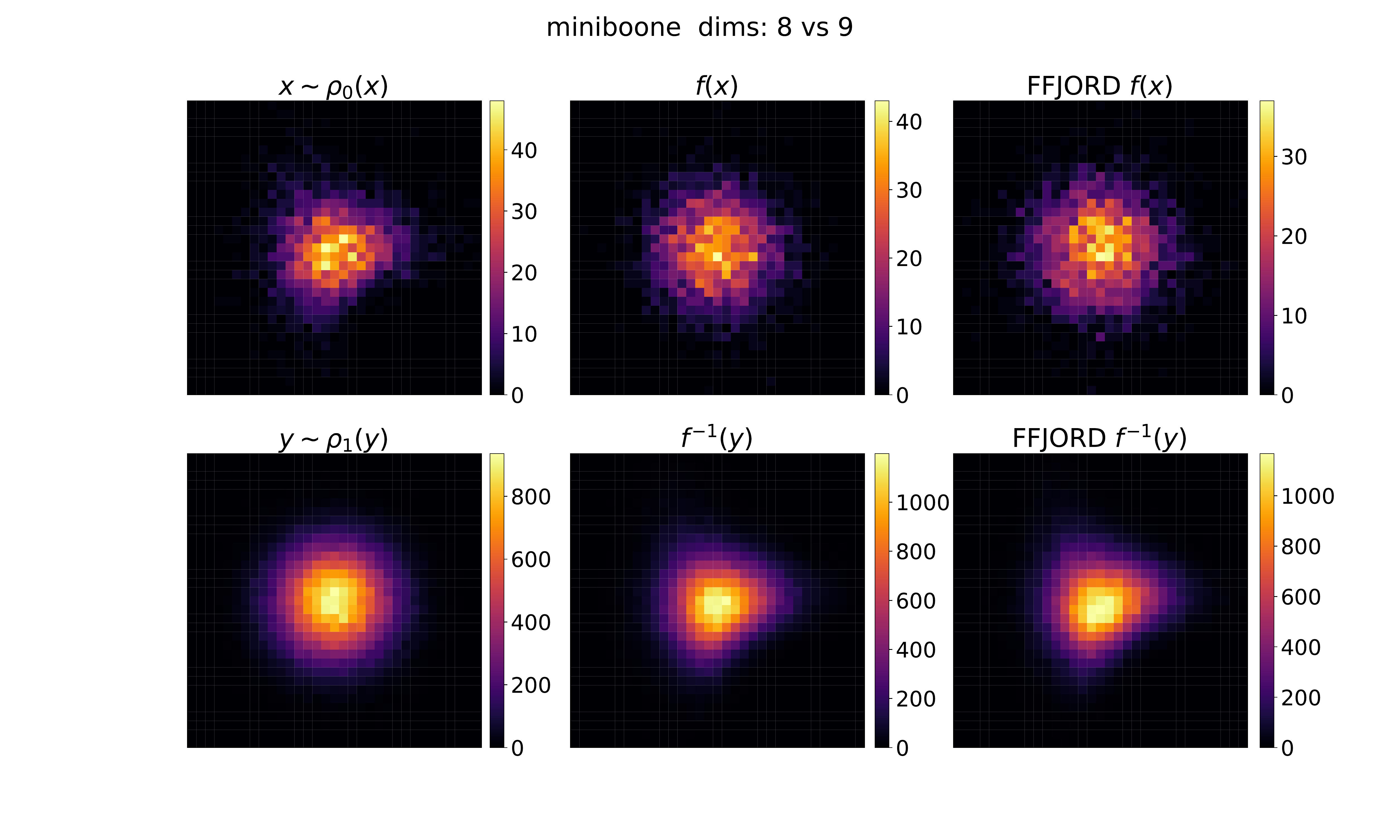}
     \includegraphics[clip, trim=2cm 2.0cm 2cm 0.5cm, width=0.3\linewidth]
         {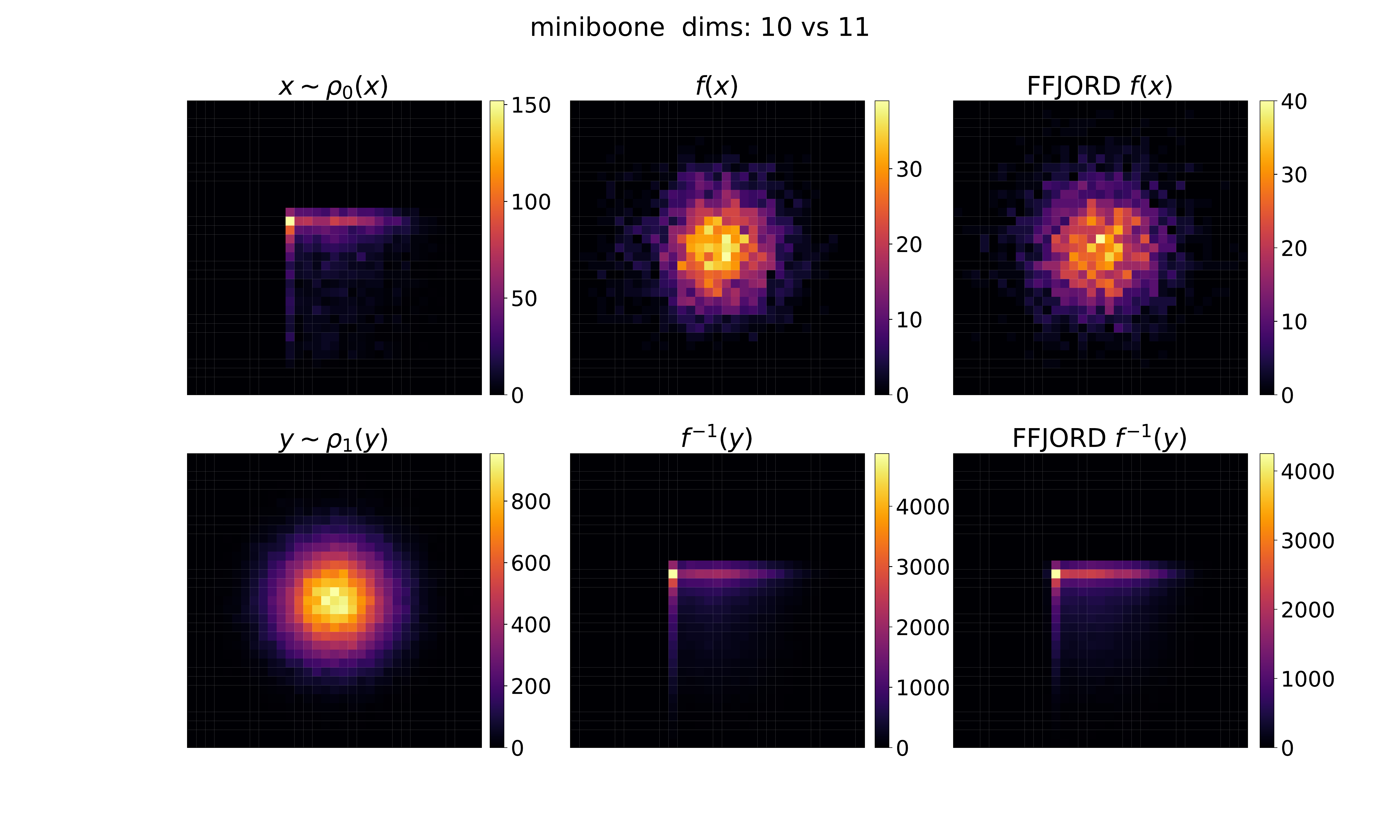}\\
     \includegraphics[clip, trim=2cm 2.0cm 2cm 0.5cm, width=0.3\linewidth]
         {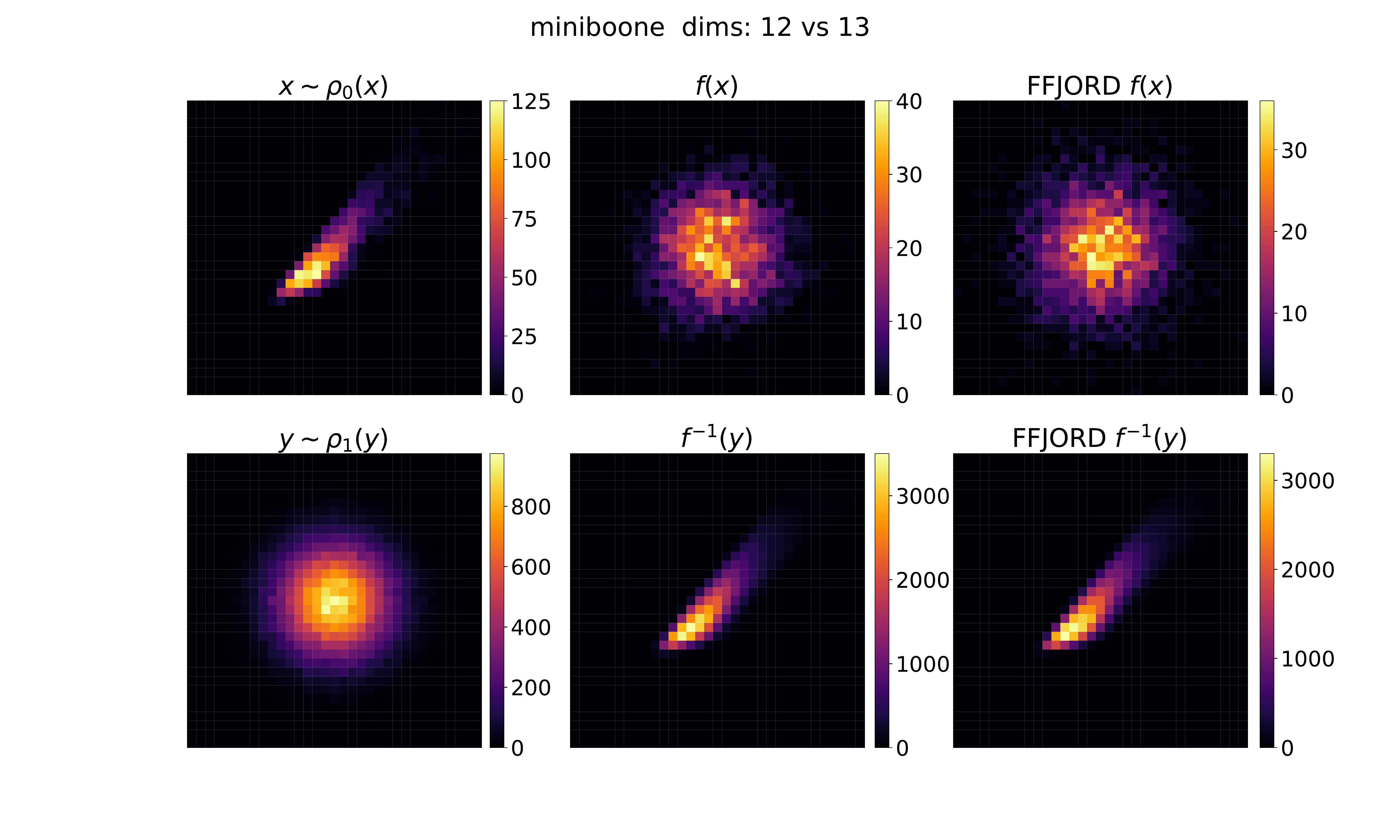}
     \includegraphics[clip, trim=2cm 2.0cm 2cm 0.5cm, width=0.3\linewidth]
         {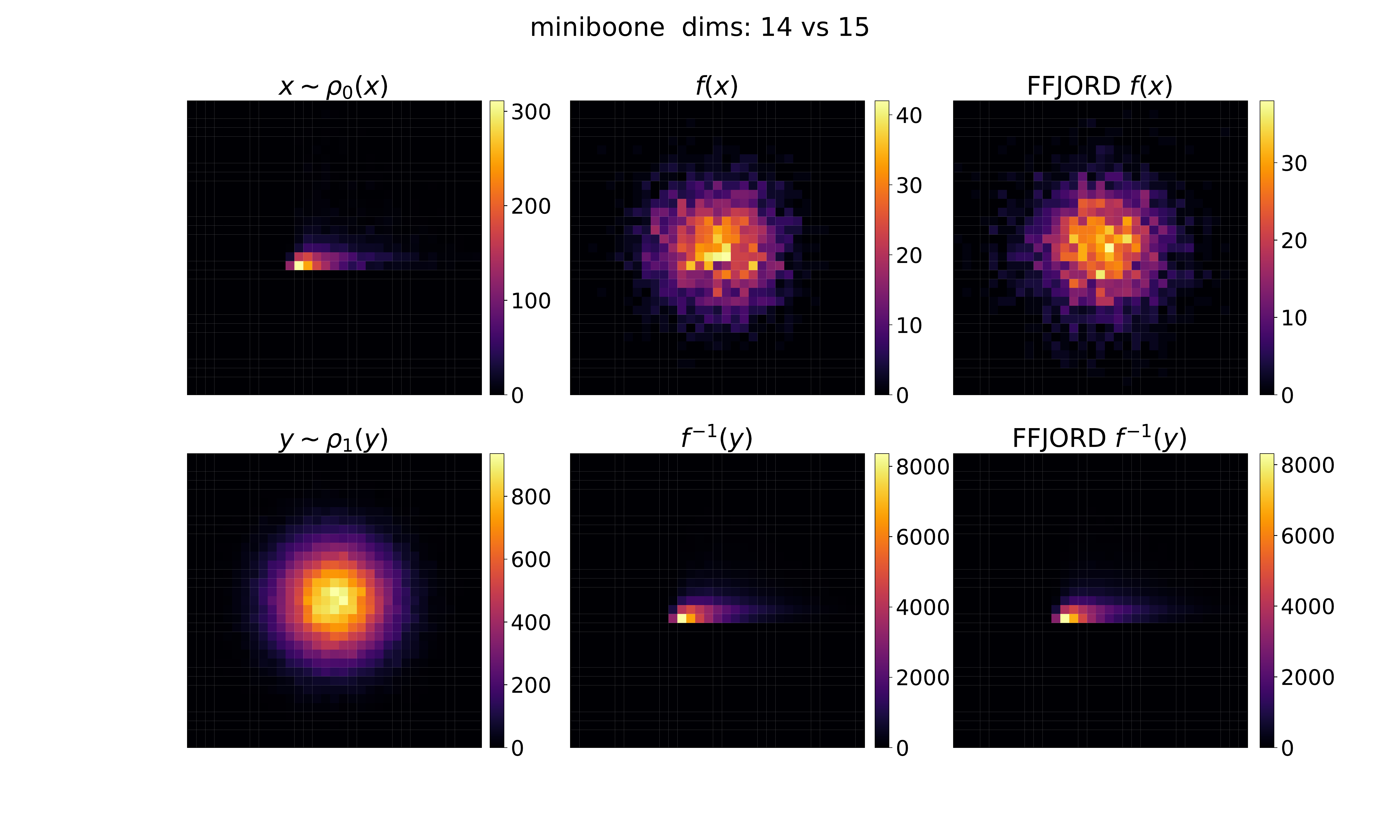}
     \includegraphics[clip, trim=2cm 2.0cm 2cm 0.5cm, width=0.3\linewidth]
         {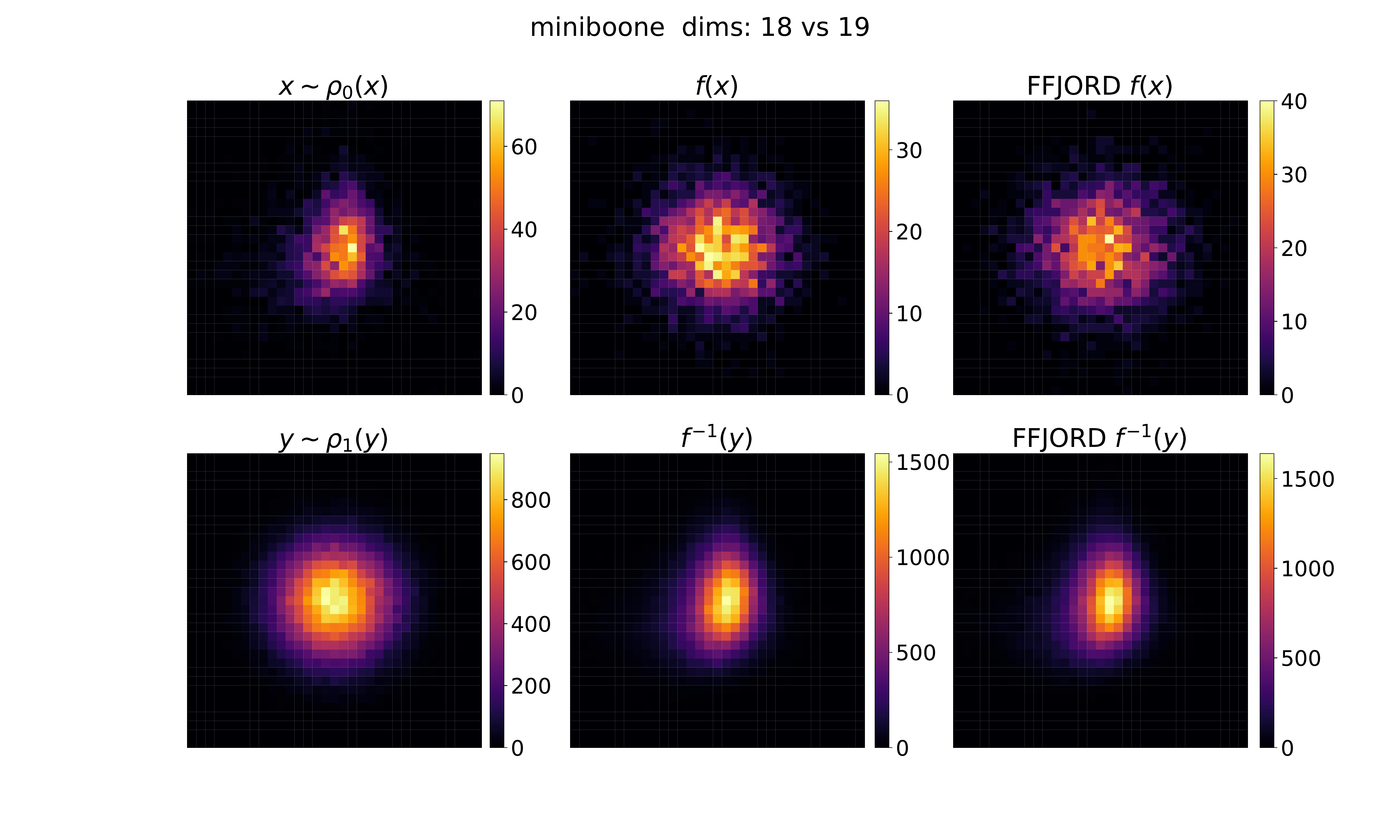}\\
     \includegraphics[clip, trim=2cm 2.0cm 2cm 0.5cm, width=0.3\linewidth]
         {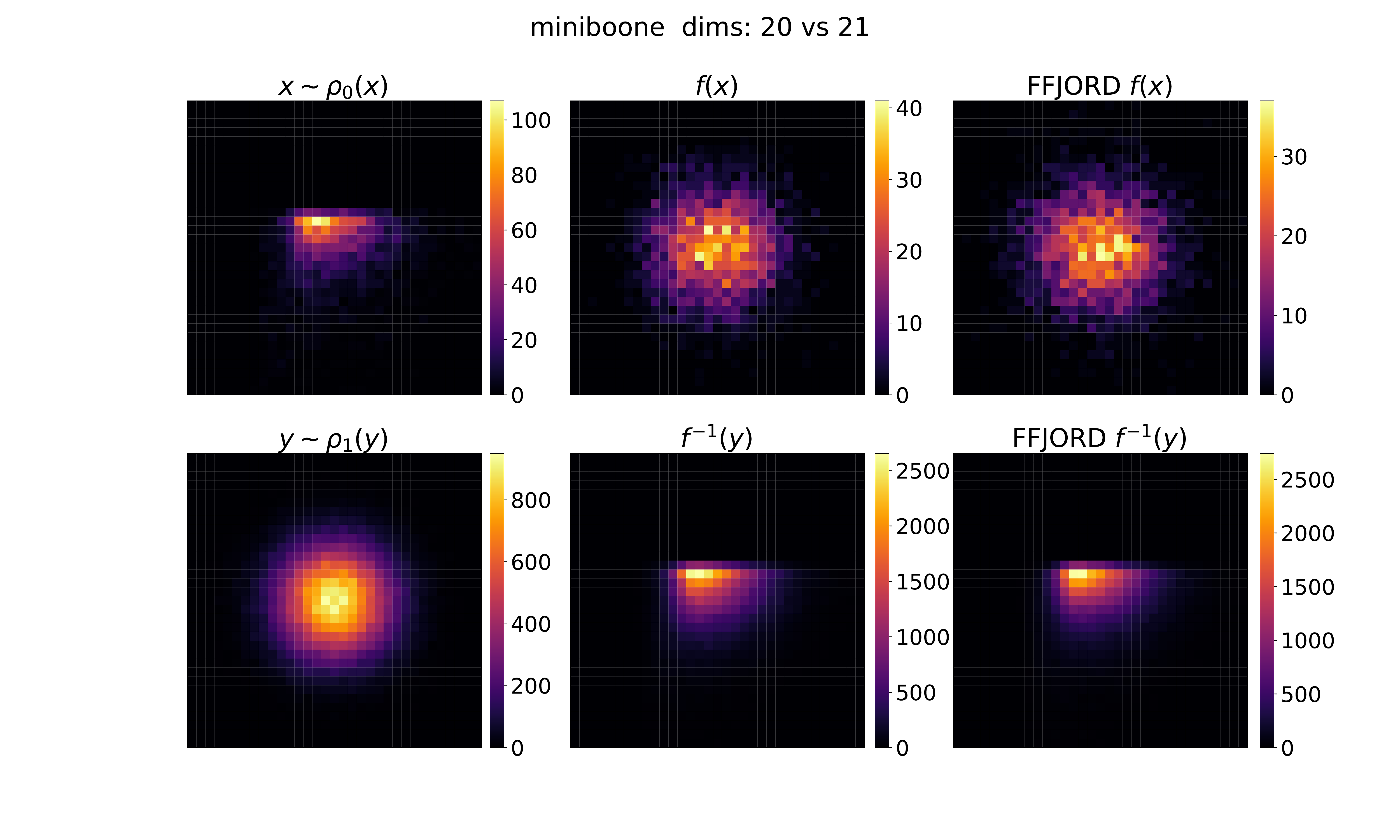}
     \includegraphics[clip, trim=2cm 2.0cm 2cm 0.5cm, width=0.3\linewidth]
         {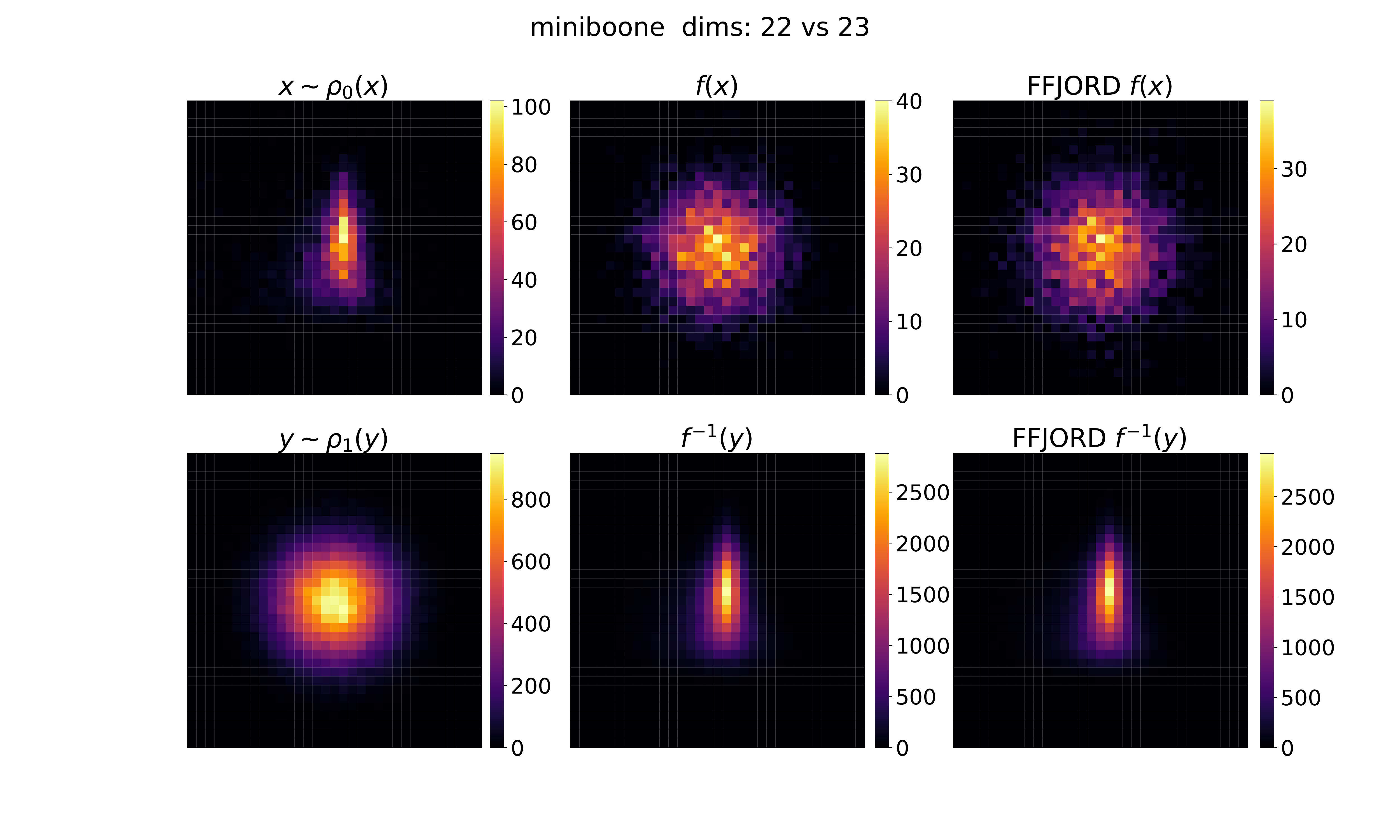}
     \includegraphics[clip, trim=2cm 2.0cm 2cm 0.5cm, width=0.3\linewidth]
         {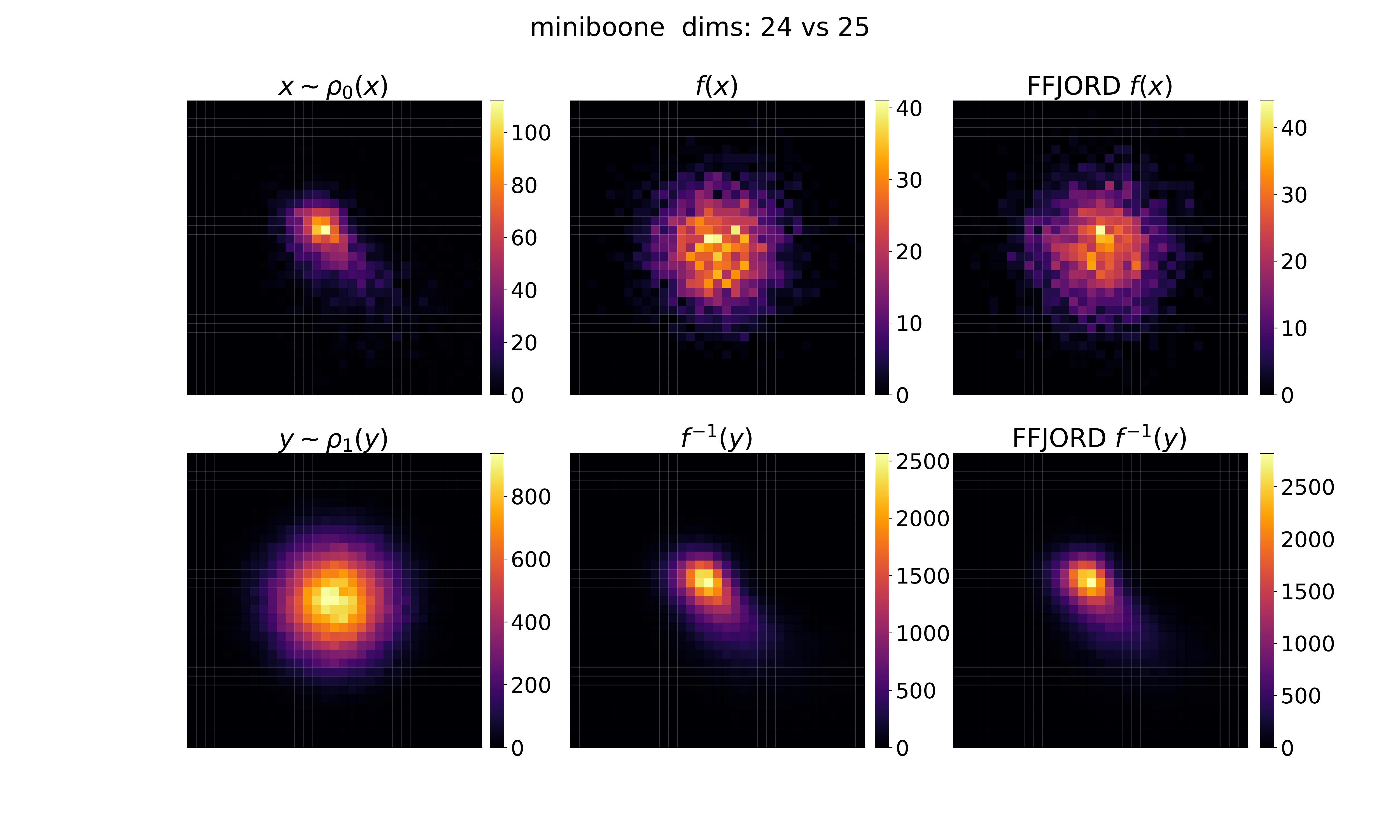}\\
     \includegraphics[clip, trim=2cm 2.0cm 2cm 0.5cm, width=0.3\linewidth]
         {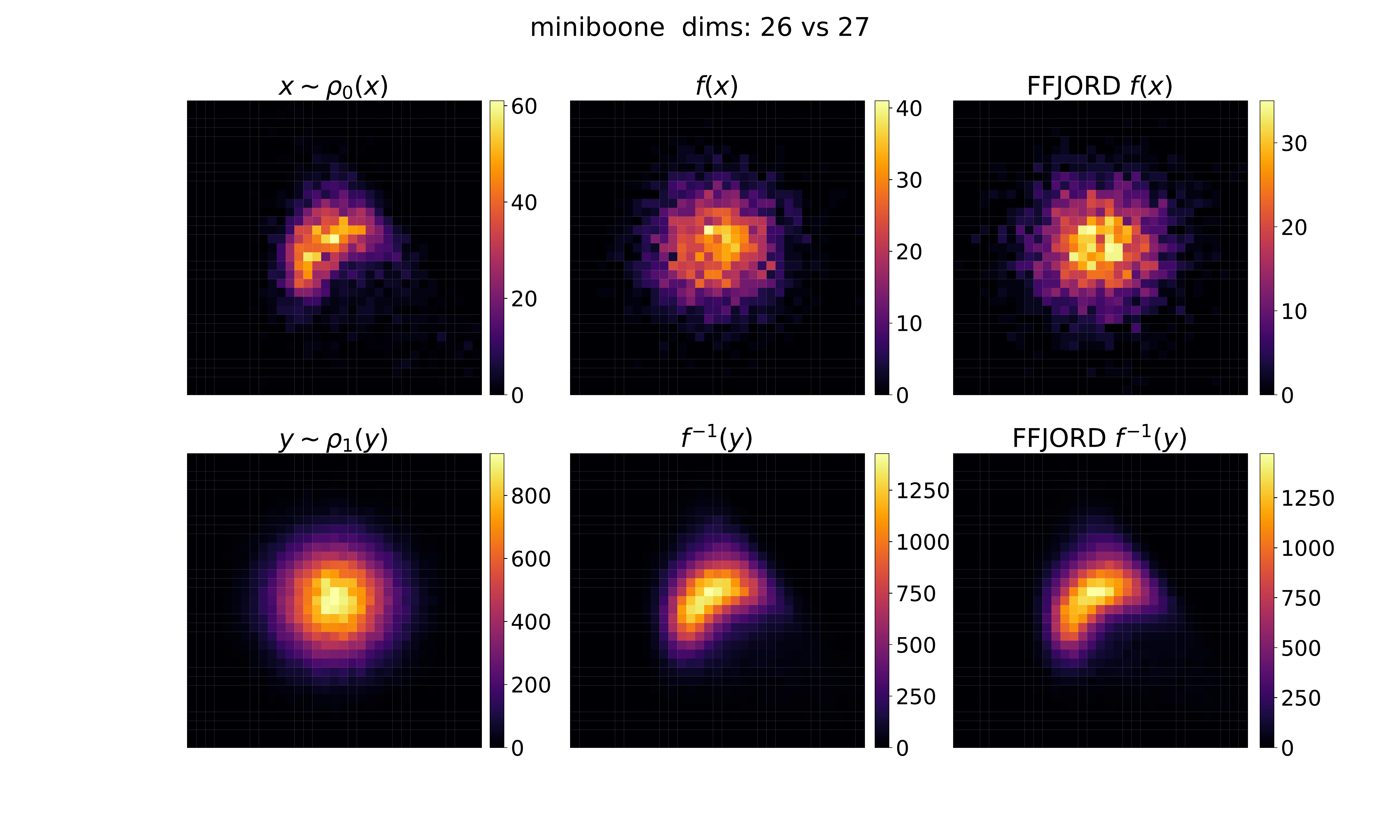}
     \includegraphics[clip, trim=2cm 2.0cm 2cm 0.5cm, width=0.3\linewidth]
         {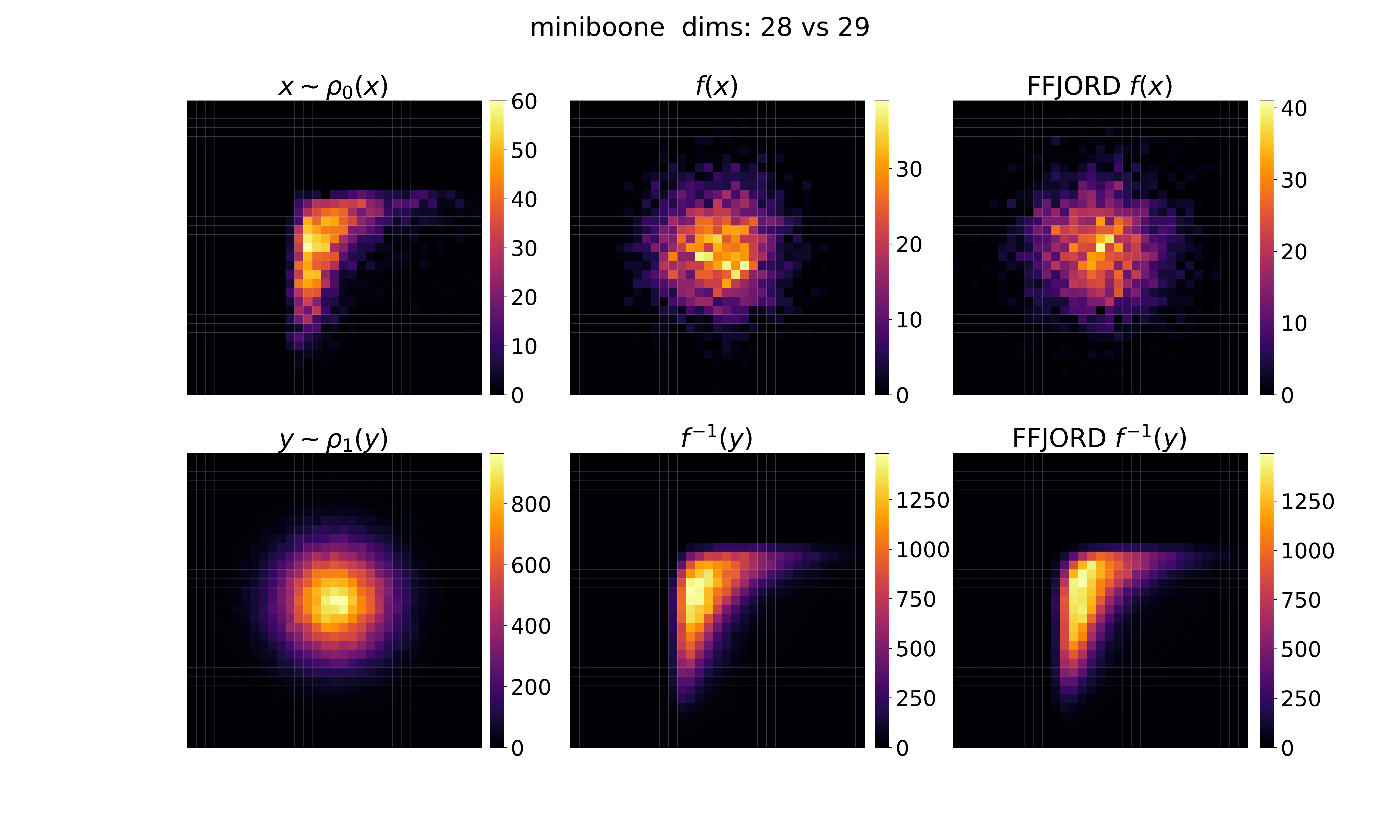}        
     \includegraphics[clip, trim=2cm 2.0cm 2cm 0.5cm, width=0.3\linewidth]
         {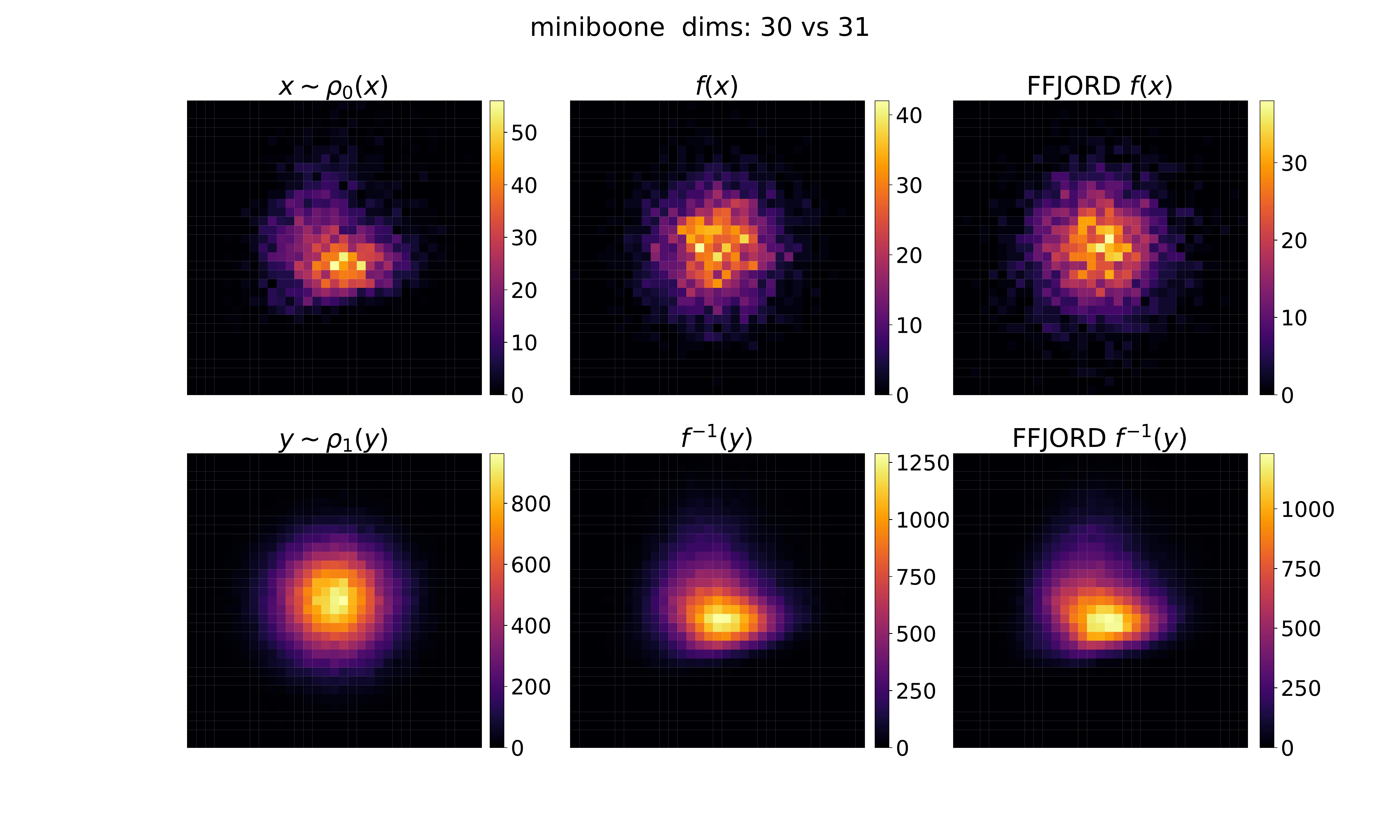} \\
     \includegraphics[clip, trim=2cm 2.0cm 2cm 0.5cm, width=0.3\linewidth]
         {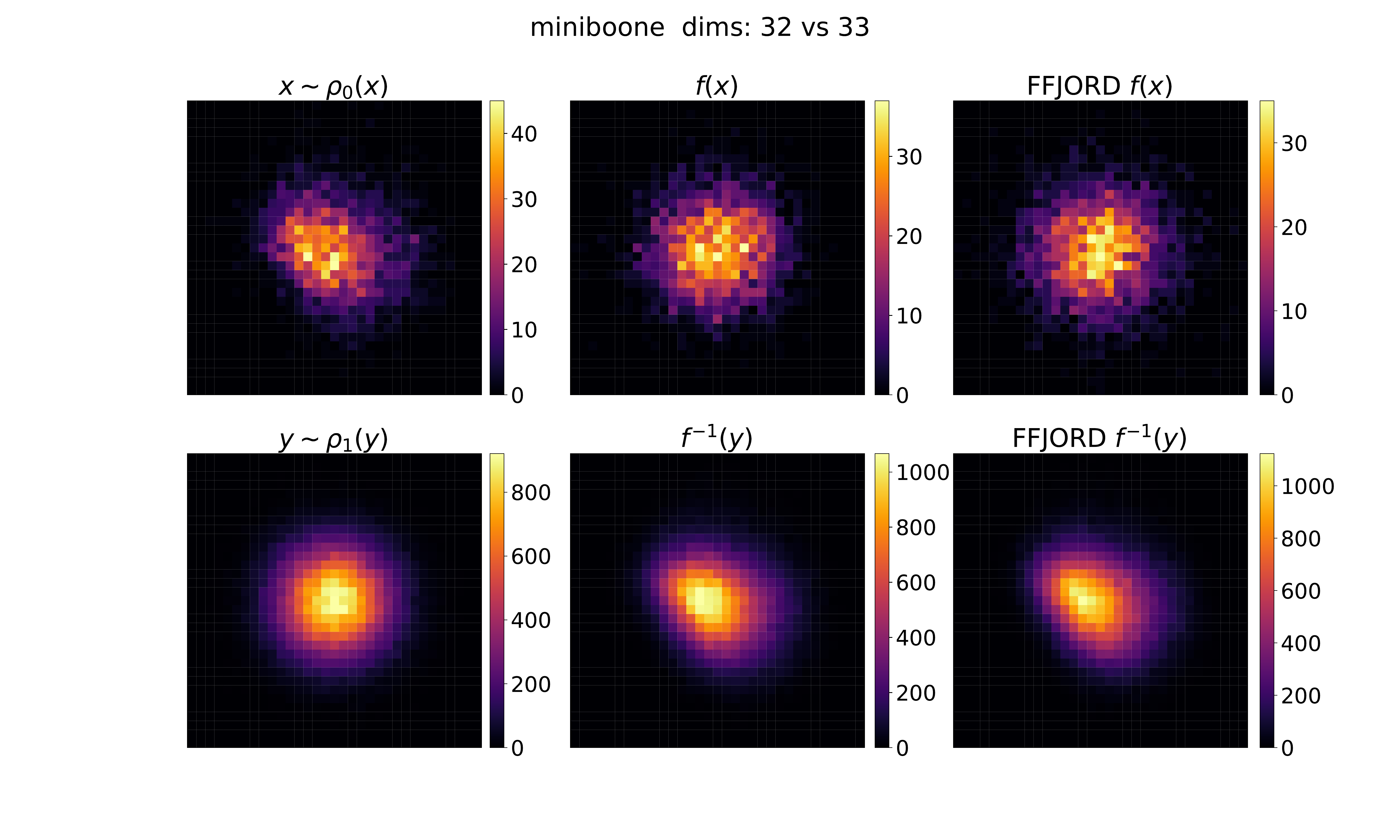}
     \includegraphics[clip, trim=2cm 2.0cm 2cm 0.5cm, width=0.3\linewidth]
         {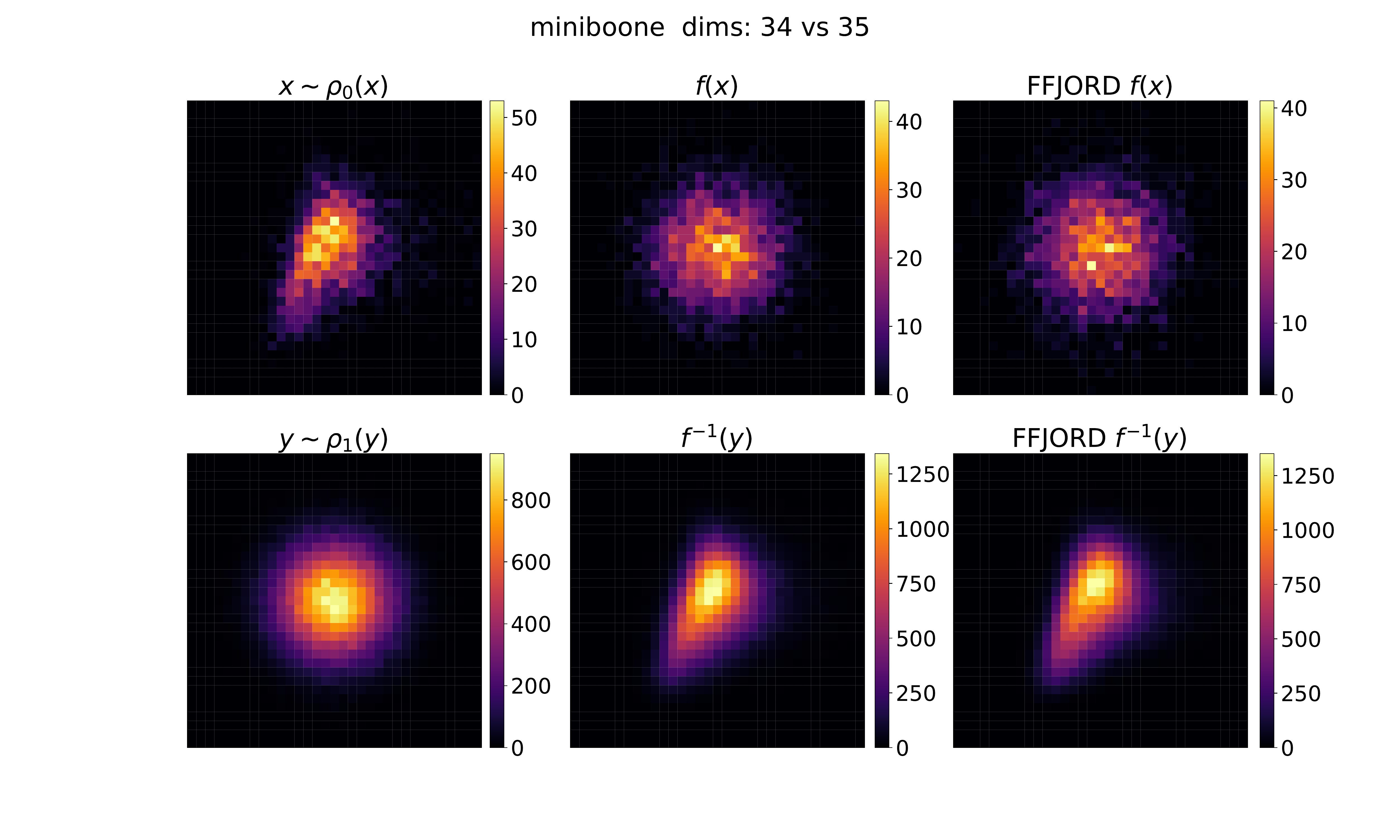}        
     \includegraphics[clip, trim=2cm 2.0cm 2cm 0.5cm, width=0.3\linewidth]
         {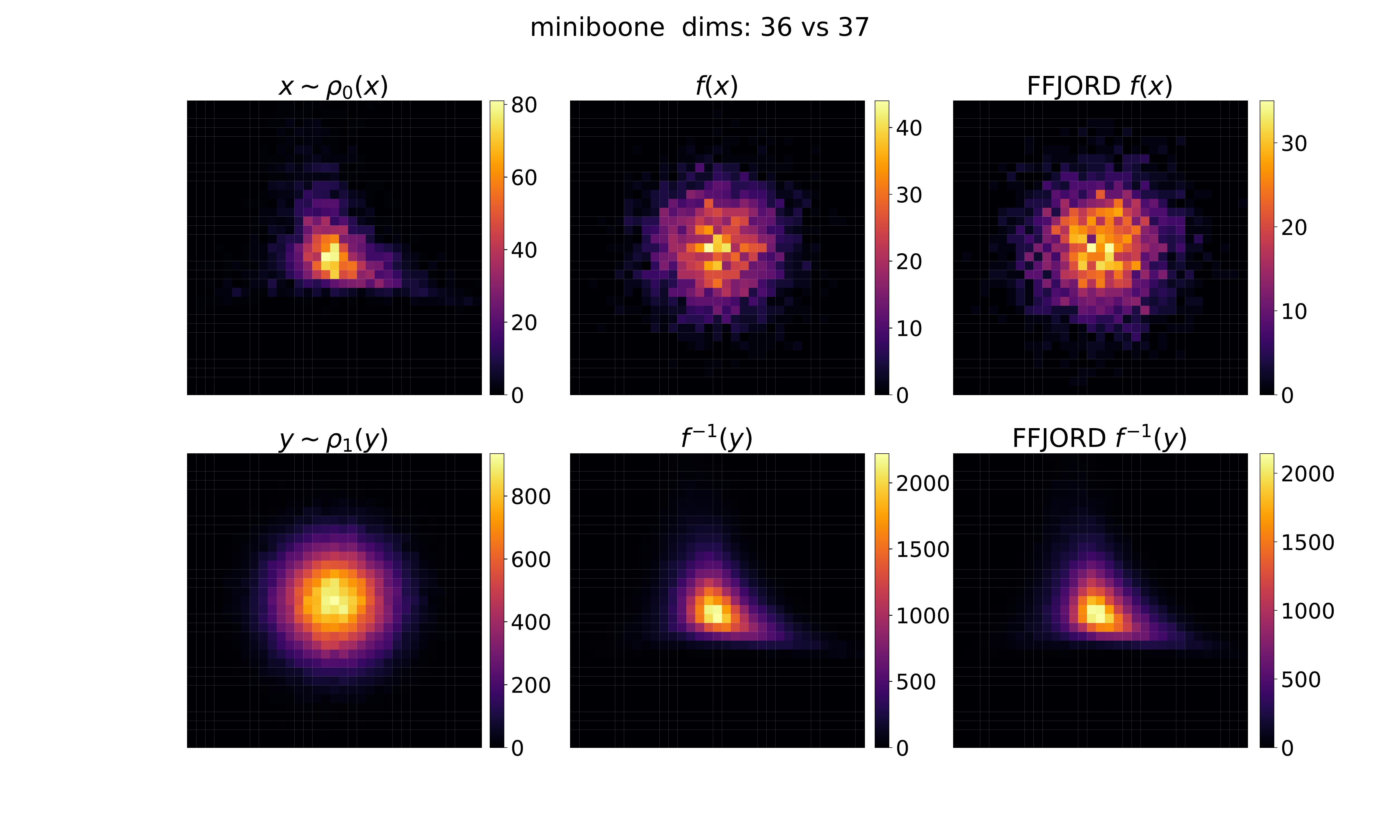} \\        
     \includegraphics[clip, trim=2cm 2.0cm 2cm 0.5cm, width=0.3\linewidth]
         {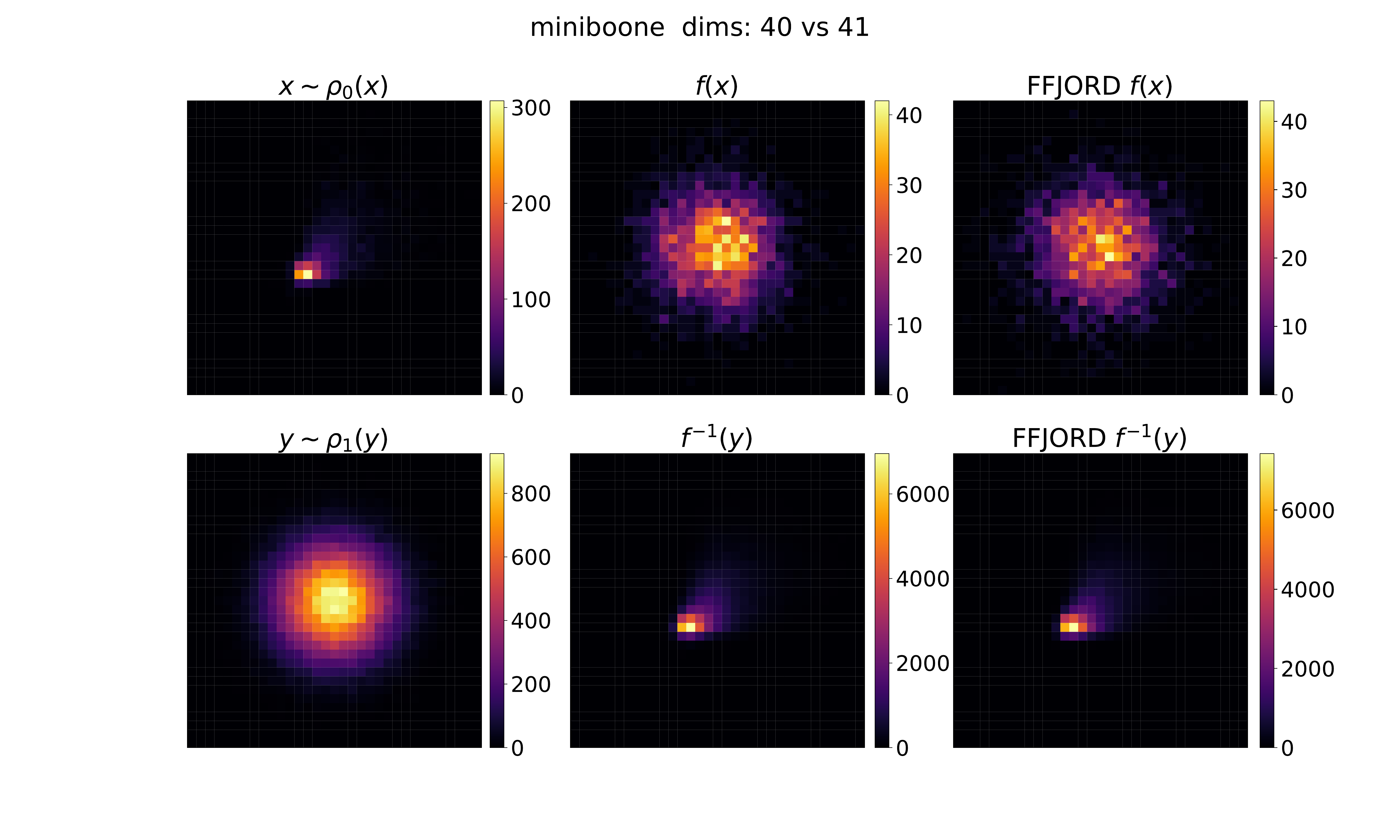}
     \caption{Other two-dimensional slices of the \miniboone{} density estimation to supplement Fig.~\ref{fig:miniboone}.  
     }
     \label{fig:miniboone_full}
 \end{figure*}

\end{document}